\documentclass{article}

\usepackage{microtype}

\usepackage{hyperref}
\usepackage{url}
\usepackage{microtype}

\usepackage{tabularx}
\usepackage{graphicx}
\usepackage{fullpage}
\usepackage{natbib}
\usepackage{delarray}
\usepackage{times}
\usepackage{subfigure}
\usepackage{color}
\usepackage{amssymb}
\usepackage{amsmath}
\usepackage{amsthm}
\usepackage{mathrsfs}
\usepackage{tikz}
\usepackage{xspace}
\usepackage{enumerate}
\usepackage{epstopdf}

\usepackage{algorithm}
\usepackage{algorithmic}
\usepackage{nameref}
\usepackage{booktabs}
\usepackage{mdframed}
\usepackage{fmtcount}
\usepackage{footnote}
\usepackage{bbm}
\usepackage{thmtools,thm-restate}
\declaretheorem{theorem}

\newcommand{\poly}{{\text{poly}}}
\newcommand{\eat}[1]{}

\newcommand{\teta}{\tilde{\eta}}
\newcommand{\calE}{{\mathcal E}}
\newcommand{\bcalE}{{\bar{\mathcal E}}}
\newcommand{\calD}{{\mathcal D}}
\newcommand{\R}{{\mathbb R}}

\newcommand{\E}{{\mathbb{E}}}

\newcommand{\inner}[2]{\left\langle #1, #2 \right\rangle}
\newcommand{\ns}[1]{\left\| #1 \right\|^2}
\newcommand{\n}[1]{\left\| #1 \right\|}

\newcommand{\indic}[1]{\mathbbm{1}\left\{#1\right\}}

\newcommand{\proj}{\mbox{Proj}}

\newcommand{\str}{S_{\text{train}}}
\newcommand{\htr}{H_{\text{train}}}
\newcommand{\wtr}{w_{\text{train}}}
\newcommand{\xtr}{X_{\text{train}}}
\newcommand{\bxtr}{\bar{X}_{\text{train}}}
\newcommand{\etatr}{\eta^*_{\text{train}}}
\newcommand{\xitr}{\xi_{\text{train}}}
\newcommand{\bxitr}{\bar{\xi}_{\text{train}}}

\newcommand{\sva}{S_{\text{valid}}}
\newcommand{\hva}{H_{\text{valid}}}
\newcommand{\wva}{w_{\text{valid}}}
\newcommand{\xva}{X_{\text{valid}}}
\newcommand{\etava}{\eta^*_{\text{valid}}}
\newcommand{\xiva}{\xi_{\text{valid}}}
\newcommand{\bt}{B_{t,\eta}}
\newcommand{\wt}{w_{t,\eta}}
\newcommand{\heta}{\hat{\eta}}

\newcommand{\hftr}{\hat{F}_{TbT}}
\newcommand{\bftr}{F_{TbT}}
\newcommand{\strk}{S_{\text{train}}^{(k)}}
\newcommand{\htrk}{H_{\text{train}}^{(k)}}
\newcommand{\wtrk}{w_{\text{train}}^{(k)}}
\newcommand{\xtrk}{X_{\text{train}}^{(k)}}
\newcommand{\xitrk}{\xi_{\text{train}}^{(k)}}

\newcommand{\hfva}{\hat{F}_{TbV}}
\newcommand{\bfva}{F_{TbV}}

\newcommand{\svak}{S_{\text{valid}}^{(k)}}
\newcommand{\hvak}{H_{\text{valid}}^{(k)}}
\newcommand{\wvak}{w_{\text{valid}}^{(k)}}
\newcommand{\xvak}{X_{\text{valid}}^{(k)}}
\newcommand{\xivak}{\xi_{\text{valid}}^{(k)}}
\newcommand{\btk}{B_{t,\eta}^{(k)}}
\newcommand{\wtk}{w_{t,\eta}^{(k)}}

\newcommand{\dtr}{\Delta_{TbT}(\eta, P)}
\newcommand{\dtrk}{\Delta_{TbT}(\eta, P_k)}
\newcommand{\dva}{\Delta_{TbV}(\eta, P)}
\newcommand{\dvak}{\Delta_{TbV}(\eta, P_k)}

\newcommand{\pr}[1]{\left( #1 \right)}
\newcommand{\br}[1]{\left[ #1 \right]}

\newcommand{\absr}[1]{\left| #1 \right|}

\newcommand{\expn}{\exp(-\Omega(n))}
\newcommand{\expd}{\exp(-\Omega(d))}
\newcommand{\expepsd}{\exp(-\Omega(\epsilon^2 d))}
\newcommand{\expm}{\exp(-\Omega(m))}
\newcommand{\expepsm}{\exp(-\Omega(\epsilon^2 m))}
\newcommand{\expepsn}{\exp(-\Omega(\epsilon^2 n))}

\newcommand{\Esgd}{\E_{\text{SGD}}}
\newcommand{\Estr}{\E}
\newcommand{\Ep}{\mathbb{E}_{P\sim \mathcal{T}}}
\newcommand{\tr}{\text{tr}}

\newcommand{\indick}{\mathbbm{1}_{\calE_k}}
\newcommand{\indicbk}{\mathbbm{1}_{\bar{\calE}_k}}

        {\hspace*{\fill}$\Box$\par\vspace{4mm}}
\newenvironment{proofof}[1]{\smallskip\noindent{\bf Proof of #1.}}%
        {\hspace*{\fill}$\Box$\par}

\newtheorem{lemma}{Lemma}

\newtheorem{definition}{Definition}

\newtheorem{claim}{Claim}

\hypersetup{
    colorlinks=true,       
    linkcolor=black,    
    citecolor=black,        
    filecolor=magenta,     
    urlcolor=blue
}

\begin{document}

\title{Guarantees for Tuning the Step Size using a Learning-to-Learn Approach}

\author{Xiang Wang\thanks{Duke University. Email: xwang@cs.duke.edu}\\ \and Shuai Yuan\thanks{Duke University. Email: shuai@cs.duke.edu}\\ \and Chenwei Wu\thanks{Duke University. Email: chenwei.wu592@duke.edu}\and Rong Ge\thanks{Duke University. Email: rongge@cs.duke.edu}}

\date{}
\maketitle
\begin{abstract}
Choosing the right parameters for optimization algorithms is often the key to their success in practice. Solving this problem using a learning-to-learn approach---using meta-gradient descent on a meta-objective based on the trajectory that the optimizer generates---was recently shown to be effective. However, the meta-optimization problem is difficult. In particular, the meta-gradient can often explode/vanish, and the learned optimizer may not have good generalization performance if the meta-objective is not chosen carefully. In this paper we give meta-optimization guarantees for the learning-to-learn approach on a simple problem of tuning the step size for quadratic loss. Our results show that the na\"ive objective suffers from meta-gradient explosion/vanishing problem. Although there is a way to design the meta-objective so that the meta-gradient remains polynomially bounded, computing the meta-gradient directly using backpropagation leads to numerical issues. We also characterize when it is necessary to compute the meta-objective on a separate validation set to ensure the generalization performance of the learned optimizer. Finally, we verify our results empirically and show that a similar phenomenon appears even for more complicated learned optimizers parametrized by neural networks.
\end{abstract}
\section{Introduction}

Choosing the right optimization algorithm and related hyper-parameters is important for training a deep neural network. 
Even for simple algorithms like gradient descent and stochastic gradient descent, choosing a good step size can be important to the convergence speed and generalization performance. Empirically, the parameters are often chosen based on past experiences or grid search. 
Recently, \citet{maclaurin2015gradient} considered the idea of tuning these parameters by optimization---that is, consider a meta-optimization problem where the goal is to find the best parameters for an optimizer. A series of works (e.g.,~\citet{andrychowicz2016learning, wichrowska2017learned}) extended such ideas and parametrized the set of optimizers by neural networks. 

Although this approach has shown empirical success, there are very few theoretical guarantees for learned optimizers. \citet{gupta2017pac} gave sample complexity bounds for tuning the step size, but they did not address how one can find the learned optimizer efficiently. In practice, the meta-optimization problem is often solved by meta-gradient descent---define a meta-objective function based on the trajectory that the optimizer generates, and then compute the meta-gradient using back-propagation~\citep{franceschi2017forward}. The optimization for meta-parameters is usually a nonconvex problem, therefore it is unclear why simple meta-gradient descent would find an optimal solution.

In this paper we consider using learning-to-learn approach to tune the step size of standard gradient descent/stochastic gradient descent algorithm. Even in this simple setting, many of the challenges still remain and we can get better learned optimizers by choosing the right meta-objective function. Though our results are proved only in the simple setting, we empirically verify the results using complicated learned optimizers with neural network parametrizations. 

\subsection{Our Results}


In this paper we focus on two basic questions on learning-to-learn for gradient descent optimizer. First, will the meta-gradient explode/vanish and is there a way to fix the problem? Second, how could we guarantee that the learned optimizer has good generalization properties?

Our first result shows that meta-gradient can explode/vanish even for tuning the step size for gradient descent on a simple quadratic objective. In this setting, we show that there is a unique local and global minimizer for the step size, and we also give a simple way to get rid of the gradient explosion/vanishing problem.

\begin{theorem}[Informal version of Theorem~\ref{thm:quadratic_exponential} and Theorem~\ref{thm:quadratic_converge}] For tuning the step size of gradient descent on a quadratic objective, if the meta-objective is the loss of the last iteration, then the meta-gradient will explode/vanish. If the meta-objective is the $\log$ of the loss of the last iteration, then the meta-gradient is polynomially bounded. Further, doing meta-gradient descent with a meta step size of $1/\sqrt{k}$ (where $k$ is the number of meta-gradient steps) provably converges to the optimal step size for the inner-optimizer.
\end{theorem}

Surprisingly, even though taking the $\log$ of the objective solves the meta-gradient explosion/vanishing problem, one cannot simply implement such an algorithm using back-propagation (which is standard in auto-differentiation tools such as those used in TensorFlow~\citep{abadi2016tensorflow}). The reason is that even though the meta-gradient is polynomially bounded, back-propagation algorithm will compute the meta-gradient as the ratio of two exponentially large/small numbers, which causes numerical issues. Detailed discussion for the first result appears in Section~\ref{sec:quadratic}.


Our second result shows that defining meta-objective on the same training set (later referred to as the ``train-by-train'' approach) could lead to overfitting; while defining meta-objective on a separate validation set (``train-by-validation'', see \citet{metz2019understanding}) can solve this issue. We consider a simple least squares setting where $y = \inner{w^*}{x}+\xi$ and $\xi\sim \mathcal{N}(0,\sigma^2)$. We show that when the number of samples is small and the noise is large, it is important to use train-by-validation; while when the number of samples is much larger train-by-train can also learn a good optimizer.

\begin{theorem}[Informal version of Theorem~\ref{thm:train_valid_GD} and Theorem~\ref{thm:large_sample}]
For a least squares problem  in $d$ dimensions, if the number of samples $n$ is a constant fraction of $d$ (e.g., $d/2$), and the samples have large noise, then the train-by-train approach performs much worse than train-by-validation. On the other hand, when the number of samples $n$ is large, train-by-train can get close to error $d\sigma^2/n$, which is optimal.
\end{theorem}

We discuss the details in Section~\ref{sec:least-squares}. In Section~\ref{sec:experiments} we show that such observations also hold empirically for more complicated learned optimizers---an optimizer parametrized by a neural network. 

\subsection{Related Work}

\paragraph{Learned optimizer}
The idea of learning an optimizer has appeared in early works decades ago~\citep{bengio1990learning,bengio1992optimization,hochreiter2001learning}. Recently, with the rise of deep learning, researchers started to consider more complex optimizers on more challenging tasks. One line of research views the optimizer as a policy and apply reinforcement learning techniques to train it~\citep{li2016learning,li2017learning,bello2017neural}. The other line of papers use gradient descent on the meta-objective to update the optimizer parameters~\citep{maclaurin2015gradient,andrychowicz2016learning,lv2017learning,wichrowska2017learned,metz2019understanding}. 

Mostly relevant to our work, \citet{metz2019understanding} highlighted several challenges in the meta-optimization for learning-to-learn approach. First, they observed the meta-gradient exploding/vanishing issue and proposed to use a gradient estimator for a variational meta-objective. They also observed that train-by-train approach can overfit the training tasks while train-by-validation generalizes well.

\paragraph{Data-driven algorithm design} In data-driven algorithm design, we aim to find an algorithm that works well on a particular distribution of tasks. \citet{gupta2017pac} first modeled this algorithm-selection process as a statistical learning problem. In particular, they analyzed the sample complexity of choosing the step size for gradient descent. But they didn't consider the meta-optimization problem. They also restricted the step size into a small range so that gradient descent is guaranteed to converge on every task. We don't have such a restriction and allow the meta-learning to choose a more aggressive step size.


Following the work by \citet{gupta2017pac}, data-driven algorithms have been studied in many problems, including partitioning and clustering \citep{balcan2016learning}, tree search \citep{balcan2018learning}, pruning \citep{alabi2019learning} and machanism design \citep{morgenstern2015pseudo, morgenstern2016learning,balcan2016sample,balcan2018general}.



\paragraph{Step size schedule for GD/SGD}
\citet{shamir2013stochastic} showed that SGD with polynomial step size scheduling can almost match the minimax rate in convex non-smooth settings, which was later tightened by \citet{harvey2018tight} for standard step size scheduling. Assuming that the number of training steps is known to the algorithm, the information-theoretically optimal bound in convex non-smooth setting was later achieved by \citet{jain2019making} which used another step size schedule, and \citet{ge2019step} showed that exponentially decaying step size scheduling can achieve near optimal rate for least squares regression. 

A closely related paper that appeared later than our work also studied the comparison between train-by-train and train-by-validation~\citep{bai2020important}. They considered a very different meta-learning problem, where the goal is to find the best common initialization for adapting to a linear predictor on each task. They proved train-by-train can work better than train-by-validation in the noiseless setting. 


\section{Preliminaries}\label{sec:prelim}
In this section, we first introduce some notations, then formulate the learning-to-learn framework.

\subsection{Notations}
For any integer $n,$ we use $[n]$ to denote $\{1,2,\cdots,n\}.$ We use $\n{\cdot }$ to denote the $\ell_2$ norm for a vector and the spectral norm for a matrix. We use $\inner{\cdot}{\cdot}$ to denote the inner product of two vectors. For a symmetric matrix $A\in\R^{d\times d},$ we denote its eigenvalues as $\lambda_1(A)\geq \cdots \geq \lambda_d(A).$ We denote the $d$-dimensional identity matrix as $I_d$ or simply as $I$ when the dimension is clear. We use $O(\cdot),\Omega(\cdot),\Theta(\cdot)$ to hide constant factor dependencies. We use $\poly(\cdot)$ to represent a polynomial on the relevant parameters with constant degree. 



\subsection{Learning-to-learn Framework}

We consider the learning-to-learn approach applied to training a distribution of learning tasks. Each task is specified by a tuple $(\mathcal{D}, \str, \sva, \ell)$. Here $\mathcal{D}$ is a distribution of samples in $X\times Y$, where $X$ is the domain for the sample and $Y$ is the domain for the label/value. The sets $\str$ and $\sva$ are samples generated independently from $\mathcal{D}$, which serve as the training and validation set (the validation set is optional). The learning task looks to find a parameter $w \in W$ that minimizes the loss function $\ell(w, x,y):W\times X\times Y\to \R$, which gives the loss of the parameter $w$ for sample $(x,y)$. 
The training loss for this task is $$\hat{f}(w) := \frac{1}{|\str|} \sum_{(x,y)\in \str} \ell(w,x,y),$$ while the population loss is $f(w) := \mathbb{E}_{(x,y)\sim \mathcal{D}}[\ell(w,x,y)].$

The goal of inner-optimization is to minimize the population loss $f(w)$. For the learned optimizer, we consider it as an update rule $u(\cdot)$ on weight $w$. The update rule is a parameterized function that maps the weight at step $\tau$ and its history to the step $\tau+1:$ $w_{\tau+1}=u(w_{\tau},\nabla \hat{f}(w_\tau), \nabla \hat{f}(w_{\tau-1}), \cdots; \theta).$  In most parts of this paper, we consider the update rule $u$ as gradient descent mapping with step size as the trainable parameter (here $\theta = \eta$ which is the step size for gradient descent). That is, $u(w; \eta) = w - \eta \nabla \hat{f}(w)$ for gradient descent and $u(w; \eta) = w - \eta \nabla_w \ell(w, x, y)$ for stochastic gradient descent where $(x,y)$ is a sample randomly chosen from the training set $\str$.

In the outer (meta) level, we consider a distribution $\mathcal{T}$ of tasks. For each task $P \sim \mathcal{T}$, we can define a meta-loss function $\Delta(\theta, P)$. 
The meta-loss function measures the performance of the optimizer on this learning task. The meta-objective, for example, can be chosen as the target training loss $\hat{f}$ at the last iteration (train-by-train), or the loss on the validation set (train-by-validation). 

The training loss for the meta-level is the average of the meta-loss across $m$ different specific tasks $P_1,P_2,...,P_m$, that is, $$\hat{F}(\theta) = \frac{1}{m} \sum_{i=1}^m \Delta(\theta, P_k).$$ The population loss for the meta-level is the expectation over all the possible specific tasks $F(\theta) = \mathbb{E}_{P\sim \mathcal{T}}[\Delta(\theta, P)].$ 

In order to train an optimizer by gradient descent, we need to compute the gradient of meta-objective $\hat{F}$ in terms of meta parameters $\theta$.  The meta parameter is updated once after applying the optimizer on the inner objective $t$ times to generate the trajectory $w_0, w_1, ..., w_t$. The meta-gradient is then computed by unrolling the optimization process and back-propagating through the $t$ applications of the optimizer. 

\section{Alleviating Gradient Explosion/Vanishing Problems}
\label{sec:quadratic}

First we consider the meta-gradient explosion/vanishing problem. More precisely, we say the meta-gradient explodes/vanishes if it is exponentially large/small with respect to the number of steps $t$ of the inner-optimizer.

In this section, we consider a simple instance of the learning-to-learn approach, where the distribution $\mathcal{T}$ only contains a single task $P$, and the task also just defines a single loss function $f$\footnote{In the notation of Section~\ref{sec:prelim}, one can think that $\mathcal{D}$ contains a single point $(0,0)$ and the loss function $f(w) = \ell(w,0,0)$.}. Therefore, in this section $\hat{F}(\eta) = F(\eta) = \Delta(\eta, P)$. We will simplify notation and only use $\hat{F}(\eta)$. 

The inner task $P$ is a simple quadratic problem, where the starting point is fixed at $w_0$ with unit norm, and the loss function is $f(w) = \frac{1}{2} w^\top H w$ for some fixed positive definite matrix $H\in\R^{d\times d}$. 

Let $\{w_{\tau,\eta}\}_{\tau = 0}^t $ be the GD sequence running on $f(w)$ starting from $w_0$ with step size $\eta.$ 
 We consider two ways of defining meta-objective: using the loss of the last point directly or using the $\log$ of this value. We first show that although choosing $\hat{F}(\eta) = f(\wt)$ does not have any bad local optimal solution, it has the meta-gradient explosion/vanishing problem. We use $\hat{F}'(\eta)$ to denote the derivative of $\hat{F}$ in $\eta.$

In the analysis, we use eigen-decomposition to transform $H$ into a diagonal matrix. We introduce related notations here: suppose the eigenvalue decomposition of $H$ is $\sum_{i=1}^d \lambda_i u_iu_i^\top.$ We denote $L := \lambda_1(H)$ and $\alpha := \lambda_d(H)$ as the largest and smallest eigenvalues of $H$. For each $i\in[d],$ let $c_i$ be $\inner{w_0}{u_i}$ and let $c_{\min}$ be $\min(|c_1|,|c_d|).$ We assume $c_{\min}>0$ and $L>\alpha$ for simplicity\footnote{If $w_0$ is uniformly sampled from the unit sphere, with high probability $c_{\min}$ is at least $\Omega(1/\sqrt{d})$; if $H$ is $XX^\top$ with $X\in\R^{d\times 2d}$ as a random Gaussian matrix, with constant probability, both $\alpha$ and $L-\alpha$ are at least $\Omega(d).$}.

\begin{restatable}
{theorem}{quadraticExponential}\label{thm:quadratic_exponential}
Let the meta-objective be $\hat{F}(\eta)=f(\wt),$ we know $\hat{F}(\eta)$ is a strictly convex function in $\eta$ with an unique minimizer. However, for any step size $0<\eta<2/L$, $$|\hat{F}'(\eta)|\leq t L^2\max(|1-\eta\alpha|^{2t-1},|1-\eta L|^{2t-1});$$ for any step size $\eta>2/L,$
$$ |\hat{F}'(\eta)|\geq c_1^2 L^2 t(\eta L-1)^{2t-1}-L^2 t.$$
\end{restatable}

Note that in Theorem~\ref{thm:quadratic_exponential}, when $0<\eta<2/L,$ $|\hat{F}'(\eta)|$ is exponentially small because $|1-\eta\alpha|, |1-\eta L|<1$ ; when $\eta>2/L,$ $|\hat{F}'(\eta)|$ is exponentially large because $\eta L-1>1$. The strict convexity of $\hat{F}(\eta)$ is proved by showing the second order derivative of $\hat{F}(\eta)$ is positive; the upper and lower bounds of $\hat{F}'(\eta)$ follows from direct calculation. 

Intuitively, gradient explosion/vanishing happens because the meta-objective becomes too small or too large. A natural idea to fix the problem is to take the $\log$ of the meta-objective to reduce its range. If we choose $\hat{F}(\eta) = \frac{1}{t}\log f(\wt)$, we have

\begin{restatable}
{theorem}{quadraticConverge}\label{thm:quadratic_converge}
Let the meta-objective be $\hat{F}(\eta) = \frac{1}{t}\log f(\wt)$. We know $\hat{F}(\eta)$ has a unique minimizer $\eta^*$ and 
$\hat{F}'(\eta)=O\pr{\frac{L^3}{c_{\min}^2 \alpha(L-\alpha)}}$
for all $\eta\geq 0.$
Let $\{\eta_k\}$ be the GD sequence running on $\hat{F}$ with meta step size $\mu_k=1/\sqrt{k}.$ Suppose the starting step size $\eta_0\leq M.$ Given any $1/L>\epsilon>0$, there exists $k'=\frac{M^6}{\epsilon^2}\poly(\frac{1}{c_{\min}},L,\frac{1}{\alpha},\frac{1}{L-\alpha})$ such that for all $k\geq k',$
$|\eta_k-\eta^*|\leq \epsilon.$
\end{restatable}

For convenience, in the above algorithmic result, we reset $\eta$ to zero once $\eta$ goes negative (this corresponds to doing a projected gradient descent on $\eta$ under constraint $\eta \ge 0$). We give a proof sketch of Theorem~\ref{thm:quadratic_converge} in Section~\ref{sec:proof_sketch_quadratic}.




Surprisingly, even though we showed that the meta-gradient is well-behaved, it cannot be effectively computed by doing back-propagation due to numerical issues. More precisely:

\begin{restatable}
{corollary}{expolodeLog}\label{cor:explode_log}
If we choose the meta-objective as $\hat{F}(\eta) = \frac{1}{t}\log f(\wt)$, when computing the meta-gradient using back-propagation, there are intermediate results that are exponentially large/small in number of inner-steps $t$.
\end{restatable}

If we use back-propagation to compute $\hat{F}'(\eta),$ we need to separately compute the numerator and denominator in Eqn.~\eqref{eqn:gradient_formula}, which are exponentially large or small as we showed in Theorem~\ref{thm:quadratic_exponential}.
Indeed, in Section~\ref{sec:experiments} we empirically verify that standard auto-differentiation tools can fail in this setting. In contrast, the meta training succeeds if we use the formula derived in Section~\ref{sec:proof_sketch_quadratic} (Eqn.~\eqref{eqn:avoid_gradient_vanishing}).
This suggests that one should be more careful about using standard back-propagation in the learning-to-learn approach. 
The proofs of the results in this section are deferred into Appendix~\ref{sec:proofs_quadratic}.

\subsection{Proof Sketch of Theorem~\ref{thm:quadratic_converge}}\label{sec:proof_sketch_quadratic}
Throughout the proof, we work in the eigenspace of $H$ which reduces the problem to having a diagonal matrix $H$. The proof goes in three steps:
\begin{itemize}
\item Claim~\ref{clm:minimizer1} shows that the meta-objective $\hat{F}$ has a unique minimizer $\eta^*$ and the minus meta-gradient always points to the minimizer.
\item Claim~\ref{clm:max2} shows meta-gradient $\hat{F}'(\eta)$ never explodes.
\item Claim~\ref{clm:min3} shows meta-gradient is large when $\eta$ is far from $\eta^*$. 
\end{itemize}

\begin{claim}\label{clm:minimizer1}
The meta-objective $\hat{F}$ has only one stationary point that is also its unique minimizer $\eta^*$. For any $\eta\in[0,\eta^*),$ $\hat{F}'(\eta)<0$ and for any $\eta\in(\eta^*,\infty),$ $\hat{F}'(\eta)>0.$ 
\end{claim}

The lemma follows from a direct calculation $\hat{F}'(\eta)$:
\begin{equation}\label{eqn:gradient_formula}
\hat{F}'(\eta)
= \frac{-2\sum_{i=1}^d c_i^2\lambda_i^2(1-\eta\lambda_i)^{2t-1}}{\sum_{i=1}^d c_i^2\lambda_i(1-\eta\lambda_i)^{2t}}.
\end{equation}
Claim~\ref{clm:minimizer1} is proved by noticing that the denominator in $\hat{F}'(\eta)$ is always positive and the numerator is strictly increasing in $\eta.$ Next, we show the meta derivative is polynomially upper bounded. 

\begin{claim}\label{clm:max2}
For any $\eta\in[0,\infty),$ we have 
$ |\hat{F}'(\eta)|\leq \frac{4L^3}{c_{\min}^2 \alpha(L-\alpha)}.$
\end{claim}
To prove this claim we observe that the numerator and denominator are both polynomially bounded once we divide them by a common factor, which is $(1-\eta\alpha)^{2t}$ when $\eta\in[0,\frac{2}{\alpha+L}]$. More precisely we have when $\eta\in[0,\frac{2}{\alpha+L}]$ 
\begin{equation}\label{eqn:avoid_gradient_vanishing}
\absr{\hat{F}'(\eta)}
= 2\frac{\absr{\sum_{i=1}^{d} \frac{c_i^2\lambda_i^2}{1-\eta\alpha}(\frac{1-\eta\lambda_i}{1-\eta\alpha})^{2t-1}}}{c_d^2\alpha + \sum_{i=1}^{d-1} c_i^2\lambda_i(\frac{1-\eta\lambda_i}{1-\eta\alpha})^{2t}}
\leq 
\frac{2\sum_{i=1}^d c_i^2\lambda_i^2 }{c_d^2\alpha(1-\eta\alpha)}
.
\end{equation}
This leads to the claimed bounds based on our assumptions. The case when $\eta$ is large is similar. Finally, we show the meta-gradient is lower bounded if $\eta$ is away from $\eta^*$ and is not too large. The proof follows from a similar calculation as above. 

\begin{claim}\label{clm:min3}
Given $\hat{M}\geq 2/\alpha$ and $1/L>\epsilon>0,$ for any $\eta\in[0,\eta^*-\epsilon]\cup[\eta^*+\epsilon,\hat{M}],$ we have
$|F'(\eta)|
\geq 2\epsilon c_{\min}^2\min\pr{\frac{\alpha^3}{L} ,\frac{1}{\hat{M}^2 }}
.$
\end{claim}

With the above three claims, we are ready to sketch the proof of Theorem~\ref{thm:quadratic_converge}. Due to Claim~\ref{clm:minimizer1}, we know the minus meta-gradient always points to the minimizer $\eta^*$. This alone is not sufficient to prove the convergence result because the iterates might significantly overshoot the minimizer if $|\hat{F}'|$ is too large or the iterates might converge very slowly if $|\hat{F}'|$ is too small. Fortunately, these two problematic cases can be excluded by Claim~\ref{clm:max2} and Claim~\ref{clm:min3}. 

\section{Generalization for Trained Optimizer}
\label{sec:least-squares}
Next we consider the generalization ability of simple trained optimizers. In this section we consider a simple family of least squares problems. Let $\mathcal{T}$ be a distribution of tasks where every task $(\mathcal{D}(w^*),\str,\sva,\ell)$ is determined by a parameter $w^* \in \R^d$ that is sampled uniformly at random from the unit sphere. 
For each individual task, $(x,y) \sim \mathcal{D}(w^*)$ is generated by first choosing $x\sim \mathcal{N}(0, I_d)$ and then computing $y = \inner{w^*}{x} + \xi$ where $\xi \sim \mathcal{N}(0, \sigma^2)$ with $\sigma\geq 1$. The loss function $\ell(w,x,y)$ is just the squared loss $\ell(w,x,y) = \frac{1}{2}(y - \inner{w}{x})^2$. That is, the tasks are just standard least-squares problems with ground-truth equal to $w^*$ and noise level $\sigma^2$. 

We consider two different ways to define the meta-objective.
\paragraph{Train-by-train:}
 In the train-by-train setting, the training set $\str$ contains $n$ independent samples, and the meta-loss function is chosen to be the training loss. That is, in each task $P$, we first choose $w^*$ uniformly at random, then generate $(x_1,y_1),...,(x_n,y_n)$ as the training set $\str$. The meta-loss function $\Delta_{TbT(n)}(\eta,P)$ is defined to be 
$$
\Delta_{TbT(n)}(\eta, P) = \frac{1}{2n}\sum_{i=1}^n (y_i - \inner{\wt}{x_i})^2.$$

Here $\wt$ is the result of running $t$ iterations of gradient descent starting from point $0$ with step size $\eta.$ Note we truncate a sequence and declare the meta loss is high once the weight norm exceeds certain threshold\footnote{Specifically, if at the $\tau$-th step $\n{w_{\tau,\eta}}\geq 40\sigma,$ we freeze the training on this task and set $w_{\tau',\eta}=40\sigma u$ for all $\tau\leq \tau'\leq t$, for some arbitrary vector $u$ with unit norm. Setting the weight to a large vector is just one way to declare the loss is high.}. We can safely do this because we assume the ground truth weight $w^*$ has unit norm, so if the weight norm is too high, it means the inner training has diverged and the step size is too large. 

As before, the empirical meta-objective in train-by-train setting is the average of the meta-loss across $m$ different specific tasks $P_1,P_2,...,P_m$, that is,
\begin{equation}
    \hat{F}_{TbT(n)}(\eta) = \frac{1}{m} \sum_{k=1}^m \Delta_{TbT(n)}(\eta, P_k). \label{eqn:train_GD}
\end{equation}

\paragraph{Train-by-validation:} In the train-by-validation setting, the specific tasks are generated by sampling $n_1$ training samples and $n_2$ validation samples for each task, and the meta-loss function is the validation loss. That is, in each specific task $P$, we first choose $w^*$ uniformly at random, then generate $(x_1,y_1),...,(x_{n_1},y_{n_1})$ as the training set $\str$ and $(x'_1,y'_1),...,(x'_{n_2},y'_{n_2})$ as the validation set $\sva$. The meta-loss function $\Delta_{TbV(n_1,n_2)}(\eta,P)$ is defined to be 
$$
\Delta_{TbV(n_1,n_2)}(\eta, P) = \frac{1}{2n_2}\sum_{i=1}^{n_2} (y'_i - \inner{\wt}{x'_i})^2.
$$
Here again $\wt$ is the result of running $t$ iterations of the gradient descent on the training set starting from point 0, and we use the same truncation as before. 
The empirical meta-objective is defined as
\begin{equation}
    \hat{F}_{TbV(n_1,n_2)}(\eta) = \frac{1}{m} \sum_{k=1}^m \Delta_{TbV(n_1,n_2)}(\eta, P_k), \label{eqn:valid_GD}
\end{equation} 
where each $P_k$ is independently sampled according to the described procedure.

We first show that when the number of samples is small (in particular $n < d$) and the noise is a large enough constant, train-by-train can be much worse than train-by-validation, even when $n_1+n_2 = n$ (the total number of samples used in train-by-validation is the same as in train-by-train)

\begin{restatable}
{theorem}{thmTrainValidGD}\label{thm:train_valid_GD}
Let $\hat{F}_{TbT(n)}(\eta)$ and $\hat{F}_{TbV(n_1,n_2)}(\eta)$ be as defined in Equation~\eqref{eqn:train_GD} and Equation~\eqref{eqn:valid_GD} respectively. Assume $n,n_1,n_2\in[d/4,3d/4].$ Assume noise level $\sigma$ is a large constant $c_1$. Assume unroll length $t\geq c_2$, number of training tasks $m\geq c_3\log(mt)$ and dimension $d\geq c_4\log(mt)$ for certain constants $c_2,c_3,c_4.$ With probability at least $0.99$ in the sampling of training tasks, we have
\[
\etatr=\Theta(1) \text{ and } \E \ns{w_{t,\etatr}-w^*}=\Omega(1)\sigma^2,
\]
for all $\etatr\in\arg\min_{\eta\geq 0} \hat{F}_{TbT(n)}(\eta);$ 
\[
\etava=\Theta(1/t)\mbox{ and }\E \ns{w_{t,\etava}-w^*}=\ns{w^*}-\Omega(1)
\]
for all $\etava\in \arg\min_{\eta\geq 0} \hat{F}_{TbV(n_1,n_2)}(\eta).$ In both equations the expectation is taken over new tasks.
\end{restatable}

In Theorem~\ref{thm:train_valid_GD}, $w_{t,\etatr}$ and $w_{t,\etava}$ are the results obtained on the new task and $w^*$ is the ground truth of the new task. If $\sigma$ is a large enough constant, we know $\E \ns{w_{t,\etatr}-w^*}$ is larger than $\E \ns{w_{t,\etava}-w^*}$ by some constant. The probability $0.99$ is an arbitrary number, which can be replaced by any constant smaller than $1$. 


Note that in this case, the number of samples $n$ is smaller than $d$, so the least square problem is under-determined and the optimal training loss would go to 0 (there is always a way to simultaneously satisfy all $n$ equations). This is exactly what train-by-train would do---it will choose a large constant learning rate which guarantees the optimizer converges exponentially to the empirical risk minimizer (ERM)\footnote{In an under-determined problem, there are actually multiple ERM solutions. Here, we focus on the unique ERM solution in the span of training data. This is also the solution that GD converges to when the initialization is $0$.}. However, when the noise is large making the training loss go to 0 will overfit to the noise and hurt the generalization performance. In contrast, train-by-validation will choose a smaller learning rate which allows it to leverage the signal in the training samples without overfitting to noise.

We separately give a proof sketch for the train-by-train setting and train-by-validation setting in Section~\ref{sec:proof_sketch_train} and Section~\ref{sec:proof_sketch_valid}, respectively. The detailed proof of Theorem~\ref{thm:train_valid_GD} is deferred to Appendix~\ref{sec:proofs_GD}. We also prove similar results for SGD in Appendix~\ref{sec:proofs_SGD} 

We emphasize that neural networks are often over-parameterized, which corresponds to the case when $d > n$.  
Therefore in order to train neural networks, it is usually better to use train-by-validation. On the other hand, we show when the number of samples is large ($n \gg d$), train-by-train can also perform well.

\begin{restatable}
{theorem}{thmLargeSample}\label{thm:large_sample}
Let $\hat{F}_{TbT(n)}(\eta)$ be as defined in Equation~\eqref{eqn:train_GD}.
 Assume noise level is a constant $c_1.$ Given any $1>\epsilon>0,$ assume training set size $n\geq \frac{cd}{\epsilon^2}\log(\frac{nm}{\epsilon d})$, unroll length $t\geq c_2\log(\frac{n}{\epsilon d})$, number of training tasks $m\geq \frac{c_3n^2}{\epsilon^4 d^2}\log(\frac{tnm}{\epsilon d})$ and dimension $d\geq c_4$ for certain constants $c,c_2,c_3,c_4.$ With probability at least $0.99$ in the sampling of training tasks, we have 
 \[
\E \ns{w_{t,\etatr}-w^*}\leq (1+\epsilon)\frac{d\sigma^2}{n},
\]
for all $\etatr\in\arg\min_{\eta\geq 0} \hat{F}_{TbT(n)}(\eta),$ where the expectation is taken over new tasks.
\end{restatable}

Therefore if the learning-to-learn approach is applied to a traditional optimization problem that is not over-parameterized, train-by-train can work well. In this case, the empirical risk minimizer (ERM) already has good generalization performance, and train-by-train optimizes the convergence towards the ERM. We defer the proof of Theorem~\ref{thm:large_sample} into Appendix~\ref{sec:proofs_large_sample}.

\subsection{Proof Sketch for Train-by-train}\label{sec:proof_sketch_train}
In this section, we will give a proof sketch for the first half of Theorem~\ref{thm:train_valid_GD} (train-by-train with small number of samples). At the end of this section, we will briefly discuss the proof of Theorem~\ref{thm:large_sample} (train-by-train with large number of samples). For convenience, we denote $\hftr$ as the empirical meta-objective and $\bftr$ as the population meta-objective. We implicit assume the conditions in Theorem~\ref{thm:train_valid_GD} hold in the following lemmas.


Our meta-optimization problem works on a distribution of tasks. Since different tasks can have different smoothness condition, it's possible that under the same step size, the inner training converges on some tasks, but diverges on others. One way to avoid this issue is to restrict the step size into a small range under which the inner training converges on all tasks~\citep{gupta2017pac}. But this is too conservative and may lead to suboptimal step size. Instead, we allow any positive step size and truncate the inner training if the weight norm goes too large. This approach resolves the diverging issues and also allow the meta-learning algorithm to choose a more aggressive step size. As we explain later, this brings some technical challenges into our proof. 

In order to prove $\E\ns{w_{t,\etatr}-w^*}$ is large, we only need to show the population meta-objective $\bftr(\etatr)$ is small. This is because $\bftr(\etatr)$ measures the distance between $w_{t,\etatr}$ and the ERM solution while ERM solution is far from $w^*$. Since $\etatr$ minimizes the empirical meta-objective, we know $\hftr(\etatr)$ is small. Thus we only need to show $\bftr$ and $\hftr$ are similar. 
This is easy to prove for small step sizes when the inner training always converges, but is difficult when the inner training can diverge and gets truncated. To address this problem we break the step size into three intervals separated by $1/L$ and $\tilde{\eta}$ ($L$ is a large constant that bounds the smoothness on all tasks). Intuitively, when $\eta\leq 1/L$ almost all inner training converges and larger step size leads to faster convergence and smaller $\hftr$; on the other hand, when $\eta > \tilde{\eta}$, we show $\hftr(\eta)$ is always large so the minimizer of $\hftr$ cannot be in this region. Therefore, the optimal step size must be in $[1/L, \teta]$. We only need to prove in the interval $[1/L, \tilde{\eta}]$ the empirical meta-objective $\hftr$ is close to the population meta-objective $\bftr$. This proof is still nontrivial since the inner training can still diverge on a small fraction of sampled tasks.



We first show that for $\eta\in[0,1/L],$ the empirical meta-objective $\hftr$ strictly decreases as $\eta$ increases and $\hftr$ is exponentially small in $t$ at step size $1/L$.

\begin{restatable}
{lemma}{smallStepTrain}\label{lem:small_step_train}
With probability at least $1-m\expd,$ $\hftr(\eta)$ is monotonically decreasing in $[0,1/L]$ and 
$$ \hftr(1/L)\leq 2L^2\sigma^2\pr{1-\frac{1}{L^2}}^t.$$
\end{restatable}

Next we show that the minimizer cannot be larger than $\tilde{\eta}$ for suitably chosen $\tilde{\eta}$ (see the precise definition in the appendix). Intuitively, this is because when $\eta$ is too large the inner-optimizer would diverge on a significant fraction of the sampled tasks.

\begin{restatable}
{lemma}{divergeTrainGD}\label{lem:diverge_train_GD}
With probability at least $1-\expm,$
$$\hftr(\eta)\geq \frac{\sigma^2}{10L^8}$$
for all $\eta>\teta.$
\end{restatable}



By Lemma~\ref{lem:small_step_train} and Lemma~\ref{lem:diverge_train_GD}, we know when $t$ is large enough, the optimal step size $\etatr$ must lie in $[1/L,\teta].$ We can also show $1/L< \teta< 3/L,$ so $\etatr$ is a constant. To relate the empirical loss at $\etatr$ to the population loss, we prove the following uniform convergence result when $\eta \in [1/L,\teta].$ 

\begin{lemma}\label{lem:genera_train_GD_informal}
With probability at least $1-m\expd-O(t+m)\expm,$ 
$$|\bftr(\eta)-\hftr(\eta)| \leq \frac{\sigma^2}{L^3},$$
for all $\eta\in[1/L,\teta].$
\end{lemma}

The proof of this Lemma involves constructing special $\epsilon$-nets for $\bftr$ and $\hftr$ and showing that for each fixed $\eta$, $|\bftr(\eta)-\hftr(\eta)|$ is small with high probability using concentration inequalities. 

Combining the above lemmas, we know the population meta-objective $\bftr$ is small at $\etatr,$ which means $w_{t,\etatr}$ is close to the ERM solution. Since the ERM solution overfits to the noise in the training samples, we know $\E \n{w_{t,\etatr}-w^*}$ has to be large. 

\paragraph{Train-by-train with large number of samples:} The proof of Theorem~\ref{thm:large_sample} follows the same strategy as above. We prove that under the optimal step size $\etatr$, $w_{t,\etatr}$ converges to the ERM solution. But with more samples, the ERM solution $w_{\text{ERM}}$ becomes closer to the ground truth $w^*$. More precisely, we can prove $\E\ns{w_{\text{ERM}}-w^*}$ is roughly $\frac{d\sigma^2}{n},$ which leads to the bound in Theorem~\ref{thm:large_sample}.


\subsection{Proof Sketch for Train-by-Validation}\label{sec:proof_sketch_valid}
In this section, we give a proof sketch for the second half of Theorem~\ref{thm:train_valid_GD}. We denote $\hfva$ as the empirical meta-objective and $\bfva$ as the population meta-objective.

The overall proof strategy is similar as before: we will show the empirical meta-objective is high when the step size is beyond certain threshold, and only prove generalization result for step sizes below this threshold. Under the train-by-validation meta-objective, the optimal step size $\etava$ is in order $\Theta(1/t).$ So we will choose a smaller threshold step size to be $1/L.$


When $\eta < 1/L$, we show that the learned signal is linear in $\eta t$ while the fitted noise is quadratic in $\eta t.$ So there exists certain step size in the order $\Theta(1/t)$ such that our model can leverage the signal in the training set without overfitting the noise. More precisely, we prove the following lemma.  

\begin{lemma}\label{lem:expectation_small_informal}
There exist $\eta_1,\eta_2,\eta_3=\Theta(1/t)$ with $\eta_1<\eta_2<\eta_3$ such that 
\begin{align*}
&\bfva(\eta_2)\leq \frac{1}{2}\ns{w^*}-\frac{9}{10}C+\frac{\sigma^2}{2}\\
&\bfva(\eta)\geq \frac{1}{2}\ns{w^*}-\frac{6}{10}C+\frac{\sigma^2}{2},\forall \eta\in[0,\eta_1]\cup [\eta_3,1/L] 
\end{align*}
where $C$ is a positive constant.
\end{lemma}

We then show whenever $\eta$ is large, either the gradient descent diverges and the sequence gets truncated or it converges and overfits the noise. In both cases, the meta-objective must be high.

\begin{lemma}\label{lem:large_valid_GD_informal}
With probability at least $1-\expm,$
$$\hfva(\eta)\geq C'\sigma^2+\frac{1}{2}\sigma^2,$$
for all $\eta\geq 1/L,$ where $C'$ is a positive constant independent with $\sigma.$
\end{lemma}

To relate the behavior of $\bfva$ to the behavior of $\hfva$, we prove the following uniform convergence result for step sizes in $[0,1/L].$ The proof is similar as in Lemma~\ref{lem:genera_train_GD_informal}.

\begin{lemma}\label{lem:genera_valid_GD_informal}
With probability at least $1-O(1/\epsilon)\expepsm$, 
$$|\hfva(\eta)-\bfva(\eta)|\leq \epsilon,$$
for all $\eta\in[0,1/L].$
\end{lemma}

By choosing a small enough $\epsilon$ in Lemma~\ref{lem:genera_valid_GD_informal}, we ensure that the behavior of $\hfva$ is similar as that of $\bfva$ in Lemma~\ref{lem:expectation_small_informal}. 
Combing with Lemma~\ref{lem:large_valid_GD_informal}, we know $\etava=\Theta(1/t)$ and $\bfva(\etava)\leq \frac12\ns{w^*}+\frac12\sigma^2-\Omega(1).$ 
This concludes our proof since $\bfva(\eta)=\frac12 \E\ns{\wt - w^*}+\frac12\sigma^2.$

\begin{figure}[t]
     \centering
     \includegraphics[width=3in]{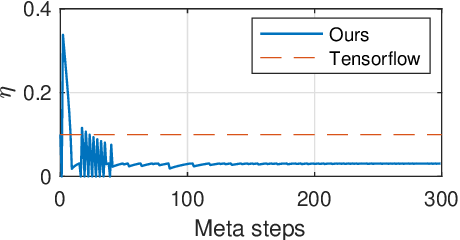}
     \caption{Meta training trajectory for $\eta$ ($t=80$, $\eta_0=0.1$).}
     \label{fig:dim20_t80_init_eta0.1}
\end{figure}

\section{Experiments}
\label{sec:experiments}
In this section, we give experiment results on both synthetic data and realistic data to verify our theory.\footnote{Our code is available at \href{ https://github.com/Kolin96/learning-to-learn}{https://github.com/Kolin96/learning-to-learn}.}






\paragraph{Optimizing step size for quadratic objective}

We first validate the results in Section~\ref{sec:quadratic}. We fixed a 20-dimensional quadratic objective as the inner problem and vary the number of inner steps $t$ and initial value $\eta_0$. We compute the meta-gradient directly using the formula in Eqn.~\eqref{eqn:avoid_gradient_vanishing}. In this way, we avoid the computation of exponentially small/large intermediate terms. We use the algorithm suggested in Theorem~\ref{thm:quadratic_converge}, except we choose the meta-step size to be $1/(100\sqrt{k})$ as the constants in the theorem were not optimized. 

An example training curve of $\eta$ for $t=80$ and $\eta_0=0.1$ is shown in Figure~\ref{fig:dim20_t80_init_eta0.1}, and we can see that $\eta$ converges quickly within 300 steps. 
Similar convergence also holds for larger $t$ or larger initial $\eta_0$. 
In contrast, we also implemented the meta-training with Tensorflow, where the code was adapted from the previous work of \citet{wichrowska2017learned}. Experiments show that in many settings (especially with large $t$ and large $\eta_0$) the implementation does not converge. In Figure~\ref{fig:dim20_t80_init_eta0.1}, under the TensorFlow implementation, the step size is stuck at the initial value throughout the meta training because the meta-gradient explodes and gives NaN value. More details can be found in Appendix \ref{sec:experiment_details}.

\paragraph{Train-by-train vs. train-by-validation, synthetic data}

Here we validate our theoretical results in Section~\ref{sec:least-squares} using the least-squares model defined there. We fix the input dimension $d$ to be $1000$.

\begin{figure}[h]
\begin{minipage}[t]{0.48\linewidth}
     \centering
     \includegraphics[width=2in]{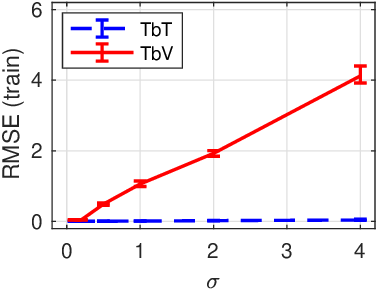}
\end{minipage}
\hfill
\begin{minipage}[t]{0.48\linewidth}
     \centering
     \includegraphics[width=2in]{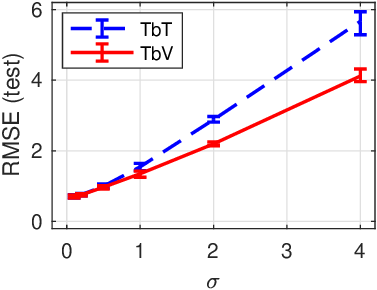}
\end{minipage}
\caption{Training and testing RMSE for different $\sigma$ values (500 samples)}
\label{fig:fixed_samplesize_rmse}
\end{figure}

In the first experiment, we fix the size of the data ($n = 500$ for train-by-train, $n_1 = n_2 = 250$ for train-by-validation).
Under different noise levels, we find the optimal $\eta^*$ by a grid search on its meta-objective for train-by-train and train-by-validation settings respectively. We then use the optimal $\eta^*$ found in each of these two settings to test on 10 new least-squares problem. The mean RMSE, as well as its range over the 10 test cases, are shown in Figure~\ref{fig:fixed_samplesize_rmse}. We can see that for all of these cases, the train-by-train model overfits easily, while the train-by-validation model performs much better and does not overfit. Also, when the noise becomes larger, the difference between these two settings becomes more significant.

\begin{figure}[h]
\begin{minipage}[t]{0.48\linewidth}
     \centering
     \includegraphics[width=2in]{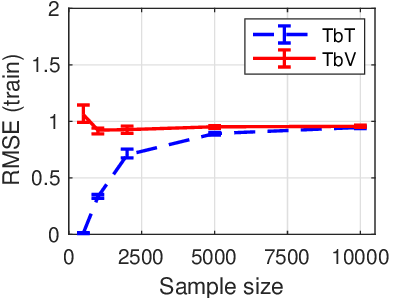}
\end{minipage}
\hfill
\begin{minipage}[t]{0.48\linewidth}
     \centering
     \includegraphics[width=2in]{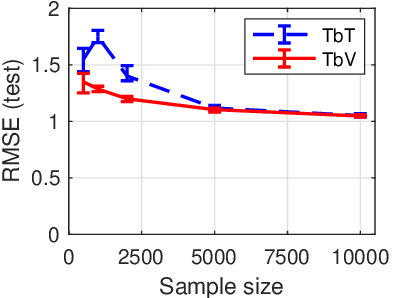}
\end{minipage}
\caption{Training and testing RMSE for different samples sizes ($\sigma=1$)}
\label{fig:fixed_sigma_rmse}
\end{figure}

In the next experiment, we fix $\sigma = 1$ and change the sample size. For train-by-validation, we always split the samples evenly into training and validation set.
From Figure~\ref{fig:fixed_sigma_rmse}, we can see that the gap between these two settings is decreasing as we use more data, as expected by Theorem~\ref{thm:large_sample}. 

\paragraph{Train-by-train vs. train-by-validation, MLP optimizer on MNIST}
Here we consider the more interesting case of a %
multi-layer perceptron (MLP) optimizer on MNIST data set. We use the same MLP optimizer as in~\citet{metz2019understanding}, and details of this optimizer is discussed in Appendix~\ref{sec:experiment_details}. As the inner problem, we use a two-layer fully-connected network of 100 and 20 hidden units with ReLU activations. The inner objective is the classic 10-class cross entropy loss, and we use mini-batches of 32 samples at inner training. In all the following experiments, we use SGD as a baseline with step size tuned by grid search against validation loss. 
For each optimizer, we run 5 independent tests and collect training accuracy and test accuracy for evaluation. The plots show the mean of the 5 tests\footnote{We didn't show the measure of the spread because the results of these 5 tests are so close to each other, such that the range or standard deviation marks will not be readable in the plots.}.

\begin{figure}[H]
\centering
\subfigure[1000 samples, no noise]{
    \includegraphics[width=3in]{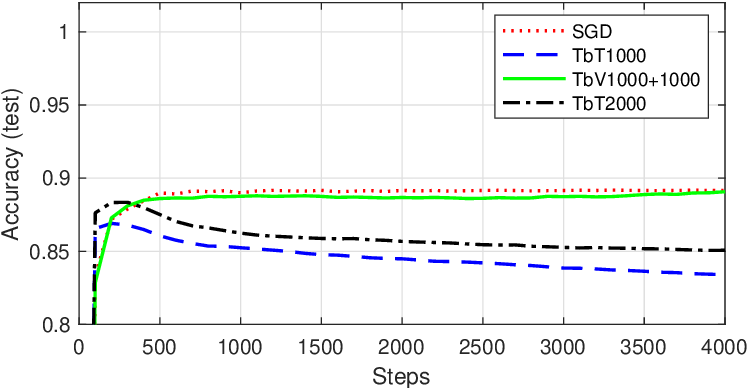}
}
\subfigure[1000 samples, 20\% noise]{
    \includegraphics[width=3in]{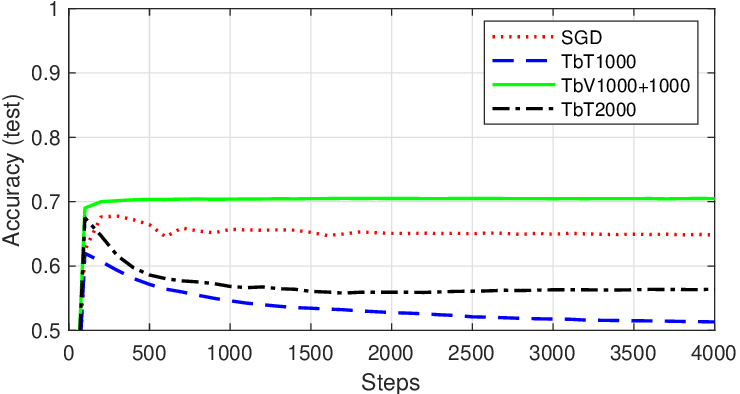}
}
\subfigure[All samples, no noise]{
    \includegraphics[width=3in]{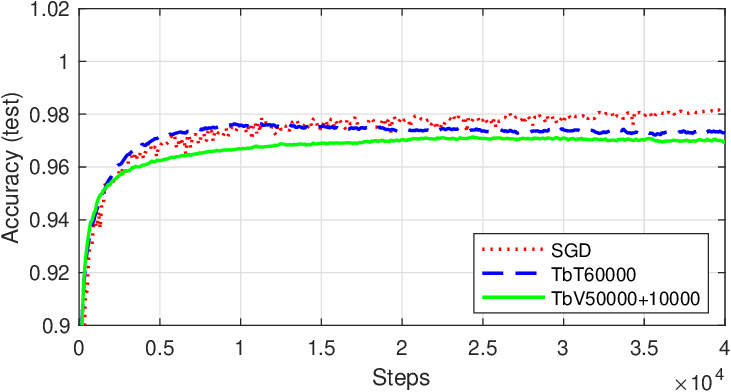} 
}
\caption{The test accuracy of different optimizers in various settings. Comparison between (a) and (b) shows that the advantage of train-by-validation over train-by-train increases when the samples have more noise; comparison between (a) and (c) shows that when the number of samples increases, train-by-train gets comparable performance as train-by-validation.}\label{fig:compare}
\end{figure}

\begin{figure}[h]
    \centering
    \includegraphics[width=3in]{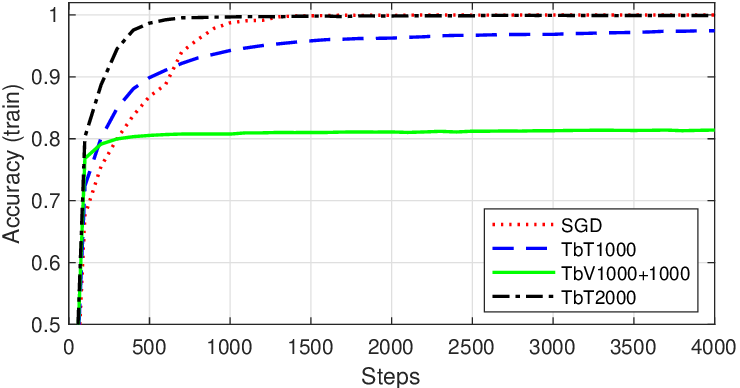}
\caption{Training accuracy for 1000 samples and 20\% noise (same setting as in Figure~\ref{fig:compare}(b))}
\label{fig:mnist_acc_noise_train}
\end{figure}

In Figure~\ref{fig:compare}, we show the test accuracy for different optimizers for different sample size and noise level. In this figure, ``TbT$x$'' represents train-by-train approach with $x$ training samples; ``TbV$x+y$'' represents train-by-validation approach with $x$ training samples and $y$ validation samples. In Figure~\ref{fig:compare}(a) the optimizer is applied to 1000 randomly sub-sampled data (split between training and validation for train-by-validation); in Figure~\ref{fig:compare}(b) we use the same amount of data, except we add 20\% label noise; in Figure~\ref{fig:compare}(c) we use the whole MNIST dataset without label noise. Comparing Figure~\ref{fig:compare}(a) and (b), we see that when the noise is large train-by-validation significantly outperforms train-by-train. Figure~\ref{fig:mnist_acc_noise_train} gives the training accuracy in the same setting as Figure~\ref{fig:compare}(b), which clearly shows that train-by-validation can avoid overfitting to noisy labels. Comparing Figure~\ref{fig:compare}(a) and (c), we see that when the number of samples is large enough there is no significant difference between train-by-train and train-by-validation.

\section{Conclusions}
In this paper, we have proved optimization and generalization guarantees for tuning the step size for quadratic loss. 
From the optimization perspective, we considered a simple task whose objective is a quadratic function. We proved that the meta-gradient can explode/vanish if the meta-objective is simply the loss of the last iteration; we then showed that the log-transformed meta-objective has polynomially bounded meta-gradient and can be successfully optimized. To study the generalization issues, we considered the least squares problem---when the number of samples is small and the noise is large, train-by-validation approach generalizes better than train-by-train; while when the number of samples is large, train-by-train can also work well. 

Although our theoretical results are proved for quadratic loss, this simple setting already yields interesting phenomenons and requires non-trivial techniques to analyze. We have also verified our theoretical results on an optimizer parameterized by neural networks and on MNIST dataset. There are still many open problems, including extending similar analysis to more complicated optimizers, or generalizing the idea to prevent numerical issues to neural network optimizers. We hope our work can lead to more theoretical understanding of the learning-to-learn approach.


\section*{Acknowledgements}
Rong Ge, Xiang Wang and Chenwei Wu are supported in part by NSF Award CCF-1704656, CCF-1845171 (CAREER), CCF-1934964 (Tripods), a Sloan Research Fellowship, and a Google Faculty Research Award. Part of the work was done when Rong Ge and Xiang Wang were visiting Instituted for Advanced Studies for ``Special Year on Optimization, Statistics, and Theoretical Machine Learning'' program. We acknowledge the valuable early discussions with Yatharth Dubey.

\bibliography{ref}
\bibliographystyle{icml2021}

\clearpage
\appendix

In the appendix, we first give the missing proofs for the theorems in the main paper. Later in Appendix~\ref{sec:experiment_details} we give details for the experiments.

\paragraph{Notations:} Besides the notations defined in Section~\ref{sec:prelim}, we define more notations that will be used in the proofs.

For a matrix $X\in\R^{n\times d}$ with $n\leq d,$ we denote its singular values as $\sigma_1(X)\geq \cdots \geq \sigma_n(X).$ 

For a positive semi-definite matrix $A\in \R^{d\times d},$ we denote $u^\top A u$ as $\ns{u}_A.$ For a matrix $X\in\R^{d\times n},$ let $\proj_X\in\R^{d\times d}$ be the projection matrix onto the column span of $X$. That means, $\proj_X = SS^\top$, where the columns of $S$ form an orthonormal basis for the column span of $X.$ 

For any event $\calE,$ we use $\indic{\calE}$ to denote its indicator function: $\indic{\calE}$ equals $1$ when $\calE$ holds and equals $0$ otherwise. We use $\bar{\calE}$ to denote the complementary event of $\calE.$
\section{Proofs for Section~\ref{sec:quadratic} \--- alleviating gradient explosion/vanishing problem for quadratic objective}\label{sec:proofs_quadratic}
In this section, we prove the results in Section~\ref{sec:quadratic}. Recall the meta learning problem as follows:

The inner task is a fixed quadratic problem, where the starting point is fixed at $w_0$, and the loss function is $f(w) = \frac{1}{2} w^\top H w$ for some fixed positive definite matrix $H\in\R^{d\times d}$. Suppose the eigenvalue decomposition of $H$ is $\sum_{i=1}^d \lambda_i u_iu_i^\top.$ In this section, we assume $L = \lambda_1(H)$ and $\alpha = \lambda_d(H)$ are the largest and smallest eigenvalues of $H$ with $L>\alpha$. We assume the starting point $w_0$ has unit $\ell_2$ norm. For each $i\in[d],$ let $c_i$ be $\inner{w_0}{u_i}$ and let $c_{\min}=\min(|c_1|,|c_d|).$ We assume $c_{\min}>0$ for simplicity, which is satisfied if $w_0$ is chosen randomly from the unit sphere.

Let $\{w_{\tau,\eta}\}$ be the GD sequence running on $f(w)$ starting from $w_0$ with step size $\eta.$ For the meta-objective, we consider using the loss of the last point directly, or using the $\log$ of this value. In Section~\ref{sec:gradient_vanishes_expolodes}, we first show that although choosing $\hat{F}(\eta) = f(\wt)$ does not have any bad local optimal solution, it has the gradient explosion/vanishing problem (Theorem~\ref{thm:quadratic_exponential}). Then, in Section~\ref{sec:gradient_vanish_expolode_fix}, we show choosing $\hat{F}(\eta) = \frac{1}{t}\log f(\wt)$ leads to polynomially bounded meta-gradient and further show meta-gradient descent converges to the optimal step size (Theorem~\ref{thm:quadratic_converge}). Although the meta-gradient is polynomially bounded, if we simply use back-propogation to compute the meta-gradient, the intermediate results can still be exponentially large/small (Corollary~\ref{cor:explode_log}). This is also proved in Section~\ref{sec:gradient_vanish_expolode_fix}.

\subsection{Meta-gradient vanishing/explosion}\label{sec:gradient_vanishes_expolodes}
In this section, we show although choosing $\hat{F}(\eta) = f(\wt)$ does not have any bad local optimal solution, it has the meta-gradient explosion/vanishing problem. Recall Theorem~\ref{thm:quadratic_exponential} as follows.

\quadraticExponential*

Intuitively, if we write $w_{t,\eta}$ in the basis of the eigen-decomposition of $H$, then each coordinate evolve exponentially in $t$. The gradient of the standard objective is therefore also exponential in $t$.

\begin{proofof}{Theorem~\ref{thm:quadratic_exponential}}
According to the gradient descent iterations, we have
\begin{align*}
\wt = w_{t-1,\eta}-\eta\nabla f(w_{t-1,\eta})=w_{t-1,\eta}-\eta Hw_{t-1,\eta}=(I-\eta H)w_{t-1,\eta}= (I-\eta H)^t w_0.
\end{align*} 
Therefore, 
$\hat{F}(\eta):=f(\wt)=\frac{1}{2}w_0^\top (I-\eta H)^{2t}Hw_0.$
Taking the derivative of $\hat{F}(\eta),$
\[
\hat{F}'(\eta)= -tw_0^\top (I-\eta H)^{2t-1}H^2w_0= -t\sum_{i=1}^d c_i^2\lambda_i^2(1-\eta\lambda_i)^{2t-1},
\]
where $c_i=\inner{w_0}{u_i}.$
Taking the second derivative of $F(\eta),$
\begin{align*}
F''(\eta)=& t(2t-1)w_0^\top (I-\eta H)^{2t-2}H^3w_0= t(2t-1)\sum_{i=1}^d c_i^2\lambda_i^3(1-\eta\lambda_i)^{2t-2}.
\end{align*}
Since $L>\alpha,$ we have $\hat{F}''(\eta)>0$ for any $\eta$. That means $\hat{F}(\eta)$ is a strictly convex function in $\eta$ with a unique minimizer. 

For any fixed $\eta\in(0,2/L)$ we know $|1-\eta\lambda_i|<1$ for all $i\in[d].$ We have
\begin{align*}
\absr{\hat{F}'(\eta)}\leq& t\sum_{i=1}^d c_i^2\lambda_i^2|1-\eta\lambda_i|^{2t-1}\\
\leq& t\sum_{i=1}^d c_i^2\max_{i\in[d]} \pr{\lambda_i^2|1-\eta\lambda_i|^{2t-1}}\\
\leq& tL^2 \max\pr{|1-\eta\alpha|^{2t-1},|1-\eta L|^{2t-1}}, 
\end{align*} 
where the last inequality uses $\sum_{i=1}^d c_i^2=1.$ Note for $\eta\in(0,2/L),$ it's guaranteed that $|1-\eta\lambda_i|$ takes the maximum at $|1-\eta\alpha|$ or $|1-\eta L|.$

For any fixed $\eta\in(2/L,\infty),$ we know $\eta L-1>1.$ We have 
\begin{align*}
&\hat{F}'(\eta)\\
=& -tc_1^2 L^2(1-\eta L)^{2t-1} -t\sum_{i\neq 1: (1-\eta\lambda_i)\leq 0} c_i^2\lambda_i^2(1-\eta\lambda_i)^{2t-1}-t\sum_{i\neq 1: (1-\eta\lambda_i)> 0} c_i^2\lambda_i^2(1-\eta\lambda_i)^{2t-1}\\
\geq& tc_1^2 L^2(\eta L-1)^{2t-1}-t\sum_{i=1}^d c_i^2\lambda_i^2\geq tc_1^2 L^2(\eta L-1)^{2t-1}-L^2 t,
\end{align*}
where the last inequality uses $\sum_{i=1}^d c_i^2=1.$
\end{proofof}

\subsection{Alleviating meta-gradient vanishing/explosion}\label{sec:gradient_vanish_expolode_fix}

We prove when the the meta objective is chosen as $\frac{1}{t}\log f(\wt)$, the meta-gradient is polynomially bounded. Furthermore, we show meta-gradient descent can converge to the optimal step size within polynomial iterations. Recall Theorem~\ref{thm:quadratic_converge} as follows.

\quadraticConverge*

When we take the $\log$ of the function value, the derivative of the function value with respect to $\eta$ becomes much more stable. We will first show some structural result on $\hat{F}(\eta)$ \--- it has a unqiue minimizer and the gradient is polynomially bounded. Further the gradient is only close to 0 when the point $\eta$ is close to the unique minimizer. Then using such structural result we prove that meta-gradient descent converges.

\begin{proofof}{Theorem~\ref{thm:quadratic_converge}}
The proof consists of three claims. In the first claim, we show that $\hat{F}$ has a unique minimizer and the minus meta derivative always points to the minimizer. In the second claim, we show that $\hat{F}$ has bounded derivative. In the last claim, we show that for any $\eta$ that is outside the $\epsilon$-neighborhood of $\eta^*$, $|\hat{F}'(\eta)|$ is lower bounded. Finally, we combine these three claims to finish the proof.

\begin{claim}\label{clm:minimizer}
The meta objective $\hat{F}$ has only one stationary point that is also its unique minimizer $\eta^*$. For any $\eta\in[0,\eta^*),$ $\hat{F}'(\eta)<0$ and for any $\eta\in(\eta^*,\infty),$ $\hat{F}'(\eta)>0.$ Furthermore, we know $\eta^*\in[1/L, 1/\alpha].$  
\end{claim}
We can compute the derivative of $\hat{F}$ in $\eta$ as follows,
\begin{align}\label{eqn:metagrad_hand}
\hat{F}'(\eta)
= \frac{-2w_0^\top (I-\eta H)^{2t-1}H^2w_0}{w_0^\top (I-\eta H)^{2t}Hw_0}
= \frac{-2\sum_{i=1}^d c_i^2\lambda_i^2(1-\eta\lambda_i)^{2t-1}}{\sum_{i=1}^d c_i^2\lambda_i(1-\eta\lambda_i)^{2t}}.
\end{align} 
It's not hard to verify that the denominator $\sum_{i=1}^d c_i^2\lambda_i(1-\eta\lambda_i)^{2t}$ is always positive. Denote the numerator $-2\sum_{i=1}^d c_i^2\lambda_i^2(1-\eta\lambda_i)^{2t-1}$ as $g(\eta).$ Since $g'(\eta)>0$ for any $\eta\in[0,\infty)$, we know $g(\eta)$ is strictly increasing in $\eta.$ Combing with the fact that $g(0)<0$ and $g(\infty)>0,$ we know there is a unique point (denoted as $\eta^*$) where $g(\eta^*)=0$ and $g(\eta)<0$ for all $\eta\in[0,\eta^*)$ and $g(\eta)>0$ for all $\eta\in(\eta^*,\infty).$ Since the denominator in $\hat{F}'(\eta)$ is always positive and the numerator equals $g(\eta),$ we know there is a unique point $\eta^*$ where $\hat{F}'(\eta^*)=0$ and $\hat{F}'(\eta)<0$ for all $\eta\in[0,\eta^*)$ and $\hat{F}'(\eta)>0$ for all $\eta\in(\eta^*,\infty).$ It's clear that $\eta^*$ is the minimizer of $\hat{F}.$

Also, it's not hard to verify that for any $\eta\in[0,1/L),$ $\hat{F}'(\eta)<0$ and for any $\eta\in(1/\alpha,\infty),$ $\hat{F}'(\eta)>0$. This implies that $\eta^*\in[1/L, 1/\alpha].$

\begin{claim}\label{clm:max}
For any $\eta\in[0,\infty),$ we have 
\[ |\hat{F}'(\eta)|\leq \frac{4L^3}{c_{\min}^2 \alpha(L-\alpha)}:=D_{\max}.\]
\end{claim}
For any $\eta\in[0,\frac{2}{\alpha+L}],$ we have $|1-\eta\lambda_i|\leq 1-\eta\alpha$ for all $i.$ Dividing the numerator and denominator in $\hat{F'}(\eta)$ by $(1-\eta\alpha)^{2t}$, we have
\begin{align*}
\absr{\hat{F}'(\eta)}
= 2\frac{\absr{\sum_{i=1}^{d} \frac{c_i^2\lambda_i^2}{1-\eta\alpha}(\frac{1-\eta\lambda_i}{1-\eta\alpha})^{2t-1}}}{c_d^2\alpha + \sum_{i=1}^{d-1} c_i^2\lambda_i(\frac{1-\eta\lambda_i}{1-\eta\alpha})^{2t}}
\leq \frac{2\sum_{i=1}^d c_i^2\lambda_i^2 }{c_d^2\alpha(1-\eta\alpha)}\leq \frac{2(\alpha+L)\sum_{i=1}^d c_i^2\lambda_i^2 }{c_d^2\alpha(L-\alpha)}\leq \frac{4L^3 }{c_d^2\alpha(L-\alpha)},
\end{align*}
where the second last inequality uses $\eta\leq \frac{2}{\alpha+L}.$

Similarly for any $\eta\in(\frac{2}{\alpha+L},\infty),$ we have $|1-\eta\lambda_i|\leq \eta L-1$ for all $i.$ Dividing the numerator and denominator in $\hat{F'}(\eta)$ by $(\eta L-1)^{2t}$, we have
\begin{align*}
\hat{F}'(\eta)
= 2\frac{\absr{\sum_{i=1}^{d} \frac{c_i^2\lambda_i^2}{\eta L-1}(\frac{1-\eta\lambda_i}{\eta L-1})^{2t-1}}}{c_1^2 L + \sum_{i=2}^{d} c_i^2\lambda_i(\frac{1-\eta\lambda_i}{\eta L-1})^{2t}}
\leq \frac{2\sum_{i=1}^d c_i^2\lambda_i^2 }{c_1^2 L(\eta L-1)}\leq \frac{2(\alpha+L)\sum_{i=1}^d c_i^2\lambda_i^2 }{c_1^2 L(L-\alpha)}\leq \frac{4L^3}{c_1^2 L(L-\alpha)}
\end{align*}
where the last inequality uses $\eta\geq \frac{2}{\alpha+L}.$

Overall, we know for any $\eta\geq 0,$
\[ |\hat{F}'(\eta)|\leq \frac{4L^3}{L-\alpha}\max\pr{\frac{1}{c_d^2 \alpha},\frac{1}{c_1^2 L}}\leq \frac{4L^3}{c_{\min}^2 \alpha(L-\alpha)}.\]

\begin{claim}\label{clm:min}
Given $\hat{M}\geq 2/\alpha$ and $1/L>\epsilon>0,$ for any $\eta\in[0,\eta^*-\epsilon]\cup[\eta^*+\epsilon,\hat{M}],$ we have
\[|F'(\eta)|\geq \min\pr{\frac{2\epsilon c_d^2\alpha^3}{L} ,\frac{2\epsilon c_1^2 L^2}{(\hat{M} L-1)^2}}
\geq 2\epsilon c_{\min}^2\min\pr{\frac{\alpha^3}{L} ,\frac{1}{\hat{M}^2 }}
:=D_{\min}(\hat{M}).\]
\end{claim}

If $\eta\in[0,\eta^*-\epsilon]$ and $\eta\leq \frac{2}{\alpha+L},$ we have 
\begin{align*}
\hat{F}'(\eta)
= -2\frac{\sum_{i=1}^d c_i^2\lambda_i^2(1-\eta\lambda_i)^{2t-1}}{\sum_{i=1}^d c_i^2\lambda_i(1-\eta\lambda_i)^{2t}}
= -2\frac{\sum_{i=1}^d c_i^2\lambda_i^2(1-\eta\lambda_i)^{2t-1}-\sum_{i=1}^d c_i^2\lambda_i^2(1-\eta^*\lambda_i)^{2t-1} }{\sum_{i=1}^d c_i^2\lambda_i(1-\eta\lambda_i)^{2t}},
\end{align*} 
where the second equality holds because $\sum_{i=1}^d c_i^2\lambda_i^2(1-\eta^*\lambda_i)^{2t-1}=0.$ For the numerator, we have
\begin{align*}
\sum_{i=1}^d c_i^2\lambda_i^2(1-\eta\lambda_i)^{2t-1}-\sum_{i=1}^d c_i^2\lambda_i^2(1-\eta^*\lambda_i)^{2t-1}
\geq& c_d^2\alpha^2\pr{(1-\eta\alpha)^{2t-1}-(1-\eta^*\alpha)^{2t-1}}\\
\geq& c_d^2\alpha^2\pr{(1-\eta\alpha)^{2t-1}-(1-\eta\alpha-\epsilon\alpha)^{2t-1}};
\end{align*}
for the denominator, we have 
\begin{align*}
\sum_{i=1}^d c_i^2\lambda_i(1-\eta\lambda_i)^{2t}\leq \pr{\sum_{i=1}^d c_i^2\lambda_i}(1-\eta\alpha)^{2t},
\end{align*}
where the second inequality holds because $|1-\eta\lambda_i|\leq 1-\eta\alpha$ for all $i.$
Overall, we have when $\eta\in[0,\eta^*-\epsilon]$ and $\eta\leq \frac{2}{\alpha+L},$
\begin{align*}
\absr{\hat{F}'(\eta)}
\geq& 2\frac{c_d^2\alpha^2\pr{(1-\eta\alpha)^{2t-1}-(1-\eta\alpha-\epsilon\alpha)^{2t-1}}}{\pr{\sum_{i=1}^d c_i^2\lambda_i}(1-\eta\alpha)^{2t}}\\
\geq& \frac{2\epsilon c_d^2\alpha^3}{\pr{\sum_{i=1}^d c_i^2\lambda_i}(1-\eta\alpha)}
\geq \frac{2\epsilon c_d^2\alpha^3}{L},
\end{align*}
where the last inequality holds because $(1-\eta\alpha)\leq 1$ and $\sum_i^d c_i^2 \lambda_i\leq L.$

Similarly, if $\eta\in[0,\eta^*-\epsilon]$ and $\eta\geq \frac{2}{\alpha+L},$ we have 
\begin{align*}
\absr{\hat{F}'(\eta)}
\geq& 2\frac{c_1^2 L^2\pr{(1-\eta L)^{2t-1}-(1-\eta L-\epsilon L)^{2t-1}}}{\pr{\sum_{i=1}^d c_i^2\lambda_i}(1-\eta L)^{2t}}\\
=& 2\frac{c_1^2 L^2\pr{(\eta L+\epsilon L-1)^{2t-1}-(\eta L-1)^{2t-1}}}{\pr{\sum_{i=1}^d c_i^2\lambda_i}(\eta L-1)^{2t}}\\
\geq& \frac{2\epsilon c_1^2 L^3}{\pr{\sum_{i=1}^d c_i^2\lambda_i}(\eta L-1)^2}
\geq \frac{2\epsilon c_1^2 \alpha^2 L^2}{(L-\alpha)^2},
\end{align*}
where the last inequality holds because $\eta\leq \eta^*-\epsilon\leq 1/\alpha$ and $\sum_i^d c_i^2 \lambda_i\leq L.$

If $\eta\in[\eta^*+\epsilon,\infty)$ and $\eta\leq \frac{2}{\alpha+L},$ we have 
\begin{align*}
\absr{\hat{F}'(\eta)}
\geq& 2\frac{c_d^2\alpha^2\pr{(1-\eta\alpha+\epsilon\alpha)^{2t-1}-(1-\eta\alpha)^{2t-1}}}{\pr{\sum_{i=1}^d c_i^2\lambda_i}(1-\eta\alpha)^{2t}}\\
\geq& \frac{2\epsilon c_d^2\alpha^3}{L},
\end{align*}

If $\eta\in[\eta^*+\epsilon,\infty)$ and $\eta\geq \frac{2}{\alpha+L},$ we have 
\begin{align*}
\absr{\hat{F}'(\eta)}
\geq& 2\frac{c_1^2 L^2\pr{(1-\eta L+\eta\epsilon)^{2t-1}-(1-\eta L)^{2t-1}}}{\pr{\sum_{i=1}^d c_i^2\lambda_i}(1-\eta L)^{2t}}\\
\geq& \frac{2\epsilon c_1^2 L^3}{\pr{\sum_{i=1}^d c_i^2\lambda_i}(\eta L-1)^2}
\geq \frac{2\epsilon c_1^2 L^2}{(\hat{M} L-1)^2},
\end{align*}
where the last inequality uses the assumption that $\eta\leq \hat{M}.$

With the above three claims, we are ready to prove the optimization result. By Claim~\ref{clm:minimizer}, we know $F'(\eta)<0$ for any $\eta\in[0,\eta^*)$ and $F'(\eta)>0$ for any $\eta\in(\eta^*,\infty).$ So the opposite gradient descent always points to the minimizer. 

Since $\mu_k=1/\sqrt{k},$ when $k\geq k_1:=\frac{D_{\max}^2}{\epsilon^2}$ we know $\mu_k \leq \frac{\epsilon}{D_{\max}}.$ By Claim~\ref{clm:max}, we know $|\hat{F}'(\eta)|\leq D_{\max}$ for all $\eta\geq 0$, which implies $|\mu_k \hat{F}'(\eta)|\leq \epsilon$ for all $k\geq k_1.$ That means, meta gradient descent will never overshoot the minimizer by more than $\epsilon$ when $k\geq k_1.$ In other words, after $k_1$ meta iterations, once $\eta$ enters the $\epsilon$-neighborhood of $\eta^*$, it will never leave this neighborhood.

We also know that at meta iteration $k_1,$ we have $\eta_{k_1}\leq \max(1/\alpha+D_{\max},M):=\hat{M}.$ Here, $1/\alpha+D_{\max}$ comes from the case that the eta starts from the left of $\eta^*$ and overshoot to the right of $\eta^*$ by $D_{\max}.$ Since $\eta^*\in[1/L,1/\alpha],$ we have $|\eta_{k_1}-\eta^*|\leq \max(1/\alpha,1/\alpha+D_{\max}-1/L,M-1/L):=R.$ By Claim~\ref{clm:min}, we know that $|\hat{F}'(\eta)|\geq D_{\min}(\hat{M})$ for any $\eta\in[0,\eta^*-\epsilon]\cup[\eta^*+\epsilon, \hat{M}].$ 
Choosing some $k_2$ satisfying $\sum_{k=k_1}^{k_2}1/\sqrt{k}\geq \frac{R}{D_{\min}},$ we know for any $k\geq k_2,$ $|\eta_k-\eta^*|\leq \epsilon.$ Plugging in all the bounds for $D_{\min},D_{\max}$ from Claim~\ref{clm:min} and Claim~\ref{clm:max}, we know there exists $k_1=\frac{1}{\epsilon^2}\poly(\frac{1}{c_{\min}},L,\frac{1}{\alpha},\frac{1}{L-\alpha}), k_2=\frac{M^6}{\epsilon^2}\poly(\frac{1}{c_{\min}},L,\frac{1}{\alpha},\frac{1}{L-\alpha})$ satisfying these conditions.
\end{proofof}

Next, we show although the meta-gradient is polynomailly bounded, the intermediate results can still vanish or explode if we use back-propogation to compute the meta-gradient.

\expolodeLog*

\begin{proofof}{Corollary~\ref{cor:explode_log}}
This is done by direct calculation. If we use back-propagation to compute the derivative of $\frac{1}{t}\log(f(\wt)),$ we need to first compute $\frac{\partial f(\wt)}{\partial}\frac{1}{t}\log(f(\wt))$ that equals $\frac{1}{tf(\wt)}$. Same as the analysis in Theorem~\ref{thm:quadratic_exponential}, we can show $\frac{1}{tf(\wt)}$ is exponentially large when $\eta<2/L$ and is exponentially small when $\eta>2/L.$
\end{proofof}



\section{Proofs of train-by-train v.s. train-by-validation (GD)}\label{sec:proofs_GD}
In this section, we show when the number of samples is small and when the noise level is a large constant, train-by-train overfits to the noise in training tasks while train-by-validation generalizes well. We separately prove the results for train-by-train and train-by-validation in Theorem~\ref{thm:train_GD} and Theorem~\ref{thm:valid_GD}, respectively. Then, Theorem~\ref{thm:train_valid_GD} is simply a combination of Theorem~\ref{thm:train_GD} and Theorem~\ref{thm:valid_GD}.

Recall that in the train-by-train setting, each task $P$ contains a training set $\str$ with $n$ samples. The inner objective is defined as 
$\hat{f}(w)=\frac{1}{2n}\sum_{(x,y)\in\str}\pr{\inner{w}{x}-y}^2.$
Let $\{w_{\tau,\eta}\}$ be the GD sequence running on $\hat{f}(w)$ from initialization $0$ (with truncation).  The meta-loss on task $P$ is defined as the inner objective of the last point, $\Delta_{TbT(n)}(\eta,P)=\hat{f}(\wt)=\frac{1}{2n}\sum_{(x,y)\in\str}\pr{\inner{\wt}{x}-y}^2.$ The empirical meta objective $\hat{F}_{TbT(n)}(\eta)$ is the average of the meta-loss across $m$ different tasks. We show that under $\hat{F}_{TbT(n)}(\eta)$, the optimal step size is a constant and the learned weight is far from ground truth $w^*$ on new tasks. We prove Theorem~\ref{thm:train_GD} in Section~\ref{sec:proofs_train_GD}.

\begin{restatable}
{theorem}{thmTrainGD}\label{thm:train_GD}
Let the meta objective $\hat{F}_{TbT(n)}(\eta)$ be as defined in Equation~\ref{eqn:train_GD} with $n\in[d/4,3d/4].$ Assume noise level $\sigma$ is a large constant $c_1$. Assume unroll length $t\geq c_2$, number of training tasks $m\geq c_3\log(mt)$ and dimension $d\geq c_4\log(m)$ for certain constants $c_2,c_3,c_4.$ With probability at least $0.99$ in the sampling of the training tasks, we have 
\begin{align*}
\etatr=\Theta(1) \text{ and } \E\ns{w_{t,\etatr}-w^*}=\Omega(1)\sigma^2,
\end{align*}
for all $\etatr\in\arg\min_{\eta\geq 0} \hat{F}_{TbT(n)}(\eta),$ where the expectation is taken over new tasks.
\end{restatable}

In Theorem~\ref{thm:train_GD}, $\Omega(1)$ is an absolute constant independent with $\sigma.$
Intuitively, the reason that train-by-train performs badly in this setting is because there is a way to set the step size to a constant such that gradient descent converges very quickly to the empirical risk minimizer, therefore making the train-by-train objective very small. However, when the noise is large and the number of samples is smaller than the dimension, the empirical risk minimizer (ERM) overfits to the noise and is not the best solution.

In the train-by-validation setting, each task $P$ contains a training set $\str$ with $n_1$ samples and a validation set with $n_2$ samples. The inner objective is defined as 
$\hat{f}(w)=\frac{1}{2n_1}\sum_{(x,y)\in\str}\pr{\inner{w}{x}-y}^2.$ Let $\{w_{\tau,\eta}\}$ be the GD sequence running on $\hat{f}(w)$ from initialization $0$ (with truncation). For each task $P$, the meta-loss $\Delta_{TbV(n_1,n_2)}(\eta,P)$ is defined as the loss of the last point $\wt$ evaluated on the validation set $\sva.$ That is, $\Delta_{TbV(n_1,n_2)}(\eta,P)=\frac{1}{2n_2}\sum_{(x,y)\in\sva}\pr{\inner{\wt}{x}-y}^2.$ The empirical meta objective $\hat{F}_{TbV(n_1,n_2)}(\eta)$ is the average of the meta-loss across $m$ different tasks $P_1,P_2,...,P_m$. We show that under $\hat{F}_{TbV(n_1,n_2)}(\eta)$, the optimal step size is $\Theta(1/t)$ and the learned weight is better than initialization $0$ by a constant on new tasks. Theorem~\ref{thm:valid_GD} is proved in Section~\ref{sec:proofs_valid_GD}.  

\begin{restatable}
{theorem}{thmValidGD}\label{thm:valid_GD}
Let the meta objective $\hat{F}_{TbV(n_1,n_2)}(\eta)$ be as defined in Equation~\ref{eqn:valid_GD} with $n_1,n_2\in[d/4,3d/4]$. Assume noise level $\sigma$ is a large constant $c_1$. Assume unroll length $t\geq c_2$, number of training tasks $m\geq c_3$ and dimension $d\geq c_4\log(t)$ for certain constants $c_2,c_3,c_4.$ With probability at least $0.99$ in the sampling of training tasks, we have
\[ \etava=\Theta(1/t)\mbox{ and }\E\ns{w_{t,\etava}-w^*}=\ns{w^*}-\Omega(1)\]
for all $\etava\in \arg\min_{\eta\geq 0}\hat{F}_{TbV(n_1,n_2)}(\eta),$ where the expectation is taken over new tasks.
\end{restatable}

Intuitively, train-by-validation is optimizing the right objective. As long as the meta-training problem has good generalization performance (that is, good performance on a few tasks implies good performance on the distribution of tasks), then train-by-validation should be able to choose the optimal learning rate. The step size of $\Theta(1/t)$ here serves as regularization similar to early-stopping, which allows gradient descent algorithm to achieve better error on test data.

\paragraph{Notations}
We define more quantities that are useful in the analysis. In the train by train setting, given a task $P_k:=(\calD(w^*_k),\strk,\ell).$ The training set $\strk$ contains $n$ samples $\{x_i^{(k)}, y_i^{(k)}\}_{i=1}^n$ with $y_i^{(k)}=\inner{w^*_k}{x_i^{(k)}}+\xi_i^{(k)}.$

Let $\xtrk$ be an $n \times d$ matrix with its $i$-th row as $(x_i^{(k)})^\top$. Let $\htrk:=\frac{1}{n} (\xtrk)^\top \xtrk$ be the covariance matrix of the inputs in $\strk.$ Let $\xitrk$ be an $n$-dimensional column vector with its $i$-th entry equal to $\xi_i^{(k)}$.

Since $n\leq d,$ with probability $1$, we know $\xtrk$ is full row rank. Therefore, $\xtrk$ has pseudo-inverse $(\xtrk)^\dagger$ such that $\xtrk(\xtrk)^\dagger=I_{n}.$ It's not hard to verify that there exists $\wtrk = \proj_{(\xtrk)^\top}w^*_k + (\xtrk)^{\dagger}\xitrk$ such that $y_i^{(k)} = \inner{\wtrk}{x_i^{(k)}}$ for every $(x_i^{(k)},y_i^{(k)})\in\strk.$ 
Here, $\proj_{(\xtrk)^\top}$ is the projection matrix onto the column span of $(\xtrk)^\top$. We also denote $\proj_{(\xtrk)^\top}w^*_k$ as $(\wtrk)^*$. We use $\btk$ to denote $(I-(I-\eta\htrk)^t).$ Let $\wtk$ be the weight obtained by running GD on $\strk$ with step size $\eta$ (with truncation).

With the above notations, it's not hard to verify that for task $P_k$, the inner objective $\hat{f}(w)=\frac{1}{2}\ns{w-\wtrk}_{\htrk}.$ The meta-loss on task $P_k$ is just $\Delta_{TbT(n)}(\eta,P_k)=\frac{1}{2}\ns{\wt-\wtrk}_{\htrk}.$

In the train-by-validation setting, each task $P_k$ contains a training set $\strk$ with $n_1$ samples and a validation set $\svak$ with $n_2$ samples. Similar as above, for the training set $\strk,$ we can define $\xitrk,\xtrk,\htrk,\wtrk,\btk,\wtk;$ for the validation set $\svak,$ we can define $\xivak,\xvak,\hvak,\wvak.$ With these notations, the inner objective is $\hat{f}(w)=\frac{1}{2}\ns{w-\wtrk}_{\htrk}$ and the meta-loss is $\Delta_{TbV(n_1,n_2)}(\eta,P_k)=\frac{1}{2}\ns{\wt-\wvak}_{\hvak}.$

We also use these notations without index $k$ to refer to the quantities defined on task $P.$ In the proofs, we ignore the subsripts on $n,n_1,n_2$ and simply write $\dtrk,\dvak,\hftr,\hfva,\bftr,\bfva.$

\subsection{Overall Proof Strategy}\label{sec:strategy}

In this section (and the next), we follow similar proof strategies that consists of three steps. 

\paragraph{Step 1:} First, we show for both train-by-train and train-by-validation, there is a good step size that achieves small empirical meta-objective (however the step sizes and the empirical meta-objective they achieve are different in the two settings). This does not necessarily mean that the actual optimal step size is exactly the good step size that we propose, but it gives an upperbound on the empirical meta-objective for the optimal step size.

\paragraph{Step 2:} Second, we define a threshold step size such that for any step size larger than it, the empirical meta-objective must be higher than what was achieved at the good step size in Step 1. This immediately implies that the optimal step size cannot exceed this threshold step size.

\paragraph{Step 3:} Third, we show the meta-learning problem has good generalization performance, that is, if a learning rate $\eta$ performs well on the training tasks, it must also perform well on the task distribution, and vice versa. Thanks to Step 1 and Step 2, we know the optimal step size cannot exceed certain threshold and then only need to prove generalization result within this range. The generalization result is not surprising as we only have a single trainable parameter $\eta$, however we also emphasize that this is non-trivial as we will not restrict the step size $\eta$ to be small enough that the algorithms do not diverge. Instead we use a truncation to alleviate the diverging problem (this allows us to run the algorithm on distribution of data whose largest possible learning rate is unknown). 

Combing Step 1, 2, 3, we know the population meta-objective has to be small at the optimal step size. Finally, we show that as long as the population meta-objective is small, the performance of the algorithms satisfy what we stated in Theorem~\ref{thm:train_valid_GD}. The last step is easier for the train-by-validation setting, because its meta-objective is exactly the correct measure that we are looking at; for the train-by-train setting we instead look at the property of empirical risk minimizer (ERM), and show that anything close to the ERM is going to behave similarly. 

\subsection{Train-by-train (GD)}\label{sec:proofs_train_GD}
Recall Theorem~\ref{thm:train_GD} as follows. 

\thmTrainGD*

According to the data distribution, we know $\xtr$ is an $n\times d$ random matrix with each entry i.i.d. sampled from standard Gaussian distribution. In the following lemma, we show that the covariance matrix $\htr$ is approximately isotropic when $d/4\leq n\leq 3d/4$. Specifically, we show $\frac{\sqrt{d}}{\sqrt{L}} \leq \sigma_i(\xtr) \leq \sqrt{Ld} \text{  and  } \frac{1}{L} \leq \lambda_i(\htr) \leq L$ for all $i\in[n]$ with $L=100.$ We use letter $L$ to denote the upper bound of $\n{\htr}$ to emphasize that this bounds the smoothness of the inner objective. Throughout this section, we use letter $L$ to denote constant $100.$ The proof of Lemma~\ref{lem:isotropic} follows from random matrix theory. We defer its proof into Section~\ref{sec:technical_train_GD}. 

\begin{lemma}\label{lem:isotropic}
Let $X\in\R^{n\times d}$ be a random matrix with each entry i.i.d. sampled from standard Gaussian distribution. Let $H:=1/nX^\top X.$ Assume $n=cd$ with $c\in[\frac{1}{4},\frac{3}{4}].$ Then, with probability at least $1-\expd,$ there exists constant $L=100$ such that 
\[\frac{\sqrt{d}}{\sqrt{L}} \leq \sigma_i(X) \leq \sqrt{Ld} \text{  and  } \frac{1}{L} \leq \lambda_i(H) \leq L,\]
for all $i\in[n].$
\end{lemma}

In this section, we always assume the size of each training set is within $[d/4,3d/4]$ so Lemma~\ref{lem:isotropic} holds. Since $\n{\htr}$ is upper bounded by $L$ with high probability, we know the GD sequence converges to $\wtr$ for $\eta\in[0,1/L].$ In Lemma~\ref{lem:small_step_train}, we prove that the empirical meta objective $\hftr$ monotonically decreases as $\eta$ increases until $1/L.$ Also, we show $\hftr$ is exponentially small in $t$ at step size $1/L$. This serves as step 1 in Section~\ref{sec:strategy}.
The proof is deferred into Section~\ref{sec:small_train_GD}.

\smallStepTrain*

When the step size is larger than $1/L,$ the GD sequence can diverge, which incurs a high loss in meta objective. Later in Definition~\ref{def:teta}, we define a step size $\teta$ such that the GD sequence gets truncated with descent probability for any step size that is larger than $\teta.$ In Lemma~\ref{lem:diverge_train_GD}, we show with high probability, the empirical meta objective is high for all $\eta>\teta.$ This serves as step 2 in the proof strategy described in Section~\ref{sec:strategy}.
The proof is deferred into Section~\ref{sec:large_train_GD}.

\divergeTrainGD*
By Lemma~\ref{lem:small_step_train} and Lemma~\ref{lem:diverge_train_GD}, we know the optimal step size must lie in $[1/L,\teta].$ We can also show $1/L< \teta< 3L,$ so $\etatr$ is a constant. To relate the empirical loss at $\etatr$ to the population loss. We prove a generalization result for step sizes within $[1/L,\teta].$ The following lemma is a formal version of Lemma~\ref{lem:genera_train_GD_informal}.
This serves as step 3 in Section~\ref{sec:strategy}. 
The proof is deferred into Section~\ref{sec:large_train_GD2}.

\begin{restatable}
{lemma}{generaTrainGD}\label{lem:genera_train_GD}
Suppose $\sigma$ is a large constant $c_1$. Assume $t\geq c_2,d\geq c_4$ for certain constants $c_2,c_4.$ With probability at least $1-m\expd-O(t+m)\expm,$
\[|\bftr(\eta)-\hftr(\eta)| \leq \frac{\sigma^2}{L^3},\]
for all $\eta\in[1/L,\teta],$
\end{restatable}

Combining the above lemmas, we know the population meta objective $\bftr$ is small at $\etatr,$ which means $w_{t,\etatr}$ is close to the ERM solution. Since the ERM solution overfits to the noise in training tasks, we know $\n{w_{t,\etatr}-w^*}$ has to be large. 
We present the proof of Theorem~\ref{thm:train_GD} as follows.

\begin{proofof}{Theorem~\ref{thm:train_GD}}
We assume $\sigma$ is a large constant in this proof.
According to Lemma~\ref{lem:small_step_train}, we know with probability at least $1-m\expd,$ $\hftr(\eta)$ is monotonically decreasing in $[0,1/L]$ and $\hftr(1/L)\leq 2L^2\sigma^2(1-1/L^2)^t.$ This implies that the optimal step size $\etatr\geq 1/L$ and $\hftr(\etatr)\leq 2L^2\sigma^2(1-1/L^2)^t.$ By Lemma~\ref{lem:diverge_train_GD}, we know with probability at least $1-\expm,$ $\hftr(\eta)\geq \frac{\sigma^2}{10L^8}$ for all $\eta>\teta,$ where $\teta$ is defined in Definition~\ref{def:teta}. As long as $t\geq c_2$ for certain constant $c_2,$ we know $\frac{\sigma^2}{10L^8}>2L^2\sigma^2(1-1/L^2)^t,$ which then implies that the optimal step size $\etatr$ lies in $[1/L,\teta].$ According to Lemma~\ref{lem:range_teta}, we know $\teta\in(1/L,3L).$ Therefore $\etatr$ is a constant.

According to Lemma~\ref{lem:genera_train_GD}, we know with probability at least $1-m\expd-O(t+m)\expm,$ $|\bftr(\eta)-\hftr(\eta)| \leq \frac{\sigma^2}{L^3},$ for all $\eta\in[1/L,\teta].$ As long as $t$ is larger than some constant, we have $\hftr(\etatr)\leq \frac{\sigma^2}{L^3}.$ Combing with the generalization result, we have $\bftr(\etatr)\leq \frac{2\sigma^2}{L^3}.$ Next, we show that under a small population loss, $\Estr\ns{w_{t,\etatr}-w^*}$ has to be large.

Let $\calE_1$ be the event that $\sqrt{d}/\sqrt{L} \leq \sigma_i(\xtr)\leq  \sqrt{Ld}$ and $1/L \leq \lambda_i(\htr) \leq L$ for all $i\in[n]$ and $\sqrt{d}\sigma/4 \leq \n{\xitr}\leq \sqrt{d}\sigma.$ We have
\begin{align*}
\Estr \ns{w_{t,\etatr}-\wtr}_{\htr}
\geq& \frac{1}{L}\Estr \ns{w_{t,\etatr}-\wtr}\indic{\calE_1}\\
\geq& \frac{1}{L} \pr{\Estr\n{w_{t,\etatr}-\wtr^*-(\xtr)^\dagger\xitr}\indic{\calE_1}}^2\\
\geq& \frac{1}{L} \pr{\Estr\n{(\xtr)^\dagger\xitr}\indic{\calE_1}-\Estr\n{w_{t,\etatr}-\wtr^*}\indic{\calE_1} }^2.
\end{align*}
Since $\Estr \ns{w_{t,\etatr}-\wtr}_{\htr}\leq \frac{4\sigma^2}{L^3},$ this then implies 
\[\Estr\n{(\xtr)^\dagger\xitr}\indic{\calE_1}-\Estr\n{w_{t,\etatr}-\wtr^*}\indic{\calE_1}\leq \sqrt{L\frac{4\sigma^2}{L^3}}=\frac{2\sigma}{L}.\]
Conditioning on $\calE_1,$ we can lower bound $\n{(\xtr)^\dagger\xitr}$ by $\frac{\sigma}{4\sqrt{L}}.$ According to Lemma~\ref{lem:isotropic} and Lemma~\ref{lem:norm_vector}, we know $\Pr[\calE_1]\geq 1-\expd.$ As long as $d$ is at least certain constant, we have $\Pr[\calE_1]\geq 0.9.$ This then implies $\Estr\n{(\xtr)^\dagger\xitr}\indic{\calE_1}\geq \frac{9\sigma}{40\sqrt{L}}.$ Therefore, we have 
\begin{align*}
\Estr\n{w_{t,\etatr}-\wtr^*}\indic{\calE_1}\geq \frac{9\sigma}{40\sqrt{L}}-\frac{2\sigma}{L}=\frac{9\sigma}{4L}-\frac{2\sigma}{L}=\frac{\sigma}{4L},
\end{align*} 
where the first equality uses $L=100.$ Then, we have 
\begin{align*}
\Estr\ns{w_{t,\etatr}-w^*}\geq \Estr\ns{w_{t,\etatr}-\wtr^*}\indic{\calE_1}\geq \pr{\Estr\n{w_{t,\etatr}-\wtr^*}\indic{\calE_1}}^2\geq \frac{\sigma^2}{16L^2},
\end{align*}
where the first inequality holds because for any $\str,$ $\wtr^*$ is the projection of $w^*$ on the subspace of $\str$ and $w_{t,\etatr}$ is also in this subspace. Taking a union bound for all the bad events, we know this result holds with probability at least $0.99$ as long as $\sigma$ is a large constant $c_1$ and $t\geq c_2,m\geq c_3\log(mt)$ and $d\geq c_4\log(m)$ for certain constants $c_2,c_3,c_4.$
\end{proofof}

\subsubsection{Behavior of $\hftr$ for $\eta\in[0,1/L]$}\label{sec:small_train_GD}
In this section, we prove the empirical meta objective $\hftr$ is monotonically decreasing in $[0,1/L]$. Furthermore, we show $\hftr(1/L)$ is exponentially small in $t$. 

\smallStepTrain*

\begin{proofof}{Lemma~\ref{lem:small_step_train}}
For each $k\in[m],$ let $\calE_k$ be the event that $\sqrt{d}/\sqrt{L} \leq \sigma_i(\xtr)\leq  \sqrt{Ld}$ and $1/L \leq \lambda_i(\htr) \leq L$ for all $i\in[n]$ and $\sqrt{d}\sigma/4 \leq \n{\xitr}\leq \sqrt{d}\sigma$. Here, $L$ is constant $100$ from Lemma~\ref{lem:isotropic}. According to Lemma~\ref{lem:isotropic} and Lemma~\ref{lem:norm_vector}, we know for each $k\in[m],$ $\calE_k$ happens with probability at least $1-\expd.$ Taking a union bound over all $k\in[m],$ we know $\cap_{k\in[m]}\calE_k$ holds with probability at least $1-m\expd.$ From now on, we assume $\cap_{k\in[m]}\calE_k$ holds.

Let's first consider each individual loss function $\dtrk$. 
Let $\{\hat{w}_{\tau,\eta}^{(k)}\}$ be the GD sequence without truncation. We have
\begin{align*}
\hat{w}_{\tau,\eta}^{(k)}-\wtrk 
=& \hat{w}_{\tau-1,\eta}^{(k)}-\wtrk-\eta\htrk(\hat{w}_{\tau-1,\eta}^{(k)}-\wtrk)\\
=& (I-\eta\htrk)(\hat{w}_{\tau-1,\eta}^{(k)}-\wtrk)=-(I-\eta\htrk)^t\wtrk.
\end{align*}
For any $\eta\in[0,1/L],$ we have $\n{\hat{w}_{\tau,\eta}^{(k)}}\leq \n{\wtrk}=\n{(\wtrk)^*+(\xtrk)^\dagger\xitrk}\leq 2\sqrt{L}\sigma$ for any $\tau.$ 
Therefore, $\n{\wtk}$ never exceeds the norm threshold and never gets truncated. 

Noticing that $\dtrk=\frac{1}{2}(\wtk-\wtrk)^\top \htrk(\wtk-\wtrk),$ we have
$$\dtrk=\frac{1}{2}(\wtrk)^\top \htrk(I-\eta\htrk)^{2t}\wtrk.$$
Taking the derivative of $\dtrk$ in $\eta,$ we have 
$$\frac{\partial }{\partial \eta }\dtrk=-t(\wtrk)^\top (\htrk)^2(I-\eta\htrk)^{2t-1}\wtrk.$$
Conditioning on $\calE_k,$ we know $1/L \leq \lambda_i(\htrk)\leq L$ for all $i\in[n]$ and $\htrk$ is full rank in the row span of $\xtrk$. Therefore, we know $\frac{\partial }{\partial \eta }\dtrk<0$ for all $\eta\in[0,1/L).$ Here, we assume $\n{\wtrk}>0$, which happens with probability $1$. 

Overall, we know that conditioning on $\cap_{k\in[m]}\calE_k$, every $\dtrk$ is strictly decreasing for $\eta\in[0,1/L].$ Since $\hftr(\eta):=\frac{1}{m}\sum_{k=1}^m \dtrk,$ we know $\hftr(\eta)$ is strictly decreasing when $\eta\in[0,1/L].$

At step size $\eta=1/L,$ we have 
\begin{align*}
\dtrk=& \frac{1}{2}(\wtrk)^\top \htrk(I-\eta\htrk)^{2t}\wtrk\\
\leq&  \frac{1}{2}L\pr{1-\frac{1}{L^2}}^t\ns{\wtrk}\leq 2L^2\sigma^2\pr{1-\frac{1}{L^2}}^t,
\end{align*}
where we upper bound $\ns{\wtrk}$ by $4L\sigma^2$ at the last step. Therefore, we have $\hftr(1/L)\leq 2L^2\sigma^2(1-\frac{1}{L^2})^t.$
\end{proofof}

\subsubsection{Lower bounding $\hftr$ for $\eta\in(\teta,\infty)$}\label{sec:large_train_GD}
In this section, we prove that the empirical meta objective is lower bounded by $\Omega(\sigma^2)$ with high probability for $\eta\in(\teta,\infty).$ Step size $\teta$ is defined such that there is a descent probability of diverging for any step size larger than $\teta.$ Then, we show the contribution from these truncated sequence will be enough to provide an $\Omega(\sigma^2)$ lower bound for $\hftr.$ The proof of Lemma~\ref{lem:diverge_train_GD} is given at the end of this section.

\divergeTrainGD*

We define $\teta$ as the smallest step size such that the contribution from the truncated sequence in the population meta objective exceeds certain threshold. The precise definition is as follows.
\begin{definition}\label{def:teta}
Given a training task $P,$ let $\calE_1$ be the event that $\sqrt{d}/\sqrt{L} \leq \sigma_i(\xtr)\leq  \sqrt{Ld}$ and $1/L \leq \lambda_i(\htr) \leq L$ for all $i\in[n]$ and $\sqrt{d}\sigma/4 \leq \n{\xitr}\leq \sqrt{d}\sigma.$ Let $\bar{\calE}_2(\eta)$ be the event that the GD sequence is truncated with step size $\eta.$ Define $\teta$ as follows,
$$\teta=\inf\left\{\eta\geq 0 \middle| \Estr \frac{1}{2}\ns{\wt-\wtr}_{\htr}\indic{\calE_1\cap \bar{\calE}_2(\eta)}\geq \frac{\sigma^2}{L^6}\right\}.$$
\end{definition}

In the next lemma, we prove that for any fixed training set, $\indic{\calE_1\cap \bar{\calE}_2(\eta')}\geq \indic{\calE_1\cap \bar{\calE}_2(\eta)}$ for any $\eta'\geq \eta.$ This immediately implies that $\Pr[\calE_1\cap \bar{\calE}_2(\eta)]$ and $\Estr \frac{1}{2}\ns{\wt-\wtr}_{\htr}\indic{\calE_1\cap \bar{\calE}_2(\eta)}$ is non-decreasing in $\eta.$

Basically we need to show, conditioning on $\calE_1,$ if a GD sequence gets truncated at step size $\eta,$ it must be also truncated for larger step sizes. Let $\{w_{\tau,\eta}'\}$ be the GD sequence without truncation. We only need to show that for any $\tau,$ if $\n{w_{\tau,\eta}'}$ exceeds the norm threshold, $\n{w_{\tau,\eta'}'}$ must also exceed the norm threshold for any $\eta'\geq \eta.$ This is easy to prove if $\tau$ is odd because in this case $\n{w_{\tau,\eta}'}$ is always non-decreasing in $\eta.$ The case when $\tau$ is even is trickier because there indeed exists certain range of $\eta$ such that $\n{w_{\tau,\eta}'}$ is decreasing in $\eta.$ We manage to prove that this problematic case cannot happen when $\n{w_{\tau,\eta}'}$ is at least $4\sqrt{L}\sigma.$ The full proof of Lemma~\ref{lem:monotone} is deferred into Section~\ref{sec:technical_train_GD}.

\begin{lemma}\label{lem:monotone}
Fixing a task $P,$ let $\calE_1$ and $\bar{\calE}_2(\eta)$ be as defined in Definition~\ref{def:teta}. We have 
$$\indic{\calE_1\cap \bar{\calE}_2(\eta')}\geq \indic{\calE_1\cap \bar{\calE}_2(\eta)},$$
for any $\eta'\geq \eta.$ 
\end{lemma}

In the next Lemma, we prove that $\teta$ must lie within $(1/L,3L).$ We prove this by showing that the GD sequence never gets truncated for $\eta\in[0,2/L]$ and almost always gets truncated for $\eta\in[2.5L,\infty).$ The proof is deferred into Section~\ref{sec:technical_train_GD}.

\begin{lemma}\label{lem:range_teta}
Let $\teta$ be as defined in Definition~\ref{def:teta}. Suppose $\sigma$ is a large constant $c_1.$ Assume $t\geq c_2,d\geq c_4$ for some constants $c_2,c_4.$ 
We have 
$$1/L<\teta<3L.$$
\end{lemma}

Now, we are ready to give the proof of Lemma~\ref{lem:diverge_train_GD}.

\begin{proofof}{Lemma~\ref{lem:diverge_train_GD}}
Let $\calE_1$ and $\bar{\calE}_2(\eta)$ be as defined in Definition~\ref{def:teta}.
For the simplicity of the proof, we assume $\Estr \frac{1}{2}\ns{w_{t,\teta}-\wtr}_{\htr}\indic{\calE_1\cap \bar{\calE}_2(\teta)}\geq \frac{\sigma^2}{L^6}.$ We will discuss the proof for the other case at the end, which is very similar.

Conditioning on $\calE_1,$ we know $\frac{1}{2}\ns{w_{t,\teta}-\wtr}_{\htr}\leq 18L^2\sigma^2.$ Therefore, we know $\Pr[\calE_1\cap \bar{\calE}_2(\teta)]\geq \frac{1}{18L^8}.$ 
For each task $P_k$, define $\calE_1^{(k)}$ and $\bar{\calE}_2^{(k)}(\eta)$ as the corresponding events on training set $\strk.$ By Hoeffding's inequality, we know with probability at least $1-\expm,$ 
$$\frac{1}{m}\sum_{k=1}^m \indic{\calE_1^{(k)}\cap \bar{\calE}_2^{(k)}(\teta)}\geq \frac{1}{20L^8}.$$
By Lemma~\ref{lem:monotone}, we know $\indic{\calE_1^{(k)}\cap \bar{\calE}_2^{(k)}(\eta)}\geq\indic{\calE_1^{(k)}\cap \bar{\calE}_2^{(k)}(\teta)}$ for any $\eta\geq \teta.$ Then, we can lower bound $\hftr$ for any $\eta>\teta$ as follows,
\begin{align*}
\hftr(\eta)=\frac{1}{m}\sum_{k=1}^m \frac{1}{2}\ns{\wtk-\wtrk}_{\htrk}
\geq& \frac{1}{m}\sum_{k=1}^m \frac{1}{2}\ns{\wtk-\wtrk}_{\htrk}\indic{\calE_1^{(k)}\cap \bar{\calE}_2^{(k)}(\eta)}\\
\geq& 2\sigma^2\frac{1}{m}\sum_{k=1}^m\indic{\calE_1^{(k)}\cap \bar{\calE}_2^{(k)}(\eta)}\\
\geq& 2\sigma^2\frac{1}{m}\sum_{k=1}^m\indic{\calE_1^{(k)}\cap \bar{\calE}_2^{(k)}(\teta)}\geq \frac{\sigma^2}{10L^8},
\end{align*}
where the second inequality lower bounds the loss for one task by $2\sigma^2$ when the sequence gets truncated.

We have assumed $\Estr \frac{1}{2}\ns{w_{t,\teta}-\wtr}_{\htr}\indic{\calE_1\cap \bar{\calE}_2(\teta)}\geq \frac{\sigma^2}{L^6}$ in the proof. Now, we show the proof also works when $\Estr \frac{1}{2}\ns{w_{t,\teta}-\wtr}_{\htr}\indic{\calE_1\cap \bar{\calE}_2(\teta)}< \frac{\sigma^2}{L^6}$ with slight changes. According to the definition and Lemma~\ref{lem:monotone}, we know $\Estr \frac{1}{2}\ns{w_{t,\teta}-\wtr}_{\htr}\indic{\calE_1\cap \bar{\calE}_2(\eta)}> \frac{\sigma^2}{L^6}$ for all $\eta>\teta.$ At each training set $\str,$ we can define $\indic{\calE_1\cap \bar{\calE}_2(\teta')}$ as $\lim_{\eta\rightarrow \teta^+}\indic{\calE_1\cap \bar{\calE}_2(\eta)}.$ We also have $\Pr[\calE_1\cap \bar{\calE}_2(\teta')]\geq \frac{1}{18L^8}.$ The remaining proof is the same as before as we substitute $\indic{\calE_1\cap \bar{\calE}_2(\teta)}$ by $\indic{\calE_1\cap \bar{\calE}_2(\teta')}$.
\end{proofof}

\subsubsection{Generalization for $\eta\in[1/L,\teta]$}\label{sec:large_train_GD2}
In this section, we show empirical meta objective $\hftr$ is point-wise close to population meta objective $\bftr$ for all $\eta\in[1/L,\teta].$

\generaTrainGD*

In this section, we first show $\hftr$ concentrates on $\bftr$ for any fixed $\eta$ and then construct $\epsilon$-net for $\hftr$ and $\bftr$ for $\eta\in[1/L,\teta].$ We give the proof of Lemma~\ref{lem:genera_train_GD} at the end.

We first show that for a fixed $\eta,$ $\hftr(\eta)$ is close to $\bftr(\eta)$ with high probability. We prove the meta-loss on each task $\dtrk$ is $O(1)$-subexponential. Then we apply Bernstein's inequality to get the result. The proof is deferred into Section~\ref{sec:technical_train_GD}. We will assume $\sigma$ is a large constant and $t\geq c_2,d\geq c_4$ for some constants $c_2,c_4$ so that Lemma~\ref{lem:range_teta} holds and $\teta$ is a constant.

\begin{lemma}\label{lem:genera_fixed_eta_train_GD}
Suppose $\sigma$ is a constant. For any fixed $\eta$ and any $1>\epsilon>0,$ with probability at least $1-\expepsm,$
$$\absr{\hftr(\eta)-\bftr(\eta) }\leq \epsilon.$$
\end{lemma}

Next, we construct an $\epsilon$-net for $\bftr.$ By the definition of $\teta,$ we know for any $\eta\leq \teta,$ the contribution from truncated sequences in $\bftr(\eta)$ is small. We can show the contribution from the un-truncated sequences is $O(t)$-lipschitz. 

\begin{lemma}\label{lem:eps_net_train_expect}
Suppose $\sigma$ is a large constant $c_1$. Assume $t\geq c_2,d\geq c_4$ for some constant $c_2,c_4.$ There exists an $\frac{11\sigma^2}{L^4}$-net $N\subset [1/L,\teta]$ for $\bftr$ with $|N|=O(t).$ That means, for any $\eta\in[1/L,\teta],$
$$|\bftr(\eta)-\bftr(\eta')|\leq \frac{11\sigma^2}{L^4},$$
for $\eta'=\arg\min_{\eta''\in N,\eta''\leq \eta}(\eta-\eta'').$
\end{lemma}

\begin{proofof}{Lemma~\ref{lem:eps_net_train_expect}}
Let $\calE_1$ and $\bar{\calE}_2(\eta)$ be as defined in Definition~\ref{def:teta}.
For the simplicity of the proof, we assume $\Estr \frac{1}{2}\ns{w_{t,\teta}-\wtr}_{\htr}\indic{\calE_1\cap \bar{\calE}_2(\teta)}\leq \frac{\sigma^2}{L^6}.$ We will discuss the proof for the other case at the end, which is very similar.

We can divide $\Estr \frac{1}{2}\ns{w_{t,\eta}-\wtr}_{\htr}$ as follows,
\begin{align*}
&\Estr \frac{1}{2}\ns{w_{t,\eta}-\wtr}_{\htr}\\ 
=& \Estr \frac{1}{2}\ns{w_{t,\eta}-\wtr}_{\htr}\indic{\calE_1\cap \calE_2(\teta)}+\Estr \frac{1}{2}\ns{w_{t,\eta}-\wtr}_{\htr}\indic{\calE_1\cap \bar{\calE}_2(\teta)}\\
&+\Estr \frac{1}{2}\ns{w_{t,\eta}-\wtr}_{\htr}\indic{\bar{\calE}_1}.
\end{align*}

We will construct an $\epsilon$-net for the first term and show the other two terms are small. Let's first consider the third term. Since $\frac{1}{2}\ns{w_{t,\eta}-\wtr}_{\htr}$ is $O(1)$-subexponential and $\Pr[\bar{\calE}_1]\leq \expd$, we have $\Estr \frac{1}{2}\ns{w_{t,\eta}-\wtr}_{\htr}\indic{\bar{\calE}_1}=O(1)\expd.$ Choosing $d$ to be at least certain constant, we know $\frac{1}{2}\ns{w_{t,\eta}-\wtr}_{\htr}\indic{\bar{\calE}_1}\leq \sigma^2/L^4.$

Then we upper bound the second term. Since $\Estr \frac{1}{2}\ns{w_{t,\teta}-\wtr}_{\htr}\indic{\calE_1\cap \bar{\calE}_2(\teta)}\leq \frac{\sigma^2}{L^6}$ and \\$\frac{1}{2}\ns{w_{t,\teta}-\wtr}_{\htr}\geq 2\sigma^2$ when $w_{t,\teta}$ diverges, we know $\Pr[\calE_1\cap \bar{\calE}_2(\teta)]\leq \frac{1}{2L^6}.$ Then, we can upper bound the second term as follows,
\begin{align*}
\Estr \frac{1}{2}\ns{w_{t,\eta}-\wtr}_{\htr}\indic{\calE_1\cap \bar{\calE}_2(\teta)}\leq 18L^2\sigma^2\frac{1}{2L^6}=\frac{9\sigma^2}{L^4}
\end{align*}

Next, we show the first term $\frac{1}{2}\ns{w_{t,\eta}-\wtr}_{\htr}\indic{\calE_1\cap \calE_2(\teta)}$ has desirable Lipschitz condition. According to Lemma~\ref{lem:monotone}, we know $\indic{\calE_1\cap \calE_2(\eta)}\geq \indic{\calE_1\cap \calE_2(\teta)}$ for any $\eta\leq \teta.$ Therefore, conditioning on $\calE_1\cap \calE_2(\teta)$, we know $\wt$ never gets truncated for any $\eta\leq \teta.$ This means $\wt=\bt\wtr$ with $\bt=(I-(I-\eta\htr)^t).$ We can compute the derivative of $\frac{1}{2}\ns{w_{t,\eta}-\wtr}_{\htr}$ as follows,
$$\frac{\partial}{\partial \eta}\frac{1}{2}\ns{w_{t,\eta}-\wtr}_{\htr}=\inner{t\htr(I-\eta\htr)^{t-1}\wtr}{\htr(\wt-\wtr)}.$$
Since $\n{\wt}=\n{(I-(I-\eta\htr)^t)\wtr}\leq 4\sqrt{L}\sigma$ and $\n{\wtr}\leq 2\sqrt{L}\sigma,$ we have $\n{(I-\eta\htr)^t\wtr}\leq 6\sqrt{L}\sigma.$ We can bound $\n{(I-\eta\htr)^{t-1}\wtr}$ with $\n{(I-\eta\htr)^t\wtr}+\n{\wtr}$ by bounding the expanding directions using $\n{(I-\eta\htr)^t\wtr}$ and bounding the shrinking directions using $\n{\wtr}.$ Therefore, we can bound the derivative as follows,
\begin{align*}
\absr{\frac{\partial}{\partial \eta}\frac{1}{2}\ns{w_{t,\eta}-\wtr}_{\htr}}\leq tL\times 8\sqrt{L}\sigma\times 6L\sqrt{L}\sigma=48L^3\sigma^2t.
\end{align*}
Suppose $\sigma$ is a constant, we know $\Estr \frac{1}{2}\ns{w_{t,\eta}-\wtr}_{\htr}\indic{\calE_1\cap \calE_2(\teta)}$ is $O(t)$-lipschitz. Therefore, there exists an $\frac{\sigma^2}{L^4}$-net $N$ for $\Estr \frac{1}{2}\ns{w_{t,\eta}-\wtr}_{\htr}\indic{\calE_1\cap \calE_2(\teta)}$ with size $O(t)$. That means, for any $\eta\in[1/L,\teta],$
$$\absr{\Estr \frac{1}{2}\ns{w_{t,\eta}-\wtr}_{\htr}\indic{\calE_1\cap \calE_2(\teta)}-\Estr \frac{1}{2}\ns{w_{t,\eta'}-\wtr}_{\htr}\indic{\calE_1\cap \calE_2(\teta)}}\leq \frac{\sigma^2}{L^4}$$ 
for $\eta'=\arg\min_{\eta''\in N,\eta''\leq \eta}(\eta-\eta'').$ Note we construct the $\epsilon$-net in a particular way such that $\eta'$ is chosen as the largest step size in $N$ that is at most $\eta.$

Combing with the upper bounds on the second term and the third term, we have for any $\eta\in[1/L,\teta],$
$$\absr{\bftr(\eta)-\bftr(\eta')}\leq \frac{11\sigma^2}{L^4}$$ 
for $\eta'=\arg\min_{\eta''\in N,\eta''\leq \eta}(\eta-\eta'').$

In the above analysis, we have assumed $\Estr \frac{1}{2}\ns{w_{t,\teta}-\wtr}_{\htr}\indic{\calE_1\cap \bar{\calE}_2(\teta)}\leq \frac{\sigma^2}{L^6}.$ The proof can be easily generalized to the other case. We can define $\indic{\calE_1\cap \bar{\calE}_2(\teta')}$ as $\lim_{\eta\rightarrow \teta^-}\indic{\calE_1\cap \bar{\calE}_2(\eta)}.$ Then the proof works as long as we substitute $\indic{\calE_1\cap \bar{\calE}_2(\teta)}$ by $\indic{\calE_1\cap \bar{\calE}_2(\teta')}.$ We will also add $\teta$ into the $\epsilon$-net.
\end{proofof}

In order to prove $\bftr$ is close to $\hftr$ point-wise in $[1/L,\teta],$ we still need to construct an $\epsilon$-net for the empirical meta objective $\hftr.$

\begin{lemma}\label{lem:eps_net_train_empirical}
Suppose $\sigma$ is a large constant $c_1$. Assume $t\geq c_2,d\geq c_4$ for certain constants $c_2,c_4.$ With probability at least $1-m\expd,$ there exists an $\frac{\sigma^2}{L^4}$-net $N'\subset [1/L,\teta]$ for $\hftr$ with $|N|=O(t+m).$ That means, for any $\eta\in[1/L,\teta],$
$$|\hftr(\eta)-\hftr(\eta')|\leq \frac{\sigma^2}{L^4},$$
for $\eta'=\arg\min_{\eta''\in N',\eta''\leq \eta}(\eta-\eta'').$
\end{lemma}

\begin{proofof}{Lemma~\ref{lem:eps_net_train_empirical}}
For each $k\in[m],$ let $\calE_{1,k}$ be the event that $\sqrt{d}/\sqrt{L} \leq \sigma_i(\xtrk)\leq  \sqrt{Ld}$ and $1/L \leq \lambda_i(\htrk) \leq L$ for all $i\in[n]$ and $\sqrt{d}\sigma/4 \leq \n{\xitrk}\leq \sqrt{d}\sigma$. According to Lemma~\ref{lem:isotropic} and Lemma~\ref{lem:norm_vector}, we know with probability at least $1-m\expd,$ $\calE_{1,k}$'s hold for all $k\in[m].$ From now on, we assume all these events hold.

Recall that the empirical meta objective as follows,
$$\hftr(\eta):=\frac{1}{m}\sum_{k=1}^m \dtrk.$$

For any $k\in[m],$ let $\eta_{c,k}$ be the smallest step size such that $\wtk$ gets truncated. If $\eta_{c,k}>\heta,$ by similar argument as in Lemma~\ref{lem:eps_net_train_expect}, we know $\dtrk$ is $O(t)$-Lipschitz in $[1/L,\heta]$ as long as $\sigma$ is a constant. If $\eta_{c,k}\leq \heta,$ by Lemma~\ref{lem:monotone} we know $\wtk$ gets truncated for any $\eta\geq \eta_{c,k}.$ This then implies that $\dtrk$ is a constant function for $\eta\in[\eta_{c,k},\heta].$ We can also show that $\dtrk$ is $O(t)$-Lipschitz in $[1/L,\eta_{c,k}).$ There might be a discontinuity in function value at $\eta_{c,k}$, so we need to add $\eta_{c,k}$ into the $\epsilon$-net.

Overall, we know there exists an $\frac{\sigma^2}{L^4}$-net $N'$ with $|N'|=O(t+m)$ for $\hftr$. That means, for any $\eta\in[1/L,\teta],$
$$\absr{\hftr(\eta)-\hftr(\eta')}\leq \frac{\sigma^2}{L^4}$$ 
for $\eta'=\arg\min_{\eta''\in N',\eta''\leq \eta}(\eta-\eta'').$ 
\end{proofof}

Finally, we combine Lemma~\ref{lem:genera_fixed_eta_train_GD}, Lemma~\ref{lem:eps_net_train_expect} and Lemma~\ref{lem:eps_net_train_empirical} to prove that $\hftr$ is point-wise close to $\bftr$ for $\eta\in[1/L,\teta].$

\begin{proofof}{Lemma~\ref{lem:genera_train_GD}}
We assume $\sigma$ as a constant in this proof.
By Lemma~\ref{lem:genera_fixed_eta_train_GD}, we know with probability at least $1-\expepsm,$
$\absr{\hftr(\eta)-\bftr(\eta) }\leq \epsilon$ for any fixed $\eta.$ By Lemma~\ref{lem:eps_net_train_expect}, we know there exists an $\frac{11\sigma^2}{L^4}$-net $N$ for $\bftr$ with size $O(t).$ By Lemma~\ref{lem:eps_net_train_empirical}, we know with probability at least $1-m\expd$, there exists an $\frac{\sigma^2}{L^4}$-net $N'$ for $\hftr$ with size $O(t+m).$   
According to the proofs of Lemma~\ref{lem:eps_net_train_expect} and Lemma~\ref{lem:eps_net_train_empirical}, it's not hard to verify that $N\cup N'$ is still an $\frac{11\sigma^2}{L^4}$-net for $\hftr$ and $\bftr$. That means, for any $\eta\in[1/L,\teta],$ we have
$$|\bftr(\eta)-\bftr(\eta')|,|\hftr(\eta)-\hftr(\eta')|\leq \frac{11\sigma^2}{L^4},$$ 
for $\eta'=\arg\min_{\eta''\in N\cup N',\eta''\leq \eta}(\eta-\eta'').$

Taking a union bound over $N\cup N',$ we have with probability at least $1-O(t+m)\expm,$
$$\absr{\hftr(\eta)-\bftr(\eta) }\leq \frac{\sigma^2}{L^4}$$ for all $\eta\in N\cup N'.$

Overall, we know with probability at least $1-m\expd-O(t+m)\expm,$ for all $\eta\in[1/L,\teta],$
\begin{align*}
&|\bftr(\eta)-\hftr(\eta)| \\
\leq& |\bftr(\eta)-\bftr(\eta')|+|\hftr(\eta)-\hftr(\eta')| + |\hftr(\eta')-\bftr(\eta')|\\
\leq& \frac{23\sigma^2}{L^4}\leq \frac{\sigma^2}{L^3},
\end{align*}
where $\eta'=\arg\min_{\eta''\in N\cup N',\eta''\leq \eta}(\eta-\eta'').$ We use the fact that $L=100$ in the last inequality.
\end{proofof}

\subsubsection{Proofs of Technical Lemmas}\label{sec:technical_train_GD}
\begin{proofof}{Lemma~\ref{lem:isotropic}}
Recall that $\xtr$ is an $n\times d$ matix with $n=cd$ where $c\in [1/4,3/4].$
According to Lemma~\ref{lem:sig_matrix}, with probability at least $1-2\exp(-t^2/2),$ we have
$$\sqrt{d}-\sqrt{cd}-t\leq \sigma_i(\xtr)\leq \sqrt{d}+\sqrt{cd}+t,$$
for all $i\in[n].$

Since $\htr=1/n\xtr^\top \xtr,$ we know $\lambda_i(\htr)=1/n\sigma_i^2(\xtr).$ Since $c\in[\frac{1}{4},\frac{3}{4}],$ we have 
$\frac{1}{cd}(\sqrt{d}+\sqrt{cd})^2\leq 100-c' \text{ and } \frac{1}{cd}(\sqrt{d}-\sqrt{cd})^2\geq \frac{1}{100}+c',$
for some constant $c'.$ Therefore, we know with probability at least $1-\expd,$
$$\frac{1}{100} \leq \lambda_i(\htr) \leq 100,$$
for all $i\in[n].$

Similarly, since there exists constant $c''$ such that $\sqrt{d}+\sqrt{cd}\leq (10-c'')\sqrt{d}$ and $\sqrt{d}-\sqrt{cd}\geq (1/10+c'')\sqrt{d},$ we know with probability at least $1-\expd,$ 
$$\frac{1}{10}\sqrt{d} \leq \sigma_i(\xtr) \leq 10\sqrt{d},$$
for all $i\in[n].$ Choosing $L=100$ finishes the proof.
\end{proofof}

\begin{proofof}{Lemma~\ref{lem:monotone}}
We prove that for any training set $\str,$ $\indic{\calE_1\cap \bar{\calE}_2(\eta')}\geq \indic{\calE_1\cap \bar{\calE}_2(\eta')}$ for any $\eta'>\eta.$ This is trivially true if $\calE_1$ is false on $\str.$ Therefore, we focus on the case when $\calE_1$ holds for $\str.$ Suppose $\eta_c$ is the smallest step size such that the GD sequence gets truncated. Let $\{w_{\tau,\eta_c}'\}$ be the GD sequence without truncation. There must exists $\tau\leq t$ such that $\n{w_{\tau,\eta_c}'}\geq 4\sqrt{L}\sigma.$ We only need to prove that $\n{w_{\tau,\eta}'}\geq 4\sqrt{L}\sigma$ for any $\eta\geq \eta_c.$ We prove this by showing the derivative of $\ns{w_{\tau,\eta}'}$ in $\eta$ is non-negative assuming $\ns{w_{\tau,\eta}'}\geq 4\sqrt{L}\sigma.$

Recall the recursion of $w_{\tau,\eta}'$ as $w_{\tau,\eta}'=\wtr-(I-\eta\htr)^\tau \wtr.$ If $\tau$ is an odd number, it's clear that $\frac{\partial }{\partial \eta}\ns{w'_{\tau,\eta}}$ is non-negative at any $\eta\geq 0.$ From now on, we assume $\tau$ is an even number. Actually in this case, $\frac{\partial }{\partial \eta}\ns{w'_{\tau,\eta}}$ can be negative for some $\eta.$ However, we can prove the derivative must be non-negative assuming $\ns{w_{\tau,\eta}'}\geq 4\sqrt{L}\sigma.$

Suppose the eigenvalue decomposition of $\htr$ is $\sum_{i=1}^n\lambda_i u_iu_i^\top$ with $\lambda_1\geq \cdots \lambda_n.$ Denote $c_i$ as $\inner{\wtr}{u_i}.$ Let $\lambda_j$ be the smallest eigenvalue such that $(1-\eta\lambda_j)\leq -1.$ This implies $\lambda_i\leq 2/\eta$ for any $i\geq j+1.$
We can write down $\ns{w_{\tau,\eta}'}$ as follows
\begin{align*}
\ns{w_{\tau,\eta}'}
=& \sum_{i=1}^j\pr{1-(1-\eta\lambda_i)^t}^2 c_i^2+\sum_{i=j+1}^n\pr{1-(1-\eta\lambda_i)^t}^2 c_i^2\\
\leq& \sum_{i=1}^j\pr{1-(1-\eta\lambda_i)^t}^2 c_i^2+\ns{\wtr}.
\end{align*}
Since $\calE_1$ holds, we know $\ns{\wtr}\leq 4L\sigma^2$. Combining with $\ns{w'_{\tau,\eta}}\geq 16L\sigma^2,$ we have $\sum_{i=1}^j\pr{1-(1-\eta\lambda_i)^t}^2 c_i^2\geq 12L\sigma^2.$ We can lower bound the derivative as follows,
\begin{align*}
\frac{\partial }{\partial \eta}\ns{w_{\tau,\eta}}
=& \sum_{i=1}^j 2t\lambda_i(1-\eta\lambda_i)^{t-1}\pr{1-(1-\eta\lambda_i)^t} c_i^2+\sum_{i=j+1}^n2t\lambda_i(1-\eta\lambda_i)^{t-1}\pr{1-(1-\eta\lambda_i)^t} c_i^2\\
\geq& 2t\sum_{i=1}^j\lambda_i (1-\eta\lambda_i)^{t-1}\pr{1-(1-\eta\lambda_i)^t} c_i^2-2t\frac{2}{\eta}\sum_{i=j+1}^n c_i^2\\
\geq& 2t\sum_{i=1}^j\lambda_i (1-\eta\lambda_i)^{t-1}\pr{1-(1-\eta\lambda_i)^t} c_i^2-2t\times 8L\sigma^2/\eta.
\end{align*}
Then, we only need to show that $\sum_{i=1}^j \lambda_i(1-\eta\lambda_i)^{t-1}\pr{1-(1-\eta\lambda_i)^t} c_i^2$ is larger than $8L\sigma^2/\eta.$ We have 
\begin{align*}
\sum_{i=1}^j \lambda_i(1-\eta\lambda_i)^{t-1}\pr{1-(1-\eta\lambda_i)^t} c_i^2
=& \sum_{i=1}^j \lambda_i\frac{(1-\eta\lambda_i)^{t-1}}{1-(1-\eta\lambda_i)^t}\pr{1-(1-\eta\lambda_i)^t}^2 c_i^2\\
=& \sum_{i=1}^j \lambda_i\frac{(\eta\lambda_i-1)^{t-1}}{(\eta\lambda_i-1)^t-1}\pr{1-(1-\eta\lambda_i)^t}^2 c_i^2\\
=& \sum_{i=1}^j \lambda_i\frac{(\eta\lambda_i-1)^{t}}{(\eta\lambda_i-1)^t-1}\frac{1}{\eta\lambda_i-1}\pr{1-(1-\eta\lambda_i)^t}^2 c_i^2\\
\geq& \sum_{i=1}^j \frac{1}{\eta}\pr{1-(1-\eta\lambda_i)^t}^2 c_i^2\geq 12L\sigma^2/\eta> 8L\sigma^2/\eta.
\end{align*}
\end{proofof}

\begin{proofof}{Lemma~\ref{lem:range_teta}}
Similar as the analysis in Lemma~\ref{lem:small_step_train}, conditioning on $\calE_1,$ we know the GD sequence never exceeds the norm threshold for any $\eta\in[0,2/L].$ This then implies 
$$\Estr \frac{1}{2}\ns{\wt-\wtr}_{\htr}\indic{\calE_1\cap \bar{\calE}_2(\eta)}=0,$$
for all $\eta\in[0,2/L].$ 

Let $\{w_{\tau,\eta}'\}$ be the GD sequence without truncation. For any step size $\eta\in[2.5L,\infty],$ conditioning on $\calE_1,$ we have
$$\n{\wt'}\geq \pr{(\eta/L-1)^t -1}\n{\wtr}\geq \pr{1.5^t -1}\pr{\frac{\sigma}{4\sqrt{L}}-1}\geq 4\sqrt{L}\sigma,$$
where the last inequality holds as long as $\sigma\geq 5\sqrt{L},t\geq c_2$ for some constant $c_2$. Therefore, we know when $\eta\in[2.5L,\infty),$ $\indic{\calE_1\cap \bar{\calE}_2(\eta)}=\indic{\calE_1}$. Then, we have for any $\eta\geq 2.5L,$
\begin{align*}
\Estr \frac{1}{2}\ns{\wt-\wtr}_{\htr}\indic{\calE_1\cap \bar{\calE}_2(\eta)}
\geq& \frac{1}{2L}\pr{4\sqrt{L}\sigma-2\sqrt{L}\sigma}^2\Pr[\calE_1]\geq 2\sigma^2\Pr[\calE_1]\geq \frac{\sigma^2}{L^3},
\end{align*}
where the last inequality uses $\Pr[\calE_1]\geq 1-\expd$ and assume $d\geq c_4$ for some constant $c_4.$

Overall, we know $\Estr \frac{1}{2}\ns{\wt-\wtr}_{\htr}\indic{\calE_1\cap \bar{\calE}_2(\eta)}$ equals zero for all $\eta\in[0,2/L]$ and is at least $\frac{\sigma^2}{L^3}$ for all $\eta\in[2.5L,\infty).$ By definition, we know $\teta\in(1/L,3L).$ 
\end{proofof}

\begin{proofof}{Lemma~\ref{lem:genera_fixed_eta_train_GD}}
Recall that $\hftr(\eta):=\frac{1}{m}\sum_{k=1}^m \dtrk.$
We prove that each $\dtrk$ is $O(1)$-subexponential. We can further write $\dtrk$ as follows,
\begin{align*}
\dtrk =& \frac{1}{2}\ns{\wtk-w^*_k-(\xtrk)^\dagger\xitrk}_{\htrk}\\
\leq& \frac{1}{2}\ns{\wtk-w^*_k}\n{\htrk}+\frac{1}{2n}\ns{\xitrk}+\n{\wtk-w^*_k}\pr{\frac{1}{\sqrt{n}}\n{\xitrk}}\pr{\frac{1}{\sqrt{n}}\n{\xtrk}}.
\end{align*}
We can write $\n{\htrk}$ as $\sigma_{\max}^2(\frac{1}{\sqrt{n}} \xtrk).$ According to Lemma~\ref{lem:concen_gauss}, we know $\sigma_{\max}(\xtrk)-\E\sigma_{\max}(\xtrk)$ is $O(1)$-subgaussian, which implies that $\sigma_{\max}(\frac{1}{\sqrt{n}} \xtrk)-\E \sigma_{\max}(\frac{1}{\sqrt{n}} \xtrk)$ is $O(1/\sqrt{d})$-subgaussian. Since $\E \sigma_{\max}(\frac{1}{\sqrt{n}} \xtrk)$ is a constant, we know $\sigma_{\max}(\frac{1}{\sqrt{n}} \xtrk)$ is $O(1)$-subgaussian and $\sigma_{\max}^2(\frac{1}{\sqrt{n}} \xtrk)$ is $O(1)$-subexponential. Similarly, we know both $\frac{1}{2n}\ns{\xitrk}$ and $\pr{\frac{1}{\sqrt{n}}\n{\xtrk}}\pr{\frac{1}{\sqrt{n}}\n{\xitrk}}$ are $O(1)$-subexponential.

Suppose $\sigma$ is a constant, we know $\n{\wtk-w_k^*}$ is upper bounded by a constant. Then, we know $\dtrk$ is $O(1)$-subexponential. Therefore, $\hftr(\eta)$ is the average of $m$ i.i.d. $O(1)$-subexponential random variables. By standard concentration inequality, we know for any $1>\epsilon>0,$ with probability at least $1-\expepsm,$
$$\absr{\hftr(\eta)-\bftr(\eta) }\leq \epsilon.$$
\end{proofof}

\subsection{Train-by-validation (GD)}\label{sec:proofs_valid_GD}
In this section, we show that the optimal step size under $\hfva$ is $\Theta(1/t).$ Furthermore, we show under this optimal step size, GD sequence makes constant progress towards the ground truth. Precisely, we prove the following theorem.

\thmValidGD*

In this section, we still use $L$ to denote constant $100.$ We start from analyzing the behavior of the population meta-objective $\bfva$ for step sizes within $[0,1/L].$ We show the optimal step size within this range is $\Theta(1/t)$ and GD sequence moves towards $w^*$ under the optimal step size. The following lemma is a formal version of Lemma~\ref{lem:expectation_small_informal}.
This serves as step 1 in Section~\ref{sec:strategy}. 
We defer the proof of Lemma~\ref{lem:expectation_small} into Section~\ref{sec:bfva_small}.

\begin{restatable}
{lemma}{expectSmall}\label{lem:expectation_small}
Suppose noise level $\sigma$ is a large enough constant $c_1$. Assume unroll length $t\geq c_2$ and dimension $d\geq c_4$ for some constants $c_2,c_4.$ There exist $\eta_1,\eta_2,\eta_3=\Theta(1/t)$ with $\eta_1<\eta_2<\eta_3$ such that 
\begin{align*}
\bfva(\eta_2)&\leq \frac{1}{2}\ns{w^*}-\frac{9}{10}C+\frac{\sigma^2}{2}\\
\bfva(\eta)&\geq \frac{1}{2}\ns{w^*}-\frac{6}{10}C+\frac{\sigma^2}{2},\forall \eta\in[0,\eta_1]\cup [\eta_3,1/L] 
\end{align*}
where $C$ is a positive constant.
\end{restatable}

To relate the behavior of $\bfva$ to the behavior of $\hfva$, we prove the following generalization result for step sizes in $[0,1/L].$ The following lemma is a formal version of Lemma~\ref{lem:genera_valid_GD_informal}.
This serves as step 3 in Section~\ref{sec:strategy}. The proof is deferred into Section~\ref{sec:genera_GD_valid}.

\begin{restatable}
{lemma}{generaValidGD}\label{lem:genera_valid_GD}
For any $1>\epsilon>0,$ assume $d\geq c_4\log(1/\epsilon)$ for some constant $c_4$. With probability at least $1-O(1/\epsilon)\expepsm$, 
$$|\hfva(\eta)-\bfva(\eta)|\leq \epsilon,$$
for all $\eta\in[0,1/L].$
\end{restatable}

In Lemma~\ref{lem:large_valid_GD}, we show the empirical meta objective $\hfva$ is high for all step size larger than $1/L$, which then implies $\etava\in[0,1/L].$ The following lemma is a formal version of Lemma~\ref{lem:large_valid_GD_informal}. This serves as step 2 in Section~\ref{sec:strategy}. We prove this lemma in Section~\ref{sec:large_valid_GD}.

\begin{restatable}
{lemma}{largeValidGD}\label{lem:large_valid_GD}
Suppose $\sigma$ is a large constant. Assume $t\geq c_2,d\geq c_4\log(t)$ for some constants $c_2,c_4.$ With probability at least $1-\expm,$
\begin{align*}
\hfva(\eta)\geq& C'\sigma^2+\frac{1}{2}\sigma^2,
\end{align*}
for all $\eta\geq 1/L,$ where $C'$ is a positive constant independent with $\sigma.$
\end{restatable}

Combining Lemma~\ref{lem:expectation_small}, Lemma~\ref{lem:genera_valid_GD} and Lemma~\ref{lem:large_valid_GD}, we give the proof of Theorem~\ref{thm:valid_GD}.

\begin{proofof}{Theorem~\ref{thm:valid_GD}}
According to Lemma~\ref{lem:expectation_small}, we know as long as $d$ and $t$ are larger than certain constants, there exists $\eta_1,\eta_2,\eta_3=\Theta(1/t)$ with $\eta_1<\eta_2<\eta_3$ such that 
\begin{align*}
\bfva(\eta_2)&\leq \frac{1}{2}\ns{w^*}-\frac{9}{10}C+\sigma^2/2\\
\bfva(\eta)&\geq \frac{1}{2}\ns{w^*}-\frac{6}{10}C+\sigma^2/2,\forall \eta\in[0,\eta_1]\cup [\eta_3,1/L],
\end{align*}
for some positive constant $C.$

Choosing $\epsilon=\min(1,C/10)$ in Lemma~\ref{lem:genera_valid_GD}, we know as long as $d$ is larger than certain constant, with probability at least $1-\expm,$
$$|\hfva(\eta)-\bfva(\eta)|\leq C/10,$$
for all $\eta\in[0,1/L].$

Therefore,
\begin{align*}
\hfva(\eta_2)&\leq \frac{1}{2}\ns{w^*}-\frac{8}{10}C+\sigma^2/2\\
\hfva(\eta)&\geq \frac{1}{2}\ns{w^*}-\frac{7}{10}C+\sigma^2/2,\forall \eta\in[0,\eta_1]\cup [\eta_3,1/L].
\end{align*}

By Lemma~\ref{lem:large_valid_GD}, we know as long as $t\geq c_2,d\geq c_4\log(t)$ for some constants $c_2,c_4,$ with probability at least $1-\expm,$
$$\hfva(\eta)\geq C'\sigma^2+\frac{1}{2}\sigma^2,$$
for all $\eta\geq 1/L.$ As long as $\sigma \geq 1/\sqrt{C'},$ we have $\hfva(\eta)\geq 1+\frac{1}{2}\sigma^2$ for all $\eta\geq 1/L.$ Combining with $\hfva(\eta_2)\leq \frac{1}{2}\ns{w^*}-\frac{8}{10}C+\sigma^2/2$, we know $\etava\in[0,1/L].$ Furthermore, since $\hfva(\eta)\geq \frac{1}{2}\ns{w^*}-\frac{7}{10}C+\sigma^2/2,\forall \eta\in[0,\eta_1]\cup [\eta_3,1/L]$, we have $\eta_1\leq \etava\leq \eta_3.$ 

Recall that $\eta_1,\eta_3=\Theta(1/t),$ we know $\etava=\Theta(1/t).$ At the optimal step size, we have
$$\bfva(\etava)\leq \hfva(\etava)+C/10\leq \hfva(\eta_2)+C/10\leq \frac{1}{2}\ns{w^*}-\frac{7}{10}C+\sigma^2/2.$$
Since $\bfva(\etava)=\E\frac{1}{2}\ns{w_{t,\etava}-w^*}+\sigma^2/2,$ we have 
$$\E\ns{w_{t,\etava}-w^*}\leq \ns{w^*}-\frac{7}{5}C.$$
Choosing $m$ to be at least certain constant, this holds with probability at least $0.99.$
\end{proofof}

\subsubsection{Behavior of $\bfva$ for $\eta\in[0,1/L]$}\label{sec:bfva_small}
In this section, we study the behavior of $\bfva$ when $\eta\in[0,1/L].$ We prove the following Lemma.

\expectSmall*

It's not hard to verify that $\bfva(\eta)=\E 1/2\ns{\wt - w^*}+\sigma^2/2.$ For convenience, denote $Q(\eta):=1/2\ns{\wt - w^*}.$ In order to prove Lemma~\ref{lem:expectation_small}, we only need to show that $\E Q(\eta_2)\leq \frac{1}{2}\ns{w^*}-\frac{9}{10}C$ and $\E Q(\eta)\geq \frac{1}{2}\ns{w^*}-\frac{6}{10}C$ for all $\eta\in[0,\eta_1]\cup [\eta_3,1/L]$. In Lemma~\ref{lem:whp_valid}, we first show that this happens with high probability over the sampling of tasks.

\begin{lemma}\label{lem:whp_valid}
Suppose noise level $\sigma$ is a large enough constant $c_1.$ Assume unroll length $t\geq c_2$ for certain constant $c_2.$ Then, with probability at least $1-\expd$ over the sampling of tasks, there exists $\eta_1,\eta_2,\eta_3=\Theta(1/t)$ with $\eta_1<\eta_2<\eta_3$ such that 
\begin{align*}
Q(\eta_2):=\frac{1}{2}\ns{w_{t,\eta_2} - w^*}&\leq \frac{1}{2}\ns{w^*}-C\\
Q(\eta):=\frac{1}{2}\ns{\wt - w^*}&\geq \frac{1}{2}\ns{w^*}-\frac{C}{2},\forall \eta\in[0,\eta_1]\cup [\eta_3,1/L] 
\end{align*}
where $C$ is a positive constant.
\end{lemma}

Since we are in the small step size regime, we know the GD sequence converges with high probability and will not be truncated. For now, let's assume $\wt = \bt\wtr^* +\bt(\xtr)^\dagger\xitr,$ where $\bt=I-(I-\eta\htr)^t.$ We have
\begin{align*}
Q(\eta) =& \frac{1}{2}\ns{\bt\wtr^* +\bt(\xtr)^\dagger\xitr-w^*}\\
=& \frac{1}{2}\ns{\bt\wtr^* -w^*}+\frac{1}{2}\ns{\bt(\xtr)^\dagger\xitr}\\
&+\inner{\bt\wtr^* -w^*}{ \bt(\xtr)^\dagger\xitr}\\
=& \frac{1}{2}\ns{w^*}+ \frac{1}{2}\ns{\bt\wtr^*}+\frac{1}{2}\ns{\bt(\xtr)^\dagger\xitr}-\inner{\bt\wtr^*}{ w^*}\\
&+\inner{\bt\wtr^* -w^*}{ \bt(\xtr)^\dagger\xitr}.
\end{align*}  

In Lemma~\ref{lem:crossing_term2}, we show that with high probability the crossing term $\inner{\bt\wtr^* -w^*}{\bt(\xtr)^\dagger\xitr}$ is negligible for all $\eta\in[0,1/L].$ By Hoeffding's inequality, we know the crossing term is small for any fixed $\eta.$ Constructing an $\epsilon$-net for the crossing term in $\eta,$ we can take a union bound and show it's small for all $\eta\in[0,1/L].$ We defer the proof of Lemma~\ref{lem:crossing_term2} to Section~\ref{sec:technical_valid_GD}.

\begin{lemma}\label{lem:crossing_term2}
Assume $\sigma$ is a constant. For any $1>\epsilon>0,$ we know with probability at least $1-O(1/\epsilon)\expepsd$,
$$\left|\inner{B_{t,\eta}\wtr^*-w^*}{B_{t,\eta}(\xtr)^\dagger\xitr}\right|\leq \epsilon,$$
for all $\eta\in[0,1/L].$
\end{lemma}
Denote
$$G(\eta):= \frac{1}{2}\ns{w^*}+ \frac{1}{2}\ns{\bt\wtr^*}+\frac{1}{2}\ns{\bt(\xtr)^\dagger\xitr}-\inner{\bt\wtr^*}{ w^*}.$$
Choosing $\epsilon=C/4$ in Lemma~\ref{lem:crossing_term2}, 
we only need to show $G(\eta_2)\leq \ns{w^*}-5C/4$ and $G(\eta)\geq \ns{w^*}-C/4$ for all $\eta\in[0,\eta_1]\cup [\eta_3,1/L]$.

We first show that there exists $\eta_2 = \Theta(1/t)$ such that $G(\eta_2)\leq \frac{1}{2}\ns{w^*}-5C/4$ for some constant $C$. It's not hard to show that $\frac{1}{2}\ns{\bt\wtr^*}+\frac{1}{2}\ns{\bt(\xtr)^\dagger\xitr}= O(\eta^2 t^2\sigma^2).$ In Lemma~\ref{lem:improvement}, we show that the improvement $\inner{\bt\wtr^*}{ w^*}=\Omega(\eta t)$ is linear in $\eta$. Therefore there exists $\eta_2 = \Theta(1/t)$ such that $G(\eta_2)\leq \frac{1}{2}\ns{w^*}-5C/4$ for some constant $C$. We defer the proof of Lemma~\ref{lem:improvement} to Section~\ref{sec:technical_valid_GD}.
\begin{lemma}\label{lem:improvement}
For any fixed $\eta\in [0,L/t]$ with probability at least $1-\exp(-\Omega(d))$,
$$\inner{B_{t,\eta}\wtr^*}{ w^*} \geq \frac{\eta t}{16L}.$$
\end{lemma}

To lower bound $G(\eta)$ for small $\eta,$ we notice 
$$G(\eta)\geq \frac{1}{2}\ns{w^*} - \inner{\bt\wtr^*}{ w^*}.$$
We can show that $\inner{\bt\wtr^*}{ w^*}= O(\eta t).$ Therefore, there exists $\eta_1=\Theta(1/t)$ such that $\inner{\bt\wtr^*}{ w^*}\leq C/4$ for all $\eta\in[0,\eta_1].$

To lower bound $G(\eta)$ for large $\eta,$ we lower bound $G(\eta)$ using the noise square term,
$$G(\eta)\geq \frac{1}{2}\ns{\bt(\xtr)^\dagger\xitr}.$$
We show that with high probability $\ns{\bt(\xtr)^\dagger\xitr}=\Omega(\sigma^2)$ for all $\eta\in[\log(2)L/t,1/L].$ Therefore, as long as $\sigma$ is larger than some constant, there exists $\eta_3=\Theta(1/t)$ such that $G(\eta)\geq \frac{1}{2}\ns{w^*}$ for all $\eta\in [\eta_3,1/L].$ 

Combing Lemma~\ref{lem:crossing_term2} and Lemma~\ref{lem:improvement}, we give a complete proof for Lemma~\ref{lem:whp_valid}.

\begin{proofof}{Lemma~\ref{lem:whp_valid}}
Recall that 
\begin{align*}
Q(\eta)=& \frac{1}{2}\ns{\bt\wtr^* -w^*}+\frac{1}{2}\ns{\bt(\xtr)^\dagger\xitr}\\
&+\inner{\bt\wtr^* -w^*}{ \bt(\xtr)^\dagger\xitr}\\
=& G(\eta)+\inner{\bt\wtr^* -w^*}{ \bt(\xtr)^\dagger\xitr}
\end{align*}
We first show that with probability at least $1-\expd,$ there exist $\eta_1,\eta_2,\eta_3=\Theta(1/t)$ with $\eta_1<\eta_2<\eta_3$ such that $G(\eta_2)\leq 1/2\ns{w^*}-5C/4$ and $G(\eta)\geq 1/2\ns{w^*}-C/4$ for all $\eta\in[0,\eta_1]\cup [\eta_3,1/L]$.

According to Lemma~\ref{lem:isotropic}, we know with probability at least $1-\expd,$ $\sqrt{d}/\sqrt{L} \leq \sigma_i(\xtr)\leq  \sqrt{Ld}$ and $1/L \leq \lambda_i(\htr) \leq L$ for all $i\in[n]$ with $L=100.$

\paragraph{Upper bounding $G(\eta_2)$:} We can expand $G(\eta)$ as follows:
\begin{align*}
G(\eta):=& \frac{1}{2}\ns{B_{t,\eta}\wtr^*-w^*}+\frac{1}{2}\ns{B_{t,\eta}(\xtr)^\dagger\xitr}\\
=& \frac{1}{2}\ns{w^*}+ \frac{1}{2}\ns{B_{t,\eta}\wtr^*}+\frac{1}{2}\ns{B_{t,\eta}(\xtr)^\dagger\xitr}- \inner{B_{t,\eta}\wtr^*}{w^*}.
\end{align*}
Recall that $\bt=I-(I-\eta\htr)^t$, for any vector $w$ in the span of $\htr,$
$$\n{\bt w}=\n{\pr{I-(I-\eta\htr)^t} w}\leq L\eta t \n{w}.$$
According to Lemma~\ref{lem:norm_vector}, we know with probability at least $1-\expd,$ $\n{\xitr}\leq \sqrt{d}\sigma.$
Therefore, we have 
\begin{align*}
\frac{1}{2}\ns{B_{t,\eta}\wtr^*}+\frac{1}{2}\ns{B_{t,\eta}(\xtr)^\dagger\xitr}
\leq  L^2\eta^2 t^2/2 + L^3 \eta^2 t^2 \sigma^2/2 \leq L^3 \eta^2 t^2 \sigma^2,
\end{align*}
where the second inequality uses $\sigma,L\geq 1.$
According to Lemma~\ref{lem:improvement}, for any fixed $\eta\in [0,L/t]$, with probability at least $1-\exp(-\Omega(d))$, $\inner{B_{t,\eta}\wtr^*}{ w^*} \geq \frac{\eta t}{16L}.$
Therefore, 
\begin{align*}
G(\eta)\leq \frac{1}{2}\ns{w^*} + L^3 \eta^2 t^2 \sigma^2 - \frac{\eta t}{16L}\leq \frac{1}{2}\ns{w^*}- \frac{\eta t}{32L},
\end{align*}
where the second inequality holds as long as $\eta\leq \frac{1}{32L^4\sigma^2 t}.$ Choosing $\eta_2:= \frac{1}{32L^4\sigma^2 t},$ we have 
$$G(\eta_2)\leq \frac{1}{2}\ns{w^*}- \frac{1}{1024L^5\sigma^2}=\frac{1}{2}\ns{w^*}-\frac{5C}{4},$$
where $C=\frac{1}{819.2L^5 \sigma^2}.$ Note $C$ is a constant as $\sigma,L$ are constants.

\paragraph{Lower bounding $G(\eta)$ for $\eta\in[0,\eta_1]:$} Now, we prove that there exists $\eta_1=\Theta(1/t)$ with $\eta_1 < \eta_2$ such that for any $\eta\in[0,\eta_1], G(\eta)\geq \frac{1}{2}\ns{w^*}-\frac{C}{4}.$
Recall that 
\begin{align*}
G(\eta)
=& \frac{1}{2}\ns{w^*}+ \frac{1}{2}\ns{B_{t,\eta}\wtr^*}+\frac{1}{2}\ns{B_{t,\eta}(\xtr)^\dagger\xitr}- \inner{B_{t,\eta}\wtr^*}{w^*}.\\
\geq& \frac{1}{2}\ns{w^*}- \inner{B_{t,\eta}\wtr^*}{w^*}.
\end{align*}
Since $\absr{\inner{B_{t,\eta}\wtr^*}{ w^*}}\leq L\eta t,$ we know for any $\eta\in[0,\eta_1],$
$$G(\eta) \geq \frac{1}{2}\ns{w^*}- L\eta_1 t.$$
Choosing $\eta_1= \frac{C}{4L t},$ we have for any $\eta\in[0,\eta_1],$
$$G(\eta)\geq \frac{1}{2}\ns{w^*}- \frac{C}{4}.$$

\paragraph{Lower bounding $G(\eta)$ for $\eta\in [\eta_3,1/L]$:} Now, we prove that there exists $\eta_3 = \Theta(1/t)$ with $\eta_3 > \eta_2$ such that for all $\eta\in[\eta_3,1/L]$,
$$G(\eta)\geq \frac{1}{2}\ns{w^*}- \frac{C}{4}.$$
Recall that 
\begin{align*}
G(\eta)
= \frac{1}{2}\ns{B_{t,\eta}\wtr^*-w^*}+\frac{1}{2}\ns{B_{t,\eta}(\xtr)^\dagger\xitr}
\geq \frac{1}{2}\ns{B_{t,\eta}(\xtr)^\dagger\xitr}.
\end{align*}
According to Lemma~\ref{lem:norm_vector}, we know with probability at least $1-\expd,$ $\frac{\sqrt{d}\sigma}{2\sqrt{2}}\leq \n{\xitr}.$ Therefore, 
\begin{align*}
\ns{B_{t,\eta}(\xtr)^\dagger\xitr}
\geq \pr{1-e^{-\eta t/L}}^2\frac{\sigma^2}{8L}
\geq \frac{\sigma^2}{32L},
\end{align*}
where the last inequality assumes $\eta\geq \log(2)L/t.$ As long as $t\geq \log(2)L^2,$ we have $\log(2)L/t\leq 1/L.$
Choosing $\eta_3=\log(2)L/t,$ we know for all $\eta\in[\eta_3,1/L],$
$$G(\eta)\geq \frac{1}{2}\ns{B_{t,\eta}(\xtr)^\dagger\xitr}\geq \frac{\sigma^2}{64L}.$$
Note that $\frac{1}{2}\ns{w^*}= 1/2.$ Therefore, as long as $\sigma\geq 8\sqrt{L},$ we have
$$G(\eta)\geq \frac{1}{2}\ns{w^*}$$
for all $\eta\in[\eta_3,1/L].$

Overall, we have shown that there exist $\eta_1,\eta_2,\eta_3=\Theta(1/t)$ with $\eta_1<\eta_2<\eta_3$ such that $G(\eta_2)\leq 1/2\ns{w^*}-5C/4$ and $G(\eta)\geq 1/2\ns{w^*}-C/4$ for all $\eta\in[0,\eta_1]\cup [\eta_3,1/L]$. Recall that $Q(\eta) = G(\eta)+\inner{B_{t,\eta}\wtr^*-w^*}{B_{t,\eta}(\xtr)^\dagger\xitr}.$ Choosing $\epsilon=C/4$ in Lemma~\ref{lem:crossing_term2}, we know with probability at least $1-\expd,$ $\absr{\inner{B_{t,\eta}\wtr^*-w^*}{B_{t,\eta}(\xtr)^\dagger\xitr}}\leq C/4$ for all $\eta\in[0,1/L].$ Therefore, we know $Q(\eta_2)\leq 1/2\ns{w^*}-C$ and $Q(\eta)\geq 1/2\ns{w^*}-C/2$ for all $\eta\in[0,\eta_1]\cup [\eta_3,1/L]$.
\end{proofof}

Next, we give the proof of Lemma~\ref{lem:expectation_small}.

\begin{proofof}{Lemma~\ref{lem:expectation_small}}
Recall that $\bfva(\eta)=\E1/2\ns{\wt - w^*}+\frac{\sigma^2}{2}.$
For convenience, denote $Q(\eta):=1/2\ns{\wt - w^*}.$
In order to prove Lemma~\ref{lem:expectation_small}, we only need to show that $\E Q(\eta_2)\leq \frac{1}{2}\ns{w^*}-\frac{9}{10}C$ and $\E Q(\eta)\geq \frac{1}{2}\ns{w^*}-\frac{6}{10}C$ for all $\eta\in[0,\eta_1]\cup [\eta_3,1/L]$.

According to Lemma~\ref{lem:whp_valid}, as long as $\sigma$ is a large enough constant $c_1$ and $t$ is at least certain constant $c_2,$ with probability at least $1-\expd$ over the sampling of $\str,$ there exists $\eta_1,\eta_2,\eta_3=\Theta(1/t)$ with $\eta_1<\eta_2<\eta_3$ such that 
\begin{align*}
Q(\eta_2):=1/2\ns{w_{t,\eta_2} - w^*}&\leq \frac{1}{2}\ns{w^*}-C\\
Q(\eta):=1/2\ns{\wt - w^*}&\geq \frac{1}{2}\ns{w^*}-\frac{C}{2},\forall \eta\in[0,\eta_1]\cup [\eta_3,1/L] 
\end{align*}
where $C$ is a positive constant. Call this event $\calE.$ Suppose the probability that $\calE$ happens is $1-\delta$. We can write $\Estr Q(\eta)$ as follows,
\begin{align*}
\Estr Q(\eta) = \Estr[Q(\eta)|\calE]\Pr[\calE]+\Estr[Q(\eta)|\bar{\calE}]\Pr[\bar{\calE}].
\end{align*}

According to the algorithm, we know $\n{\wt}$ is always bounded by $4\sqrt{L}\sigma.$ Therefore, $Q(\eta):=1/2\ns{\wt-w^*}\leq 13L\sigma^2.$
When $\eta=\eta_2,$ we have
\begin{align*}
\Estr Q(\eta_2) \leq& \pr{\frac{1}{2}\ns{w^*}-C}(1-\delta)+13L \sigma^2\delta\\
=& \frac{1}{2}\ns{w^*}-\frac{\delta}{2}-C+(C+13L\sigma^2)\delta\\
\leq& \frac{1}{2}\ns{w^*}-\frac{9C}{10},
\end{align*}
where the last inequality assumes $\delta\leq \frac{C}{10C+130L\sigma^2}.$

When $\eta\in[0,\eta_1]\cup [\eta_3,1/L],$ we have
\begin{align*}
\Estr Q(\eta_2) \geq& \pr{\frac{1}{2}\ns{w^*}-\frac{C}{2}}(1-\delta)-13L\sigma^2\delta\\
=& \frac{1}{2}\ns{w^*}-\frac{\delta}{2}-(1-\delta)\frac{C}{2}-13L\sigma^2\delta\\
\geq& \frac{1}{2}\ns{w^*}-\frac{C}{2}-(1/2+13L\sigma^2)\delta\\
\geq& \frac{1}{2}\ns{w^*}-\frac{6C}{10},
\end{align*}
where the last inequality holds as long as $\delta\leq \frac{C}{5C+130L\sigma^2}.$

According to Lemma~\ref{lem:whp_valid}, we know $\delta\leq \expd.$ Therefore, the conditions for $\delta$ can be satisfied as long as $d$ is larger than certain constant.
\end{proofof}

\subsubsection{Generalization for $\eta\in[0,1/L]$}\label{sec:genera_GD_valid}
In this section, we show $\hfva$ is point-wise close to $\bfva$ for all $\eta\in[0,1/L].$ Recall Lemma~\ref{lem:genera_valid_GD} as follows.

\generaValidGD*

In order to prove Lemma~\ref{lem:genera_valid_GD}, let's first show that for a fixed $\eta$ with high probability $\hfva(\eta)$ is close to $\bfva(\eta)$. Similar as in Lemma~\ref{lem:genera_fixed_eta_train_GD}, we show each $\dvak$ is $O(1)$-subexponential. We defer its proof to Section~\ref{sec:technical_valid_GD}.
\begin{lemma}\label{lem:genera_fixed_eta_valid_GD}
Suppose $\sigma$ is a constant. For any fixed $\eta\in[0,1/L]$ and any $1>\epsilon>0,$ with probability at least $1-\expepsm,$
$$\absr{\hfva(\eta)-\bfva(\eta) }\leq \epsilon.$$
\end{lemma}

Next, we show that there exists an $\epsilon$-net for $\bfva$ with size $O(1/\epsilon).$ By $\epsilon$-net, we mean there exists a finite set $N_\epsilon$ of step size such that $|\bfva(\eta)-\bfva(\eta')|\leq \epsilon$ for any $\eta\in [0,1/L]$ and $\eta'\in\arg\min_{\eta\in N_\epsilon}|\eta-\eta'|.$
We defer the proof of Lemma~\ref{lem:eps_net_expect_valid_GD} to Section~\ref{sec:technical_valid_GD}.
\begin{lemma}\label{lem:eps_net_expect_valid_GD}
Suppose $\sigma$ is a constant. For any $1>\epsilon>0,$ assume $d\geq c_4\log(1/\epsilon)$ for constant $c_4.$ There exists an $\epsilon$-net $N_\epsilon$ for $\bfva$ with $|N_\epsilon|=O(1/\epsilon).$ That means, for any $\eta\in[0,1/L],$
$$|\bfva(\eta)-\bfva(\eta')|\leq \epsilon,$$
for $\eta'\in\arg\min_{\eta\in N_\epsilon}|\eta-\eta'|.$
\end{lemma}

Next, we show that with high probability, there also exists an $\epsilon$-net for $\hfva$ with size $O(1/\epsilon).$
\begin{lemma}\label{lem:eps_net_empirical_valid_GD}
Suppose $\sigma$ is a constant. For any $1>\epsilon>0,$ assume $d\geq c_4\log(1/\epsilon)$ for constant $c_4.$ With probability at least $1-\expepsm$, there exists an $\epsilon$-net $N_\epsilon'$ for $\hfva$ with $|N_\epsilon|=O(1/\epsilon).$ That means, for any $\eta\in[0,1/L],$
$$|\hfva(\eta)-\hfva(\eta')|\leq \epsilon,$$
for $\eta'\in\arg\min_{\eta\in N_\epsilon}|\eta-\eta'|.$
\end{lemma}

Combing Lemma~\ref{lem:genera_fixed_eta_valid_GD}, Lemma~\ref{lem:eps_net_expect_valid_GD} and Lemma~\ref{lem:eps_net_empirical_valid_GD}, now we give the proof of Lemma~\ref{lem:genera_valid_GD}.

\begin{proofof}{Lemma~\ref{lem:genera_valid_GD}}
The proof is very similar as in Lemma~\ref{lem:genera_train_GD}.
By Lemma~\ref{lem:genera_fixed_eta_valid_GD}, we know with probability at least $1-\expepsm,$
$\absr{\hfva(\eta)-\bfva(\eta) }\leq \epsilon$ for any fixed $\eta.$ By Lemma~\ref{lem:eps_net_expect_valid_GD} and Lemma~\ref{lem:eps_net_empirical_valid_GD}, we know as long as $d=\Omega(\log(1/\epsilon)),$ with probability at least $1-\expepsm,$ there exists $\epsilon$-net $N_\epsilon$ and $N_\epsilon'$ for $\bfva$ and $\hfva$ respectively. Here, both of $N_\epsilon$ and $N_\epsilon'$ have size $O(1/\epsilon).$ According to the proofs of Lemma~\ref{lem:eps_net_expect_valid_GD} and Lemma~\ref{lem:eps_net_empirical_valid_GD}, it's not hard to verify that $N_\epsilon\cup N_\epsilon'$ is still an $\epsilon$-net for $\hfva$ and $\bfva$. That means, for any $\eta\in[0,1/L],$ we have
$$|\bfva(\eta)-\bfva(\eta')|,|\hfva(\eta)-\hfva(\eta')|\leq \epsilon,$$ 
for $\eta'\in\arg\min_{\eta\in N_\epsilon\cup N_\epsilon'}|\eta-\eta'|.$ 

Taking a union bound over $N_\epsilon\cup N_\epsilon',$ we have with probability at least $1-O(1/\epsilon)\expepsm,$
$$\absr{\hfva(\eta)-\bfva(\eta) }\leq \epsilon$$ for any $\eta\in N_\epsilon\cup N_\epsilon'.$

Overall, we know with probability at least $1-O(1/\epsilon)\expepsm,$ for all $\eta\in[0,1/L],$
\begin{align*}
&|\bfva(\eta)-\hfva(\eta)| \\
\leq& |\bfva(\eta)-\bfva(\eta')|+|\hfva(\eta)-\hfva(\eta')| + |\hfva(\eta')-\bfva(\eta')|\\
\leq& 3\epsilon,
\end{align*}
where $\eta'\in\arg\min_{\eta\in N_\epsilon\cup N_\epsilon'}|\eta-\eta'|.$ Changing $\epsilon$ to $\epsilon'/3$ finishes the proof.
\end{proofof}

\subsubsection{Lower bounding $\hfva$ for $\eta\in[1/L,\infty)$}\label{sec:large_valid_GD}
In this section, we prove $\hfva$ is large for any step size $\eta\geq 1/L.$ Therefore, the optimal step size $\etava$ must be smaller than $\hfva.$

\largeValidGD*

When the step size is very large (larger than $3L$), we know the GD sequence gets truncated with high probability, which immediately implies the loss is high. 
The proof of Lemma~\ref{lem:very_large_valid_GD} is deferred into Section~\ref{sec:technical_valid_GD}.

\begin{lemma}\label{lem:very_large_valid_GD}
Assume $t\geq c_2,d\geq c_4$ for some constants $c_2,c_4.$ With probability at least $1-\expm,$
$$\hfva(\eta)\geq \sigma^2,$$
for all $\eta\in[3L,\infty)$
\end{lemma}

The case for step size within $[1/L,3L]$ requires more efforts. We give the proof of Lemma~\ref{lem:middle_valid_GD} in this section later.

\begin{restatable}
{lemma}{middleValidGD}\label{lem:middle_valid_GD}
Suppose $\sigma$ is a large constant. Assume $t\geq c_2,d\geq c_4\log(t)$ for some constants $c_2,c_4.$ With probability at least $1-\expm,$
\begin{align*}
\hfva(\eta)\geq& C_4\sigma^2+\frac{1}{2}\sigma^2,
\end{align*}
for all $\eta\in[1/L,3L],$ where $C_4$ is a positive constant independent with $\sigma.$
\end{restatable}

With the above two lemmas, Lemma~\ref{lem:large_valid_GD} is just a combination of them.

\begin{proofof}{Lemma~\ref{lem:large_valid_GD}}
The result follows by taking a union bound and choosing $C'=\min(C_4,1/2).$
\end{proofof}

In the remaining of this section, we give the proof of Lemma~\ref{lem:middle_valid_GD}. 
When the step size is between $1/L$ and $3L$, if the GD sequence has a reasonable probability of diverging, we can still show the loss is high similar as before. If not, we need to show the GD sequence overfits the noise in the training set, which incurs a high loss. 

Recall that the noise term is roughly $\frac{1}{2}\ns{(I-(I-\eta\htr)^t)(\xtr)^\dagger \xitr}$.
When $\eta\in[1/L,3L],$ the eigenvalues of $I-\eta\htr$ in $\str$ subspace can be negative. If all the non-zero $n$ eigenvalues of $\htr$ have the same value, there exists a step size such that the eigenvalues of $I-\eta\htr$ in subspace $\str$ is $-1.$ If $t$ is even, the eigenvalues of $I-(I-\eta\htr)^t$ in $\str$ subspace are zero, which means GD sequence does not catch any noise in $\str.$

Notice that the above problematic case cannot happen when the eigenvalues of $\htr$ are spread out. Basically, when there are two different eigenvalues, there won't exist any large $\eta$ that can cancel both directions at the same time. In Lemma~\ref{lem:eigen_diverse}, we show with constant probability, the eigenvalues of $\htr$ are indeed spread out. The proof is deferred into Section~\ref{sec:technical_valid_GD}.

\begin{lemma}\label{lem:eigen_diverse}
Let the top $n$ eigenvalues of $\htr$ be $\lambda_1\geq \cdots\geq \lambda_n$. Assume dimension $d\geq c_4$ for certain constant $c_4.$ There exist positive constants $\mu,\mu',\mu''$ such that with probability at least $\mu,$ 
$$\lambda_{\mu' n}-\lambda_{n-\mu' n+1}\geq \mu''.$$
\end{lemma}

Next, we utilize this variance in eigenvalues to prove that the GD sequence has to learn a constant fraction of the noise in training set.

\begin{lemma}\label{lem:exploit_eigen_diverse}
Suppose noise level $\sigma$ is a large enough constant $c_1$. Assume unroll length $t\geq c_2$ and dimension $d\geq c_4$ for some constants $c_2,c_4.$ Then, with probability at least $C_1$
\begin{align*}
\ns{\bt\wtr-w^*}_{\htr} \geq C_2\sigma^2,
\end{align*}
for all $\eta\in[1/L,3L],$ where $C_1,C_2$ are positive constants.
\end{lemma}
 
\begin{proofof}{Lemma~\ref{lem:exploit_eigen_diverse}}
Let $\calE_1$ be the event that $\sqrt{d}/\sqrt{L} \leq \sigma_i(\xtr)\leq  \sqrt{Ld}$ and $1/L \leq \lambda_i(\htr) \leq L$ for all $i\in[n]$ and $\sqrt{d}\sigma/4 \leq \n{\xitr}\leq \sqrt{d}\sigma$. Let $\calE_3$ be the event that $\sqrt{d}/\sqrt{L} \leq \sigma_i(\xva)\leq  \sqrt{Ld}$ and $1/L \leq \lambda_i(\hva) \leq L$ for all $i\in[n]$ and $\sqrt{d}\sigma/4 \leq \n{\xiva}\leq \sqrt{d}\sigma.$ According to Lemma~\ref{lem:isotropic} and Lemma~\ref{lem:norm_vector}, we know both $\calE_1$ and $\calE_3$ hold with probability at least $1-\expd.$

Let the top $n$ eigenvalues of $\htr$ be $\lambda_1\geq \cdots\geq \lambda_n$.
According to Lemma~\ref{lem:eigen_diverse}, assuming $d$ is larger than certain constant, we know there exist positive constants $\mu_1,\mu_2,\mu_3$ such that with probability at least $\mu_1,$
$\lambda_{\mu_2 n}-\lambda_{n-\mu_2 n+1}\geq \mu_3.$ Call this event $\calE_2.$ 

Let $S_1$ and $S_2$ be the span of the bottom and top $\mu_2 n$ eigenvectors of $\htr$ respectively. According to Lemma~\ref{lem:norm_vector}, we know $\n{\xitr}\geq \frac{\sqrt{d}}{4}\sigma$ with probability at least $1-\expd.$ Let $P_1\in\R^{n\times n}$ be a rank-$\mu_2n$ projection matrix such that the column span of $(\xtr)^\dagger P_1$ is $S_1.$ By Johnson-Lindenstrauss Lemma, we know with probability at least $1-\expd,$ $\n{\proj_{P_1}\xitr}\geq \frac{\sqrt{\mu_2}}{2}\n{\xitr}.$ Taking a union bound, with probability at least $1-\expd,$ $\n{\proj_{P_1}\xitr}\geq \frac{\sqrt{\mu_2d}\sigma}{8}.$ Similarly, we can define $P_2$ for the $S_2$ subspace and show with probability at least $1-\expd,$ $\n{\proj_{P_2}\xitr}\geq \frac{\sqrt{\mu_2d}\sigma}{8}.$ Call the intersection of both events as $\calE_4,$ which happens with with probability at least $1-\expd.$

Taking a union bound, we know $\calE_1\cap\calE_2\cap\calE_3\cap\calE_4$ holds with probability at least $\mu_1/2$ as long as $d$ is larger than certain constant. Through the proof, we assume $\calE_1\cap\calE_2\cap\calE_3\cap\calE_4$ holds.

Let's first lower bound $\n{\bt\wtr-\wtr^*}$ as follows,
\begin{align*}
\n{\bt\wtr - \wtr^*} =& \n{\bt\pr{\wtr^*+(\xtr)^\dagger\xitr} - \wtr^*} \\
\geq& \pr{\n{\bt\pr{\wtr^*+(\xtr)^\dagger\xitr}}-1}
\end{align*}

Recall that we define $S_1$ and $S_2$ as the span of the bottom and top $\mu_2 n$ eigenvectors of $\htr$ respectively. We rely on $S_1$ to lower bound $\n{\wt - w^*}$ when $\eta$ is small and rely on $S_2$ when $\eta$ is large.

\paragraph{Case 1:} Let $\sigma_{\min}^{S_1}(\bt)$ be the smallest singular value of $\bt$ within $S_1$ subspace. 
If $\eta \lambda_{n-\mu_2n+1}\leq 2-\mu_3/(2L),$ we have
$$\sigma_{\min}^{S_1}(\bt) \geq \min\pr{1-\pr{1-\frac{1}{L^2}}^t,1-\pr{1-\frac{\mu_3}{2L}}^t}\geq \frac{1}{2},$$
where the second inequality assumes $t\geq \max(L^2,2L/\mu_3)\log 2.$  Then, we have 
\begin{align*}
\n{\wt-w^*}
\geq& \pr{\sigma_{\min}^{S_1}(\bt)\pr{\n{\proj_{S_1}(\xtr)^\dagger\xitr }-1}-1}\\
\geq& \pr{\frac{1}{2}\pr{\frac{\sqrt{\mu_2}\sigma}{8\sqrt{L}}-1}-1}
\geq \frac{\sqrt{\mu_2}\sigma}{32\sqrt{L}},
\end{align*}
where the second inequality uses $\n{\proj_{P_1}\xitr}\geq \frac{\sqrt{\mu_2d}\sigma}{8}$ and the last inequality assumes $\sigma\geq \frac{48\sqrt{L}}{\sqrt{\mu_2}}.$

\paragraph{Case 2:} If $\eta \lambda_{n-\mu_2n+1}> 2-\mu_3/(2L),$ we have $\eta\lambda_{\mu_2 n}\geq 2+\mu_3/(2L)$ since $\lambda_{\mu_2 n}-\lambda_{n-\mu_2n+1}\geq \mu_3$ and $\eta\geq 1/L.$ Let $\sigma_{\min}^{S_2}(\bt)$ be the smallest singular value of $\bt$ within $S_2$ subspace. We have
$$\sigma_{\min}^{S_2}(\bt) \geq \pr{\pr{1+\frac{\mu_3}{2L}}^t-1}\geq \frac{1}{2},$$
where the last inequality assumes $t\geq 4L/\mu_3.$ Then, similar as in Case 1, we can also prove $\n{\wt-w^*}\geq \frac{\sqrt{\mu_2}\sigma}{32\sqrt{L}}.$ 

Therefore, we have 
\begin{align*}
\ns{\bt\wtr-w^*}_{\htr} = \ns{\bt\wtr-\wtr^*}_{\htr}\geq \frac{1}{L}\ns{\bt\wtr-\wtr^*}\geq \frac{\mu_2\sigma^2}{1024L^2},
\end{align*}
for all $\eta\in[1/L,3L].$ We denote $C_1:=\mu_1/2$ and $C_2=\frac{\mu_2}{1024L^2}.$
\end{proofof}

Before we present the proof of Lemma~\ref{lem:middle_valid_GD}, we still need a technical lemma that shows the noise in $\sva$ concentrates at its mean. The proof of Lemma~\ref{lem:cross_valid_GD} is deferred into Section~\ref{sec:technical_valid_GD}.

\begin{lemma}\label{lem:cross_valid_GD}
Suppose $\sigma$ is constant. For any $1>\epsilon>0,$ with probability at least $1-O(t/\epsilon)\expepsd$, $\lambda_n(\hva)\geq 1/L$ and 
\begin{align*}
\ns{\wt-\wva}_{\hva}\geq \ns{\wt-w^*}_{\hva}+(1-\epsilon)\sigma^2,
\end{align*}
for all $\eta\in[1/L,3L].$
\end{lemma}

Combing the above lemmas, we give the proof of Lemma~\ref{lem:middle_valid_GD}.

\begin{proofof}{Lemma~\ref{lem:middle_valid_GD}}
According to Lemma~\ref{lem:cross_valid_GD}, we know given $1>\epsilon>0$, with probability at least \\$1-O(t/\epsilon)\expepsd$, $\lambda_n(\hva)\geq 1/L$ and $\ns{\wt-\wva}_{\hva}\geq \ns{\wt-w^*}_{\hva}+(1-\epsilon)\sigma^2$
for all $\eta\in[1/L,3L].$ Call this event $\calE_1$. Suppose $\Pr[\calE_1]\geq 1-\delta/2,$ where $\delta$ will be specifies later. For each training set $\strk,$ we also define $\calE_1^{(k)}.$ By concentration, we know with probability at least $1-\exp(-\Omega(\delta^2 m)),$ $1/m\sum_{k=1}^m\indic{\calE_1^{(k)}}\geq 1-\delta.$

According to Lemma~\ref{lem:exploit_eigen_diverse}, we know there exist constants $C_1,C_2$ such that with probability at least $C_1,$ \\$\ns{\bt\wtr-w^*}_{\htr}\geq C_2\sigma^2$ for all $\eta\in[1/L,3L].$ Call this event $\calE_2.$ For each training set $\strk,$ we also define $\calE_2^{(k)}.$ By concentration, we know with probability at least $1-\exp(-\Omega(m)),$ $1/m\sum_{k=1}^m\indic{\calE_2^{(k)}}\geq C_1/2.$

For any step size $\eta\in[1/L,3L],$ we can lower bound $\hfva(\eta)$ as follows,
\begin{align*}
\hfva(\eta) =& \frac{1}{m}\sum_{k=1}^m \frac{1}{2}\ns{\wtk-\wvak}_{\hvak}\\
\geq& \frac{1}{m}\sum_{k=1}^m \frac{1}{2}\ns{\wtk-\wvak}_{\hvak}\indic{\calE_1^{(k)}}\\
\geq& \frac{1}{m}\sum_{k=1}^m \frac{1}{2}\ns{\wtk-w^*_k}_{\hva}\indic{\calE_1^{(k)}}+\frac{1}{2}(1-\epsilon)(1-\delta)\sigma^2\\
\geq& \frac{1}{m}\sum_{k=1}^m \frac{1}{2}\ns{\wtk-w^*_k}_{\hva}\indic{\calE_1^{(k)}\cap \calE_2^{(k)}}+\frac{1}{2}(1-\epsilon)(1-\delta)\sigma^2.
\end{align*}
As long as $\delta\leq C_1/4,$ we know $\frac{1}{m}\sum_{k=1}^m \indic{\calE_1^{(k)}\cap \calE_2^{(k)}}\geq C_1/4.$ Let $\bar{\calE}_3(\eta)$ be the event that $\wtk$ gets truncated with step size $\eta.$ We have 
\begin{align*}
&\frac{1}{m}\sum_{k=1}^m \frac{1}{2}\ns{\wtk-w^*_k}_{\hva}\indic{\calE_1^{(k)}\cap \calE_2^{(k)}}\\
=& \frac{1}{m}\sum_{k=1}^m \frac{1}{2}\ns{\wtk-w^*_k}_{\hva}\indic{\calE_1^{(k)}\cap \calE_2^{(k)}\cap \calE_3^{(k)}}
\\&+ \frac{1}{m}\sum_{k=1}^m \frac{1}{2}\ns{\wtk-w^*_k}_{\hva}\indic{\calE_1^{(k)}\cap \calE_2^{(k)}\cap \bar{\calE}_3^{(k)}}.
\end{align*}

If $\frac{1}{m}\sum_{k=1}^m\indic{\calE_1^{(k)}\cap \calE_2^{(k)}\cap \bar{\calE}_3^{(k)}}\geq C_1/8,$ we have 
\begin{align*}
\frac{1}{m}\sum_{k=1}^m \frac{1}{2}\ns{\wtk-w^*_k}_{\hva}\indic{\calE_1^{(k)}\cap \calE_2^{(k)}}
\geq& \frac{1}{m}\sum_{k=1}^m \frac{1}{2}\ns{\wtk-w^*_k}_{\hva}\indic{\calE_1^{(k)}\cap \calE_2^{(k)}\cap \bar{\calE}_3^{(k)}}\\
\geq& \frac{C_1}{8}\times \frac{9\sigma^2}{2} = \frac{9C_1 \sigma^2}{16}.
\end{align*}
Here, we lower bound $\ns{\wtk-w^*_k}_{\hva}$ by $9\sigma^2$ when the sequence gets truncated. 

If $\frac{1}{m}\sum_{k=1}^m\indic{\calE_1^{(k)}\cap \calE_2^{(k)}\cap \bar{\calE}_3^{(k)}}< C_1/8,$ we know $\frac{1}{m}\sum_{k=1}^m\indic{\calE_1^{(k)}\cap \calE_2^{(k)}\cap \calE_3^{(k)}}\geq C_1/8$. Then, we have 
\begin{align*}
\frac{1}{m}\sum_{k=1}^m \frac{1}{2}\ns{\wtk-w^*_k}_{\hva}\indic{\calE_1^{(k)}\cap \calE_2^{(k)}}
\geq& \frac{1}{m}\sum_{k=1}^m \frac{1}{2}\ns{\btk\wtr-w^*_k}_{\hva}\indic{\calE_1^{(k)}\cap \calE_2^{(k)}\cap \calE_3^{(k)}}\\
\geq& \frac{C_1}{8}\times \frac{C_2\sigma^2}{2}=\frac{C_1 C_2\sigma^2}{16}
\end{align*}
Letting $C_3=\min(\frac{9C_1}{16},\frac{C_1 C_2}{16}),$ we then have
\begin{align*}
\hfva(\eta)\geq C_3\sigma^2 + \frac{1}{2}(1-\epsilon)(1-\delta)\sigma^2\geq \frac{C_3\sigma^2}{2}+\frac{1}{2}\sigma^2,
\end{align*}
where the last inequality chooses $\delta=\epsilon=C_3/2.$ In order for $\Pr[\calE_1]\geq 1-\delta/2,$ we only need $d\geq c_4\log(t)$ for some constant $c_4.$ Replacing $C_3/2$ by $C_4$ finishes the proof.
\end{proofof}

\subsubsection{Proofs of Technical Lemmas}\label{sec:technical_valid_GD}
\begin{proofof}{Lemma~\ref{lem:crossing_term2}}
We first show that for a fixed $\eta\in[0,1/L],$ the crossing term $\absr{\inner{B_{t,\eta}\wtr^*-w^*}{ B_{t,\eta}(\xtr)^\dagger\xitr}}$ is small with high probability. We can write down the crossing term as follows:
\begin{align*}
\inner{B_{t,\eta}\wtr^*-w^*}{B_{t,\eta}(\xtr)^\dagger\xitr} = \inner{[(\xtr)^\dagger]^\top B_{t,\eta} (B_{t,\eta}\wtr^*-w^*)}{\xitr}.
\end{align*}
Noticing that $\xitr$ is independent with $[(\xtr)^\dagger]^\top B_{t,\eta} (B_{t,\eta}\wtr^*-w^*)$, we will use Hoeffding's inequality to bound $\absr{\inner{B_{t,\eta}\wtr^*-w^*}{ B_{t,\eta}(\xtr)^\dagger\xitr}}$. According to Lemma~\ref{lem:isotropic}, we know with probability at least $1-\expd,$ $\sqrt{d}/\sqrt{L} \leq \sigma_i(\xtr)\leq \sqrt{Ld}$ and $1/L \leq \lambda_i(\htr)\leq L$ for all $i\in[n]$ with $L=100.$
Since $\eta\leq 1/L,$ we know $\n{\bt}=\n{I-(I-\eta\htr)^t}\leq 1.$ Therefore, we have 
\begin{align*}
\n{[(\xtr)^\dagger]^\top B_{t,\eta} (B_{t,\eta}\wtr^*-w^*)}\leq \frac{2\sqrt{L}}{\sqrt{d}},
\end{align*}
for any $\eta\in[0,1/L].$ Then, for any $\epsilon>0,$ by Hoeffding's inequality, with probability at least $1-\expepsd,$
$$\absr{\inner{B_{t,\eta}\wtr^*-w^*}{ B_{t,\eta}(\xtr)^\dagger\xitr}}\leq \epsilon.$$

Next, we construct an $\epsilon$-net on $\eta$ and show the crossing term is small for all $\eta\in[0,1/L].$ Let 
$$g(\eta):=\inner{B_{t,\eta}\wtr^*-w^*}{ B_{t,\eta}(\xtr)^\dagger\xitr}.$$ We compute the derivative of $g(\eta)$ as follows:
\begin{align*}
g'(\eta)=&\inner{t\htr(I-\eta\htr)^{t-1}\wtr^*}{ B_{t,\eta}(\xtr)^\dagger\xitr}\\
&+\inner{B_{t,\eta}\wtr^*-w^*}{t \htr(I-\eta\htr)^{t-1}(\xtr)^\dagger\xitr}
\end{align*}
By Lemma~\ref{lem:norm_vector}, we know with probability at least $1-\expd,$ $\n{\xitr}\leq \sqrt{d}\sigma.$
Therefore,
$$|g'(\eta)|\leq L^{1.5} t\pr{1-\frac{\eta}{L}}^{t-1}\sigma+ 2L^{1.5} t\pr{1-\frac{\eta}{L}}^{t-1}\sigma=3L^{1.5} t\pr{1-\frac{\eta}{L}}^{t-1}\sigma.$$
We can control $|g'(\eta)|$ in different regimes:
\begin{itemize}
\item For $\eta\in[0, \frac{L}{t-1}],$ we have $|g'(\eta)|\leq 3L^{1.5} t\sigma.$
\item Given any $1\leq i\leq \log t-1,$ for any $\eta\in(\frac{iL}{t-1},\frac{(i+1)L}{t-1}],$ we have $|g'(\eta)|\leq \frac{3L^{1.5} t\sigma}{e^i}.$
\item For any $\eta\in(\frac{L\log t}{t-1},1/L],$ we have $|g'(\eta)|\leq 3L^{1.5} \sigma.$
\end{itemize}
Fix any $\epsilon>0,$ we know there exists an $\epsilon$-net $N_\epsilon$ with size 
\begin{align*}
|N_\epsilon|=&\frac{1}{\epsilon}\pr{\frac{L}{t-1}\sum_{i=0}^{\log t-1}\frac{3L^{1.5}t \sigma}{e^i}+\pr{\frac{1}{L}-\frac{L\log t}{t-1}}3L^{1.5} \sigma}\\
\leq& \frac{1}{\epsilon}\pr{\frac{3eL^{2.5}t \sigma}{t-1}+3\sqrt{L}\sigma}=O(\frac{1}{\epsilon})
\end{align*}
such that for any $\eta\in[0,1/L],$ there exists $\eta'\in N_\epsilon$ with $|g(\eta)-g(\eta')|\leq \epsilon.$ Note that $L=100$ and $\sigma$ is a constant. Taking a union bound over $N_\epsilon$ and all the other bad events, we have with probability at least $1-\expd-O(1/\epsilon)\expepsd,$ for all $\eta\in[0,1/L],$
$$\absr{ \inner{B_{t,\eta}\wtr^*-w^*}{ B_{t,\eta}(\xtr)^\dagger\xitr}}\leq \epsilon+\epsilon=2\epsilon.$$ As long as $1>\epsilon>0$, this happens with probability at least $1-O(1/\epsilon)\expepsd.$ 
Replacing $\epsilon$ by $\epsilon'/2$ finishes the proof.
\end{proofof}

\begin{proofof}{Lemma~\ref{lem:improvement}}
According to Lemma~\ref{lem:isotropic}, we know with probability at least $1-\exp(-\Omega(d)),$ $1/L \leq \lambda_i(\htr)\leq L$ for all $i\in[n]$ with $L=100.$
We can lower bound $\inner{B_{t,\eta}\wtr^*}{w^*}$ as follows,
\begin{align*}
\inner{B_{t,\eta}\wtr^*}{w^*} =& \inner{\pr{I-(I-\eta\htr)^t}\wtr^*}{\wtr^*}\notag\\
\geq& \lambda_{\min}\pr{I-(I-\eta\htr)^t}\ns{\wtr^*}\notag\\
\geq& \pr{1-\exp\pr{-\frac{\eta t}{L}}}\ns{\wtr^*}.
\end{align*}
By Johnson-Lindenstrauss lemma (Lemma~\ref{lem:JL_subspace}), we know with probability at least $1-2\exp(-c\epsilon^2d/4),$ 
\begin{align*}
\n{\wtr^*}\geq \frac{1}{2}(1-\epsilon)\n{w^*}=\frac{1}{2}(1-\epsilon).
\end{align*}
Then, we know with probability at least $1-2\exp(-c\epsilon^2d/4)-\expd,$ 
\begin{align*}
\inner{B_{t,\eta}\wtr^*}{ w^*}
\geq& \pr{1-\exp\pr{-\frac{\eta t}{L}}}\ns{\wtr^*}\\
\geq& \pr{1-\exp\pr{-\frac{\eta t}{L}}}\frac{1}{4}(1-\epsilon)^2\\
\geq& \frac{1-2\epsilon}{4}\pr{1-\exp\pr{-\frac{\eta t}{L}}}
\end{align*}
Since $e^x\leq 1-x+x^2/2$ for any $x\leq 0,$ we know $\exp(-\eta t/L)\leq 1-\eta t/L+ \eta^2 t^2/(2L^2).$ For any $\eta\leq L/t,$ we have $\exp(-\eta t/L)\leq 1-\eta t/(2L).$ Then with probability at least $1-2\exp(-c\epsilon^2d/4)-\expd,$
\begin{align*}
\inner{B_{t,\eta}\wtr^*}{w^*}
\geq& \frac{1-2\epsilon}{4}\frac{\eta t}{2L}\\
\geq& \frac{\eta t}{16L},
\end{align*}
where the second inequality holds by choosing  $\epsilon= 1/4.$
\end{proofof}

\begin{proofof}{Lemma~\ref{lem:genera_fixed_eta_valid_GD}}
Recall that 
$$\hfva(\eta):=\frac{1}{m}\sum_{k=1}^m \dvak$$  
For each individual loss function $\dvak,$ we have
\begin{align*}
\dvak
=& \frac{1}{2}\ns{\wtk - w^*-(\xvak)^\dagger\xivak}_{\hvak}\\
=& \frac{1}{2}\ns{\wtk - w^*}_{\hvak}+\frac{1}{2n}\ns{\xivak}+\inner{\wtk - w^*}{\frac{1}{n}(\xvak)^\top \xivak}\\
\leq& \frac{25L\sigma^2}{2}\n{\hvak} + \frac{1}{2n}\ns{\xivak} + 5\sqrt{L}\sigma\pr{\frac{1}{\sqrt{n}}\n{\xvak}}\pr{\frac{1}{\sqrt{n}}\n{\xivak}}
\end{align*}

We can write $\n{\hvak}$ as $\sigma_{\max}^2(\frac{1}{\sqrt{n}} \xvak).$ According to Lemma~\ref{lem:concen_gauss}, we know $\sigma_{\max}(\xvak)-\E\sigma_{\max}(\xvak)$ is $O(1)$-subgaussian, which implies that $\sigma_{\max}(\frac{1}{\sqrt{n}} \xvak)-\E \sigma_{\max}(\frac{1}{\sqrt{n}} \xvak)$ is $O(1/\sqrt{d})$-subgaussian. Since $\E \sigma_{\max}(\frac{1}{\sqrt{n}} \xvak)$ is a constant, we know $\sigma_{\max}(\frac{1}{\sqrt{n}} \xvak)$ is $O(1)$-subgaussian and $\sigma_{\max}^2(\frac{1}{\sqrt{n}} \xvak)$ is $O(1)$-subexponential. Similarly, we know both $\frac{1}{2n}\ns{\xivak}$ and $\pr{\frac{1}{\sqrt{n}}\n{\xvak}}\pr{\frac{1}{\sqrt{n}}\n{\xivak}}$ are $O(1)$-subexponential. 
This further implies that $\dvak$ is $O(1)$-subexponential. Therefore, $\hfva$ is the average of $m$ i.i.d. $O(1)$-subexponential random variables. By standard concentration inequality, we know for any $1>\epsilon>0,$ with probability at least $1-\expepsm,$
$$\absr{\hfva(\eta)-\bfva(\eta) }\leq \epsilon.$$
\end{proofof}

\begin{proofof}{Lemma~\ref{lem:eps_net_expect_valid_GD}}
Recall that 
\begin{align*}
\bfva(\eta)=&\E\frac{1}{2}\ns{\wt-w^*}+\sigma^2/2.
\end{align*}
We only need to construct an $\epsilon$-net for $\E\frac{1}{2}\ns{\wt-w^*}$.
Let $\calE$ be the event that $\sqrt{d}/\sqrt{L} \leq \sigma_i(\xtr)\leq  \sqrt{Ld}$ and $1/L \leq \lambda_i(\htr) \leq L$ for all $i\in[n]$ and $\n{\xitr}\leq \sqrt{d}\sigma$. We have
$$\E\frac{1}{2}\ns{\wt-w^*} = \E\br{\frac{1}{2}\ns{\wt-w^*}| \calE}\Pr[\calE]+ \E\br{\frac{1}{2}\ns{\wt-w^*}| \bar{\calE} }\Pr[\bar{\calE}]$$ 

We first construct an $\epsilon$-net for $\E\br{\frac{1}{2}\ns{\wt-w^*}| \calE}\Pr[\calE]$. Let $Q(\eta):=\frac{1}{2}\ns{\wt-w^*}.$ Fix a training set $\str$ under which event $\calE$ holds. We show that $Q(\eta)$ has desirable lipschitz property.

The derivative of $Q(\eta)$ can be computed as follows,
$$Q'(\eta)=\inner{t\htr(I-\eta\htr)^{t-1}\wtr}{\wt-w^*}.$$
Conditioning on $\calE,$ we have
$$|Q'(\eta)|= O(1)t(1-\frac{\eta}{L})^{t-1}.$$
Therefore, we have 
\begin{align*}
\absr{\frac{\partial}{\partial \eta}\E\br{\frac{1}{2}\ns{\wt-w^*}| \calE}\Pr[\calE]}= O(1)t(1-\frac{\eta}{L})^{t-1}. 
\end{align*}
Similar as in Lemma~\ref{lem:crossing_term2}, for any $\epsilon>0,$ we know there exists an $\epsilon$-net $N_\epsilon$ with size $O(1/\epsilon)$
such that for any $\eta\in[0,1/L],$ 
$$\absr{\E\br{\frac{1}{2}\ns{\wt-w^*}| \calE}\Pr[\calE]-\E\br{\frac{1}{2}\ns{w_{t,\eta'}-w^*}| \calE}\Pr[\calE]}\leq \epsilon$$
for $\eta'\in\arg\min_{\eta\in N_\epsilon}|\eta-\eta'|.$ 

Suppose the probability of $\bar{\calE}$ is $\delta.$ We have 
\begin{align*}
\E\br{\frac{1}{2}\ns{\wt-w^*}| \bar{\calE} }\Pr[\bar{\calE}]
\leq \frac{25L\sigma^2}{2} \delta\leq \epsilon,
\end{align*}
where the last inequality assumes $\delta\leq \frac{2\epsilon}{25L\sigma^2}.$ According to Lemma~\ref{lem:isotropic} and Lemma~\ref{lem:norm_vector}, we know $\delta:=\Pr[\bar{\calE}]\leq \expd.$ Therefore, given any $\epsilon>0,$ there exists constant $c_4$ such that $\delta\leq \frac{2\epsilon}{25L\sigma^2}$ as long as $d\geq c_4\log(1/\epsilon).$

Overall, for any $\epsilon>0,$ as long as $d=\Omega(\log(1/\epsilon)),$ there exists $N_\epsilon$ with size $O(1/\epsilon)$ such that for any $\eta\in[0,1/L],$ $|\bfva(\eta)-\bfva(\eta')|\leq 3\epsilon$ for $\eta'\in\arg\min_{\eta\in N_\epsilon}|\eta-\eta'|.$ Changing $\epsilon$ to $\epsilon'/3$ finishes the proof.
\end{proofof}

\begin{proofof}{Lemma~\ref{lem:eps_net_empirical_valid_GD}}
For each $k\in[m],$ let $\calE_k$ be the event that $\sqrt{d}/\sqrt{L}\leq \sigma_i(\xtrk)\leq \sqrt{L d}$ for any $i\in[n]$ and $\n{\xitrk}\leq \sqrt{d}\sigma$. Then, we can write the empirical meta objective as follows,
$$\hfva(\eta):=\frac{1}{m}\sum_{k=1}^m \dtrk\indick +\frac{1}{m}\sum_{k=1}^m \dtrk\indicbk.$$

Similar as Lemma~\ref{lem:eps_net_expect_valid_GD}, we will show that the first term has desirable Lipschitz property and the second term is small. 
Now, let's focus on the first term $\frac{1}{m}\sum_{k=1}^m \dtrk\indick$. Recall that 
\begin{align*}
\dtrk=&\frac{1}{2}\ns{\wtk-\wvak}_{\hvak}\\
=&\frac{1}{2}\ns{\btk\wtrk - w^*-(\xvak)^\dagger\xivak }_{\hvak}.
\end{align*}
Computing the derivative of $\dtrk$ in terms of $\eta,$ we have
\begin{align*}
\frac{\partial }{\partial \eta}\dtrk=\inner{t\htrk(I-\eta\htrk)^{t-1}\wtrk}{\hvak\pr{ \wtk-w^*-(\xvak)^\dagger\xivak }}
\end{align*}
Conditioning on $\calE_k,$ we can bound the derivative, 
\begin{align*}
\absr{\frac{\partial }{\partial \eta}\dtrk}= O(1)t\pr{1-\frac{\eta}{L}}^{t-1}\pr{\n{\hvak}+ \pr{\frac{1}{\sqrt{d}}\n{\xvak}}\pr{\frac{1}{\sqrt{d}}\n{\xivak}}}.
\end{align*}
Therefore, we have 
\begin{align*}
\absr{\frac{1}{m}\sum_{k=1}^m \frac{\partial }{\partial \eta}\dtrk\indick}= O(1)t\pr{1-\frac{\eta}{L}}^{t-1}\frac{1}{m}\sum_{k=1}^m\pr{\n{\hvak}+ \pr{\frac{1}{\sqrt{d}}\n{\xvak}}\pr{\frac{1}{\sqrt{d}}\n{\xivak}}}.
\end{align*}
Similar as in Lemma~\ref{lem:genera_fixed_eta_valid_GD}, we know both $\n{\hvak}$ and $\pr{\frac{1}{\sqrt{d}}\n{\xvak}}\pr{\frac{1}{\sqrt{d}}\n{\xivak}}$ are $O(1)$-subexponential. Therefore, we know with probability at least $1-\expm,$ $\frac{1}{m}\sum_{k=1}^m\pr{\n{\hvak}+ \pr{\frac{1}{\sqrt{d}}\n{\xvak}}\pr{\frac{1}{\sqrt{d}}\n{\xivak}}}=O(1).$ This further shows that with probability at least $1-\expm,$
$$\absr{\frac{1}{m}\sum_{k=1}^m \frac{\partial }{\partial \eta}\dtrk\indick}= O(1)t\pr{1-\frac{\eta}{L}}^{t-1}.$$
Similar as in Lemma~\ref{lem:crossing_term2}, we can show that for any $\epsilon>0,$ there exists an $\epsilon$-net with size $O(1/\epsilon)$ for $\frac{1}{m}\sum_{k=1}^m \dtrk\indick$.

Next, we show that the second term $\frac{1}{m}\sum_{k=1}^m \dtrk\indicbk$ is small with high probability. According to the proof in Lemma~\ref{lem:genera_fixed_eta_valid_GD}, we know 
\begin{align*}
\dtrk=O(1)\pr{\n{\hvak}+\frac{1}{d}\ns{\xivak}+\pr{\frac{1}{\sqrt{d}}\n{\xvak}}\pr{\frac{1}{\sqrt{d}}\n{\xivak}}} 
\end{align*}
Therefore, there exists constant $C$ such that 
\begin{align*}
\frac{1}{m}\sum_{k=1}^m \dtrk\indicbk\leq  C\frac{1}{m}\sum_{k=1}^m \pr{\n{\hvak}+\frac{1}{d}\ns{\xivak}+\pr{\frac{1}{\sqrt{d}}\n{\xvak}}\pr{\frac{1}{\sqrt{d}}\n{\xivak}}} \indicbk.
\end{align*}
It's not hard to verify that $\pr{\n{\hvak}+\frac{1}{d}\ns{\xivak}+\pr{\frac{1}{\sqrt{d}}\n{\xvak}}\pr{\frac{1}{\sqrt{d}}\n{\xivak}}} \indicbk$ is $O(1)$-subexponential. Suppose the expectation of $\pr{\n{\hvak}+\frac{1}{d}\ns{\xivak}+\pr{\frac{1}{\sqrt{d}}\n{\xvak}}\pr{\frac{1}{\sqrt{d}}\n{\xivak}}}$ is $\mu,$ which is a constant. Suppose the probability of $\bar{\calE_k}$ be $\delta.$ We know the expectation of $\pr{\n{\hvak}+\frac{1}{d}\ns{\xivak}+\pr{\frac{1}{\sqrt{d}}\n{\xvak}}\pr{\frac{1}{\sqrt{d}}\n{\xivak}}} \indicbk$ is $\mu\delta$ due to independence. By standard concentration inequality, for any $1>\epsilon>0,$ with probability at least $1-\expepsm,$
$$C\frac{1}{m}\sum_{k=1}^m \pr{\n{\hvak}+\frac{1}{d}\ns{\xivak}+\pr{\frac{1}{\sqrt{d}}\n{\xvak}}\pr{\frac{1}{\sqrt{d}}\n{\xivak}}} \indicbk\leq C\mu\delta+C\epsilon\leq (C+1)\epsilon,$$
where the second inequality assumes $\delta\leq \epsilon/(C\mu).$ By Lemma~\ref{lem:isotropic} and Lemma~\ref{lem:norm_vector}, we know $\delta\leq \expd.$ Therefore, as long as $d\geq c_4\log(1/\epsilon)$ for some constant $c_4$, we have $\delta\leq \epsilon/(C\mu).$

Overall, we know that as long as $d\geq c_4\log(1/\epsilon)$, with probability at least $1-\expepsm,$ there exists $N_\epsilon'$ with $|N_\epsilon'|=O(1/\epsilon)$ such that for any $\eta\in[0,1/L],$ 
$$|\hfva(\eta)-\hfva(\eta')|\leq (2C+3)\epsilon,$$
for $\eta'\in\arg\min_{\eta\in N_\epsilon}|\eta-\eta'|.$
Changing $\epsilon$ to $\epsilon'/(2C+3)$ finishes the proof.
\end{proofof}

\begin{proofof}{Lemma~\ref{lem:very_large_valid_GD}}
Let $\calE_1$ be the event that $\sqrt{d}/\sqrt{L} \leq \sigma_i(\xtr)\leq  \sqrt{Ld}$ and $1/L \leq \lambda_i(\htr) \leq L$ for all $i\in[n]$ and $\sqrt{d}\sigma/4 \leq \n{\xitr}\leq \sqrt{d}\sigma$. Let $\calE_2$ be the event that $\sqrt{d}/\sqrt{L} \leq \sigma_i(\xva)\leq  \sqrt{Ld}$ and $1/L \leq \lambda_i(\hva) \leq L$ for all $i\in[n]$ and $\sqrt{d}\sigma/4 \leq \n{\xiva}\leq \sqrt{d}\sigma.$ According to Lemma~\ref{lem:isotropic} and Lemma~\ref{lem:norm_vector}, we know both $\calE_1$ and $\calE_2$ hold with probability at least $1-\expd.$ Assuming $d\geq c_4$ for certain constant $c_4,$ we know $\Pr[\calE_1\cap \calE_2]\geq 2/3.$ Also define $\calE_1^{(k)}$ and $\calE_2^{(k)}$ on each training set $\strk.$ By concentration, we know with probability at least $1-\expm,$
$$\frac{1}{m}\sum_{k=1}^m \indic{\calE_1^{(k)}\cap \calE_2^{(k)}}\geq \frac{1}{2}.$$

It's easy to verify that conditioning on $\calE_1,$ the GD sequence always exceeds the norm threshold and gets truncated for $\eta\geq 3L$ as long as $t$ is larger than certain constant. We can lower bound $\hfva$ for any $\eta\geq 3L$ as follows,
\begin{align*}
\hfva(\eta) =& \frac{1}{m}\sum_{k=1}^m \frac{1}{2}\ns{\wtk-\wvak}_{\hvak}\\
\geq& \frac{1}{m}\sum_{k=1}^m \frac{1}{2}\ns{\wtk-\wvak}_{\hvak}\indic{\calE_1\cap \calE_2}
\geq 2\sigma^2\frac{1}{2}=\sigma^2,
\end{align*}
where the last inequality lower bounds $\ns{\wtk-\wvak}_{\hvak}$ by $2\sigma^2$ when $\wtk$ gets truncated.
\end{proofof} 

\begin{proofof}{Lemma~\ref{lem:eigen_diverse}}
We first show that with constant probability in $\xtr,$ the variance of the eigenvalues of $\htr$ is lower bounded by a constant. Let $\bar{\lambda}$ be $1/n\sum_{i=1}^n \lambda_i$. Specifically, we show $1/n\sum_{i=1}^n \lambda_i^2-\bar{\lambda}^2$ is lower bounded by a constant. 

Let's first compute the variance of the eigenvalues in expectation. Let the $i$-th row of $\xtr$ be $x_i^\top.$ We have,
\begin{align*}
\Estr\br{\bar{\lambda}^2 } 
= \frac{1}{n^2}\Estr\br{\pr{\tr\pr{\frac{1}{n}\xtr^\top\xtr}}^2}
=& \frac{1}{n^4}\Estr\br{\pr{\sum_{i=1}^n\ns{x_i}}^2}\\
=& \frac{1}{n^4}\sum_{i=1}^n \Estr \n{x_i}^4 + \frac{1}{n^4}\sum_{1\leq i\neq j\leq n}\Estr\ns{x_i}\ns{x_j}\\
=& \frac{1}{n^4}\pr{nd(d+2)+n(n-1)d^2}=\frac{d^2}{n^2}+\frac{2d}{n^3}.
\end{align*}
Similarly, we compute $\Estr \br{1/n\sum_{i=1}^n \lambda_i^2}$ as follows,
\begin{align*}
\Estr \br{\frac{1}{n}\sum_{i=1}^n \lambda_i^2}
=& \frac{1}{n^3}\Estr \br{\tr\pr{\xtr^\top\xtr\xtr^\top\xtr}}\\
=& \frac{1}{n^3}\sum_{i=1}^n \Estr \n{x_i}^4+\frac{1}{n^3}\sum_{1\leq i\neq j\leq n}\Estr \inner{x_i}{x_j}^2\\
=& \frac{1}{n^3}\pr{nd(d+2)+n(n-1)d}=\frac{d^2}{n^2}+\frac{d}{n}+\frac{d}{n^2}
\end{align*}
Therefore, we have
$$\Estr \br{\frac{1}{n}\sum_{i=1}^n \lambda_i^2-\bar{\lambda}^2}=\frac{d}{n}+\frac{d}{n^2}-\frac{2d}{n^3}\geq \frac{d}{n}\geq \frac{4}{3},$$
where the first inequality assumes $n\geq 2$ and the last inequality uses $n\leq \frac{3d}{4}.$ Since $n\geq \frac{1}{4}d,$ we know $n\geq 2$ as long as $d\geq 8.$

Let $\calE$ be the event that $\sqrt{d}/\sqrt{L} \leq \sigma_i(\xtr)\leq \sqrt{Ld}$ and $1/L \leq \lambda_i(\htr)\leq L$ for $i\in[n]$ with $L=100.$ According to Lemma~\ref{lem:isotropic}, we know $\calE$ happens with probability at least $1-\expd.$ Let $\indic{\calE}$ be the indicator function for event $\calE.$ Next we show that $\E[1/n\sum_{i=1}^n(\lambda_i-\bar{\lambda})^2\indic{\calE}]$ is also lower bounded.

It's clear that $\E\br{\bar{\lambda}^2 \indic{\calE}}$ is upper bounded by $\E\br{\bar{\lambda}^2}$. In order to lower bound $\E\br{\frac{1}{n}\sum_{i=1}^n \lambda_i^2\indic{\calE}},$ we first show that $\E\br{\frac{1}{n}\sum_{i=1}^n \lambda_i^2\indic{\bar{\calE}}}$ is small. We can decompose $\E\br{\frac{1}{n}\sum_{i=1}^n \lambda_i^2\indic{\bar{\calE}}}$ into two parts,
\begin{align*}
\E\br{\frac{1}{n}\sum_{i=1}^n \lambda_i^2\indic{\bar{\calE}}}
=& \E\br{\frac{1}{n}\sum_{i=1}^n \lambda_i^2\indic{\bar{\calE}\text{ and } \lambda_1 \leq L}}+\E\br{\frac{1}{n}\sum_{i=1}^n \lambda_i^2\indic{\lambda_1 > L}}. 
\end{align*}
The first term can be bounded by $L^2\Pr[\bar{\calE}].$ Since $\Pr[\bcalE]\leq \expd,$ we know the first term is at most $1/6$ as long as $d$ is larger than certain constant. The second term can be bounded by $\E\br{\lambda_1^2\indic{\lambda_1 > L}}.$ According to Lemma~\ref{lem:sig_matrix}, we know $\Pr[\lambda_1\geq L+t]\leq \exp(-\Omega(dt)).$ Then, it's not hard to verify that $\E\br{\lambda_1^2\indic{\lambda_1 > L}}=O(1/d)$ that is bounded by $1/6$ as long as $d$ is larger than certain constant. Overall, we know $\E\br{\frac{1}{n}\sum_{i=1}^n \lambda_i^2\indic{\calE}}\geq \E\br{\frac{1}{n}\sum_{i=1}^n \lambda_i^2}-1/3.$ Combing with the upper bounds on $\E\br{\bar{\lambda}^2 \indic{\calE}}$, we have
$\E\br{\frac{1}{n}\sum_{i=1}^n (\lambda_i-\bar{\lambda})^2\indic{\calE}}\geq 1.$

Since conditioning on $\calE,$ $\lambda_i$ is bounded by $L$ for all $i\in[n].$ In order to make $\E\br{\frac{1}{n}\sum_{i=1}^n (\lambda_i-\bar{\lambda})^2\indic{\calE}}$ lower bounded by one, there must exist positive constants $\mu_1,\mu_2$ such that with probability at least $\mu_1,$ $\calE$ holds and $\frac{1}{n}\sum_{i=1}^n (\lambda_i-\bar{\lambda})^2\geq \mu_2$.

Since $\frac{1}{n}\sum_{i=1}^n (\lambda_i-\bar{\lambda})^2\geq \mu_2$ and $\lambda_i\leq L$ for all $i\in[n],$ we know there exists a subset of eigenvalues $S\subset \{\lambda_i\}_1^n$ with size $\mu_3 n$ such that $|\lambda_i-\bar{\lambda}|\geq \mu_4$ for all $\lambda_i\in S,$ where $\mu_3,\mu_4$ are both positive constants.

If at least half of eigenvalues in $S$ are larger than $\bar{\lambda},$ we know at least $\frac{\mu_3\mu_4 n}{2L}$ number of eigenvalues are smaller than $\bar{\lambda}.$ Otherwise, the expectation of the eigenvalues will be larger than $\bar{\lambda},$ which contradicts the definition of $\bar{\lambda}.$ Similarly, if at least half of eigenvalues in $S$ are smaller than $\bar{\lambda},$ we know at least $\frac{\mu_3\mu_4 n}{2L}$ number of eigenvalues are larger than $\bar{\lambda}.$ Denote $\mu_5:=\frac{\mu_3\mu_4}{2L}.$ We know $\lambda_{\mu_5 n}-\lambda_{n-\mu_5 n+1}\geq \mu_4.$ 
\end{proofof}

\begin{proofof}{Lemma~\ref{lem:cross_valid_GD}}
Let $\calE_1$ be the event that $\sqrt{d}/\sqrt{L} \leq \sigma_i(\xtr)\leq  \sqrt{Ld}$ and $1/L \leq \lambda_i(\htr) \leq L$ for all $i\in[n]$ and $\sqrt{d}\sigma/4 \leq \n{\xitr}\leq \sqrt{d}\sigma$. Let $\calE_3$ be the event that $\sqrt{d}/\sqrt{L} \leq \sigma_i(\xva)\leq  \sqrt{Ld}$ and $1/L \leq \lambda_i(\hva) \leq L$ for all $i\in[n]$ and $\sqrt{d}\sigma/4 \leq \n{\xiva}\leq \sqrt{d}\sigma.$ According to Lemma~\ref{lem:isotropic} and Lemma~\ref{lem:norm_vector}, we know both $\calE_1$ and $\calE_3$ hold with probability at least $1-\expd.$ In this proof, we assume both properties hold and take a union bound at the end. 

We can lower bound $\ns{\wt-\wva}_{\hva}$ as follows,
\begin{align*}
\ns{\wt-\wva}_{\hva}
=& \ns{\wt-w^*-(\xva)^\dagger\xiva}_{\hva}\\
\geq& \ns{\wt-w^*}_{\hva}+\frac{1}{n}\ns{\xiva}-2\absr{\inner{\wt-w^*}{\hva(\xva)^\dagger\xiva}}.
\end{align*}

For the second term, by Lemma~\ref{lem:norm_vector}, we know for any $1>\epsilon>0,$ with probability at least $1-\expepsd,$
$$\frac{1}{n}\ns{\xiva}\geq (1-\epsilon)\sigma^2.$$

We can write down the third term as $\inner{[(\xva)^\dagger]^\top\hva(\wt-w^*)}{\xiva}$. Suppose $\sigma$ is a constant, we know $\n{[(\xva)^\dagger]^\top\hva(\wt-w^*)}=O(1/\sqrt{d}).$ Therefore, for a fixed $\eta\in[1/L,3L],$ we have with probability at least $1-\expepsd,$
$$\absr{\inner{\wt-w^*}{\hva(\xva)^\dagger\xiva}}\leq \epsilon.$$
To prove this crossing term is small for all $\eta\in[1/L,3L],$ we need to construct an $\epsilon$-net for the crossing term. Similar as in Lemma~\ref{lem:eps_net_train_empirical}, we can show there exists an $\epsilon$-net for the crossing term with size $O(t/\epsilon).$ Taking a union bound over this $\epsilon$-net, we are able to show with probability at least $1-O(t/\epsilon)\expepsd,$
$$\absr{\inner{\wt-w^*}{\hva(\xva)^\dagger\xiva}}\leq \epsilon,$$
for all $\eta\in[1/L,3L].$ 

Overall, we have with probability at least $1-O(t/\epsilon)\expepsd,$
\begin{align*}
\ns{\wt-\wva}_{\hva}
\geq& \ns{\wt-w^*}_{\hva}+\frac{1}{n}\ns{\xiva}-2\absr{\inner{\wt-w^*}{\hva(\xva)^\dagger\xiva}}\\
\geq& \ns{\wt-w^*}_{\hva}+(1-\epsilon)\sigma^2-2\epsilon\geq (1-3\epsilon)\sigma^2,
\end{align*}
for all $\eta\in[1/L,3L],$ where the last inequality uses $\sigma\geq 1.$ The proof finishes as we change $3\epsilon$ to $\epsilon'.$
\end{proofof}

\section{Proofs of train-by-train with large number of samples (GD)}\label{sec:proofs_large_sample}
In this section, we give the proof of Theorem~\ref{thm:large_sample}. We show when the size of each training set $n$ and the the number of training tasks $m$ are large enough, train-by-train also performs well. Recall Theorem~\ref{thm:large_sample} as follows.

\thmLargeSample*

In the proof, we use the same notations defined in Section~\ref{sec:proofs_GD}.
On each training task $P$, in Lemma~\ref{lem:good_point_large_sample} we show the meta-loss can be decomposed into two terms:
\begin{align*}
\dtr= \frac{1}{2}\ns{\wt-\wtr}_{\htr}+\frac{1}{2n}\ns{(I_n-\proj_{\xtr})\xitr},
\end{align*}
where $\wtr=w^*+(\xtr)^\dagger\xitr.$ Recall that $\xtr$ is a $n\times d$ matrix with its $i$-th row as $x_i^\top.$ The pseudo-inverse $(\xtr)^\dagger$ has dimension $d\times n$ satisfying $\xtr^\dagger\xtr=I_d.$ Here, $\proj_{\xtr}\in \R^{n\times n}$ is a projection matrix onto the column span of $\xtr.$

In Lemma~\ref{lem:good_point_large_sample}, we show with a constant step size, the first term in $\dtr$ is exponentially small. The second term is basically the projection of the noise on the orthogonal subspace of the data span. We show this term concentrates well on its mean. This lemma servers as step 1 in Section~\ref{sec:strategy}.
The proof of Lemma~\ref{lem:good_point_large_sample} is deferred into Section~\ref{sec:good_point_large_sample}.

\begin{restatable}
{lemma}{GoodPointLargeSample}\label{lem:good_point_large_sample}
Assume $n\geq 40d.$ Given any $1>\epsilon>0$, with probability at least $1-m\expn-\exp(-\Omega(\epsilon^4 md/n)),$
$$\hftr(2/3)\leq 20(1-\frac{1}{3})^{2t}\sigma^2+\frac{n-d}{2n}\sigma^2+\frac{\epsilon^2 d\sigma^2}{20n}.$$
\end{restatable}

In the next lemma, we show the empirical meta objective is large when $\eta$ exceeds certain threshold. We define this threshold $\heta$ such that for any step size larger than $\heta$ the GD sequence has reasonable probability being truncated. In the proof, we rely on the truncated sequences to argue the meta-objective must be high. The precise definition of $\heta$ is in Definition~\ref{def:heta}. This lemma serves as step 2 in Section~\ref{sec:strategy}.
We leave the proof of Lemma~\ref{lem:diverge_large_sample} into Section~\ref{sec:high_loss_large_sample}.

\begin{restatable}
{lemma}{DivergeLargeSample}\label{lem:diverge_large_sample}
Let $\heta$ be as defined in Definition~\ref{def:heta} with $1>\epsilon>0$. Assume $n\geq cd,t\geq c_2,d\geq c_4$ for some constants $c,c_2,c_4.$ With probability at least $1-\exp(-\Omega(\epsilon^4 md^2/n^2)),$
$$\hftr(\eta)\geq \frac{\epsilon^2 d\sigma^2}{8n}+\frac{n-d}{2n}\sigma^2-\frac{\epsilon^2 d\sigma^2}{20n},$$
for all $\eta>\heta.$
\end{restatable}

By Lemma~\ref{lem:good_point_large_sample} and Lemma~\ref{lem:diverge_large_sample}, we know when $t$ is reasonably large, $\hftr(\eta)$ is larger than $\hftr(2/3)$ for all step sizes $\eta>\heta.$ This means the optimal step size $\heta$ must lie in $[0,\heta].$ In Lemma~\ref{lem:genera_large_sample}, we show a generalization result for $\eta\in[0,\heta].$ This serves as step 3 in Section~\ref{sec:strategy}. We prove this lemma in Section~\ref{sec:genera_large_sample}.

\begin{restatable}
{lemma}{generaLargeSample}\label{lem:genera_large_sample}
Let $\heta$ be as defined in Definition~\ref{def:heta} with $1>\epsilon>0$.
Suppose $\sigma$ is a constant. Assume $n\geq c\log(\frac{n}{\epsilon d})d,t\geq c_2,d\geq c_4$ for some constants $c,c_2,c_4.$ With probability at least $1-m\expn-O(\frac{tn}{\epsilon^2 d}+m)\exp(-\Omega(m\epsilon^4 d^2/n^2)),$
$$|\bftr(\eta)-\hftr(\eta)| \leq \frac{17\epsilon^2 d\sigma^2}{n},$$
for all $\eta\in[0,\heta],$
\end{restatable}

Combining Lemma~\ref{lem:good_point_large_sample}, Lemma~\ref{lem:diverge_large_sample} and Lemma~\ref{lem:genera_large_sample}, we present the proof of Theorem~\ref{thm:large_sample} as follows.

\begin{proofof}{Theorem~\ref{thm:large_sample}}
According to Lemma~\ref{lem:good_point_large_sample}, assuming $n\geq 40d,$ given any $1/2>\epsilon>0$, with probability at least $1-m\expn-\exp(-\Omega(\epsilon^4 md/n)),$
$\hftr(2/3)\leq 20(1-\frac{1}{3})^{2t}\sigma^2+\frac{n-d}{2n}\sigma^2+\frac{\epsilon^2 d\sigma^2}{20n}.$
As long as $t\geq c_2\log(\frac{n}{\epsilon d})$ for certain constant $c_2,$ we have 
$$\hftr(2/3)\leq \frac{n-d}{2n}\sigma^2+\frac{7\epsilon^2 d\sigma^2}{100n}.$$

Let $\heta$ be as defined in Definition~\ref{def:heta} with the same $\epsilon$. According to Lemma~\ref{lem:diverge_large_sample}, as long as $n\geq cd,t\geq c_2,d\geq c_4$ with probability at least $1-\exp(-\Omega(\epsilon^4 md^2/n^2)),$
$$\hftr(\eta)\geq \frac{\epsilon^2 d\sigma^2}{8n}+\frac{n-d}{2n}\sigma^2-\frac{\epsilon^2 d\sigma^2}{20n}=\frac{n-d}{2n}\sigma^2+\frac{7.5\epsilon^2 d\sigma^2}{100n}$$
for all $\eta>\heta.$ We have $\hftr(\eta)>\hftr(2/3)$ for all $\eta\geq \heta.$ This implies that $\etatr$ is within $[0,\heta]$ and $\hftr(\etatr)\leq \hftr(2/3)\leq \frac{n-d}{2n}\sigma^2+\frac{7\epsilon^2 d\sigma^2}{100n}.$

By Lemma~\ref{lem:genera_large_sample}, assuming $\sigma$ is a constant and assuming $n\geq c\log(\frac{n}{\epsilon d})d$ for some constant $c,$ we have with probability at least $1-m\expn-O(\frac{tn}{\epsilon^2 d}+m)\exp(-\Omega(m\epsilon^4 d^2/n^2)),$
$$|\bftr(\eta)-\hftr(\eta)| \leq \frac{17\epsilon^2 d\sigma^2}{n},$$
for all $\eta\in[0,\heta].$ This then implies
\begin{align*}
\bftr(\etatr)\leq \hftr(\etatr)+\frac{17\epsilon^2 d\sigma^2}{n}
\leq \frac{n-d}{2n}\sigma^2+\frac{24\epsilon^2 d\sigma^2}{n}.
\end{align*} 

By the analysis in Lemma~\ref{lem:good_point_large_sample}, we have 
\begin{align*}
\bftr(\etatr)=& \Estr \frac{1}{2}\ns{w_{t,\etatr}-\wtr}_{\htr}+\Estr \frac{1}{2n}\ns{(I_n-\proj_{\xtr})\xitr}\\
=& \Estr \frac{1}{2}\ns{w_{t,\etatr}-\wtr}_{\htr}+\frac{n-d}{2n}\sigma^2.
\end{align*}

Therefore, we know $\Estr \frac{1}{2}\ns{w_{t,\etatr}-\wtr}_{\htr}\leq \frac{24\epsilon^2 d\sigma^2}{n}.$ Next, we show this implies $\Estr \ns{w_{t,\etatr}-w^*}$ is small. 

Let $\calE$ be the event that $1-\epsilon \leq \lambda_i(\htr)\leq 1+\epsilon$ for all $i\in[d].$ According to Lemma~\ref{lem:isotropic_eps}, we know $\Pr[\calE]\geq 1-\expepsn$ as long as $n\geq 10d/\epsilon^2.$ Then, we can decompose $\Estr \ns{w_{t,\etatr}-w^*}$ as follows,
$$\Estr \ns{w_{t,\etatr}-w^*} = \Estr \ns{w_{t,\etatr}-w^*}\indic{\calE}+\Estr \ns{w_{t,\etatr}-w^*}\indic{\bar{\calE}}.$$

Let's first show the second term is small. Due to the truncation in our algorithm, we know $\ns{w_{t,\etatr}-w^*}\leq 41^2\sigma^2,$ which then implies $\Estr \ns{w_{t,\etatr}-w^*}\indic{\bar{\calE}}\leq 41^2\sigma^2\expepsn.$ As long as $n\geq \frac{c}{\epsilon^2}\log(\frac{n}{\epsilon d})$ for some constant $c$, we have $\Estr \ns{w_{t,\etatr}-w^*}\indic{\bar{\calE}}\leq \frac{\epsilon d \sigma^2}{n}.$

We can upper bound the first term by Young's inequality,
$$\Estr \ns{w_{t,\etatr}-w^*}\indic{\calE}\leq (1+\frac{1}{\epsilon})\Estr \ns{w_{t,\etatr}-\wtr}\indic{\calE}+(1+\epsilon)\Estr \ns{\wtr-w^*}\indic{\calE}.$$
Conditioning on $\calE,$ we have $\ns{w_{t,\etatr}-\wtr}_{\htr}\geq (1-\epsilon)\ns{w_{t,\etatr}-\wtr}$ which implies $\ns{w_{t,\etatr}-\wtr}\leq  (1+2\epsilon)\ns{w_{t,\etatr}-\wtr}_{\htr}$ as long as $\epsilon\leq 1/2.$ Similarly, we also have $\ns{\wtr-w^*}\leq (1+2\epsilon)\ns{\wtr-w^*}_{\htr}.$ Then, we have 
\begin{align*}
&\Estr \ns{w_{t,\etatr}-w^*}\indic{\calE}\\
\leq& (1+\frac{1}{\epsilon})(1+2\epsilon)\Estr \ns{w_{t,\etatr}-\wtr}_{\htr}\indic{\calE}+(1+\epsilon)(1+2\epsilon)\Estr \ns{\wtr-w^*}_{\htr}\indic{\calE}\\
\leq& (5+\frac{1}{\epsilon})\Estr \ns{w_{t,\etatr}-\wtr}_{\htr}+(1+5\epsilon)\Estr \ns{\wtr-w^*}_{\htr}\\
\leq& (5+\frac{1}{\epsilon})\frac{48\epsilon^2 d\sigma^2 }{n}+(1+5\epsilon)\frac{d\sigma^2}{n}\leq (1+293\epsilon)\frac{d\sigma^2}{n}.
\end{align*}

Overall, we have $\Estr \ns{w_{t,\etatr}-w^*}\leq (1+293\epsilon)\frac{d\sigma^2}{n}+\frac{\epsilon d\sigma^2}{n}=(1+294\epsilon)\frac{d\sigma^2}{n}.$ Combining all the conditions, we know this holds with probability at least $0.99$ as long as $\sigma$ is a constant $c_1$, $n\geq \frac{cd}{\epsilon^2}\log(\frac{nm}{\epsilon d}), t\geq c_2\log(\frac{n}{\epsilon d}), m\geq \frac{c_3n^2}{\epsilon^4 d^2}\log(\frac{tnm}{\epsilon d}), d\geq c_4$ for some constants $c,c_2,c_3,c_4.$ We finish the proof by choosing $\epsilon=\epsilon'/294.$
\end{proofof}

\subsection{Upper bounding $\hftr(2/3)$}\label{sec:good_point_large_sample}
In this section, we show there exists a step size that achieves small empirical meta objective. On each training task $P$, we show the meta-loss can be decomposed into two terms:
\begin{align*}
\dtr=& \frac{1}{2n}\sum_{i=1}^n\pr{\inner{\wt-\wtr}{x_i}-\pr{\xi_i-x_i^\top \xtr^\dagger\xitr}}^2\\
=& \frac{1}{2}\ns{\wt-\wtr}_{\htr}+\frac{1}{2n}\ns{(I_n-\proj_{\xtr})\xitr},
\end{align*}
where $\wtr=w^*+(\xtr)^\dagger\xitr.$
In Lemma~\ref{lem:good_point_large_sample}, we show with a constant step size, the first term is exponentially small and the second term concentrates on its mean.

\GoodPointLargeSample*

Before we go to the proof of Lemma~\ref{lem:good_point_large_sample}, let's first show the covariance matrix $\htr$ is very close to identity when $n$ is much larger than $d$. The proof follows from the concentration of singular values of random Gaussian matrix (Lemma~\ref{lem:sig_matrix}). We leave the proof into Section~\ref{sec:technical_train_large}.
\begin{lemma}\label{lem:isotropic_eps}
Given $1>\epsilon>0$, assume $n\geq 10d/\epsilon^2.$ With probability at least $1-\expepsn,$
$$(1-\epsilon)\sqrt{n}\leq \sigma_i(\xtr)\leq (1+\epsilon)\sqrt{n} \text{ and } 1-\epsilon\leq \lambda_i(\htr)\leq 1+\epsilon,$$
for all $i\in[d].$
\end{lemma}

Now, we are ready to present the proof of Lemma~\ref{lem:good_point_large_sample}.

\begin{proofof}{Lemma~\ref{lem:good_point_large_sample}}
Let's first look at one training set $\str,$ in which $y_i=\inner{w^*}{x_i}+\xi_i$ for each sample. 
Recall the meta-loss as 
$$\dtr=\frac{1}{2n}\sum_{i=1}^n\pr{\inner{\wt}{x_i}-\inner{w^*}{x_i}-\xi_i}^2.$$
Recall that $\xtr$ is an $n\times d$ matrix with its $i$-th row as $x_i^\top.$ With probability $1$, we know $\xtr$ is full column rank. Denote the pseudo-inverse of $\xtr$ as $\xtr^\dagger\in\R^{d\times n}$ that satisfies $\xtr^\dagger\xtr=I_d$ and $\xtr\xtr^\dagger=\proj_{\xtr},$ where $\proj_{\xtr}\in\R^{n\times n}$ is a projection matrix onto the column span of $\xtr.$

Let $\wtr$ be $w^*+\xtr^\dagger\xitr,$ where $\xitr$ is an $n$-dimensional vector with its $i$-th entry as $\xi_i.$ We have,
\begin{align*}
&\dtr\\
=& \frac{1}{2n}\sum_{i=1}^n\pr{\inner{\wt-\wtr}{x_i}-\pr{\xi_i-x_i^\top \xtr^\dagger\xitr}}^2\\
=& \frac{1}{2}\ns{\wt-\wtr}_{\htr}+\frac{1}{2n}\ns{(I_n-\proj_{\xtr})\xitr}
-\frac{1}{n}\sum_{i=1}^n\inner{\wt-\wtr}{x_i\xi_i-x_i x_i^\top \xtr^\dagger\xitr}.
\end{align*}
We first show the crossing term is actually zero. We have,
\begin{align*}
\frac{1}{n}\sum_{i=1}^n\inner{\wt-\wtr}{x_i\xi_i-x_i x_i^\top \xtr^\dagger\xitr}
=& \frac{1}{n}\inner{\wt-\wtr}{\sum_{i=1}^n x_i\xi_i-\sum_{i=1}^n x_i x_i^\top \xtr^\dagger\xitr}\\
=& \frac{1}{n}\inner{\wt-\wtr}{\xtr^\top \xitr-\xtr^\top\xtr \xtr^\dagger\xitr}\\
=& \frac{1}{n}\inner{\wt-\wtr}{\xtr^\top \xitr-\xtr^\top\xitr}=0,
\end{align*}
where the second last equality holds because $\xtr\xtr^\dagger=\proj_{\xtr}.$ 

We can define $\wtrk$ as $w^*_k+(\xtrk)^\dagger \xitrk$ for every training set $\strk.$ Then, we have 
\begin{align*}
\hftr(\eta)=\frac{1}{m}\sum_{k=1}^m \frac{1}{2}\ns{\wtk - \wtrk}_{\htrk}+\frac{1}{m}\sum_{k=1}^m\frac{1}{2n}\ns{(I_n-\proj_{\xtrk})\xitrk}
\end{align*}

We first prove that the second term concentrates on its mean. We can concatenate $m$ noise vectors $\xitrk$ into a single noise vector $\bxitr$ with dimension $nm.$ We can also construct a data matrix $\bxtr\in\R^{nm\times dm}$ that  consists of $\xtrk$ as diagonal blocks. Then the second term can be written as 
$$\frac{1}{2}\ns{\frac{1}{\sqrt{nm}}(I_{nm}-\proj_{\bxtr})\bxitr}.$$
According to Lemma~\ref{lem:norm_vector}, with probability at least $1-\exp(-\Omega(\epsilon^4 md^2/n)),$
$$\pr{1-\frac{\epsilon^2 d}{n}}\sigma\leq \frac{1}{\sqrt{nm}}\n{\bxitr}\leq \pr{1+\frac{\epsilon^2 d}{n}}\sigma.$$
By Johnson-Lindenstrauss Lemma (Lemma~\ref{lem:JL_subspace}), we know with probability at least $1-\exp(-\Omega(\epsilon^4 md)),$
$$\frac{1}{\sqrt{nm}}\n{\proj_{\bxtr}\bxitr}\geq (1-\epsilon^2)\frac{\sqrt{md}}{\sqrt{mn}}\frac{1}{\sqrt{nm}}\n{\bxitr}\geq (1-\epsilon^2)\sqrt{\frac{d}{n}}(1-\frac{\epsilon^2 d}{n})\sigma.$$
Therefore, we have $\ns{\frac{1}{\sqrt{nm}}\bxitr}\leq (1+\frac{3\epsilon^2 d}{n})\sigma^2$ and $\ns{\frac{1}{\sqrt{nm}}\proj_{\bxtr}\bxitr}\geq (1-2\epsilon^2)\frac{d}{n}\sigma^2.$ Overall, we know with probability at least $1-\exp(-\Omega(\epsilon^4 md/n)),$ 
$$\frac{1}{2}\ns{\frac{1}{\sqrt{nm}}(I_{nm}-\proj_{\bxtr})\bxitr}\leq \frac{n-d}{2n}\sigma^2+\frac{5\epsilon^2 d\sigma^2}{2n}.$$

Now, we show the first term in meta objective is small when we choose a right step size. According to Lemma~\ref{lem:isotropic_eps}, we know as long as $n\geq 40d,$ with probability at least $1-\expn,$ $\sqrt{n}/2\leq \sigma_i(\xtrk)\leq 3\sqrt{n}/2 \text{ and } 1/2\leq \lambda_i(\htrk)\leq 3/2,$
for all $i\in[d].$ According to Lemma~\ref{lem:norm_vector}, we know with probability at least $1-\expn,$ $\n{\xitrk}\leq 2\sqrt{n}\sigma$. Taking a union bound on $m$ tasks, we know all these events hold with probability at least $1-m\expn.$ 

For each $k\in[m],$ we have $\n{\wtrk}\leq 1+\frac{2}{\sqrt{n}}2\sqrt{n}\sigma\leq 5\sigma.$
It's easy to verify that for any step size at most $2/3,$ the GD sequence will not be truncated since we choose the threshold norm as $40\sigma.$ Then, for any step size $\eta \leq 2/3,$ we have 
\begin{align*}
\frac{1}{m}\sum_{k=1}^m \frac{1}{2}\ns{\wtk - \wtrk}_{\htrk} =& \frac{1}{m}\sum_{k=1}^m \frac{1}{2}\ns{(I-\eta\htrk)^t\wtrk}_{\htrk}\\
\leq& \frac{3}{4}(1-\frac{\eta}{2})^{2t}25\sigma^2\leq 20(1-\frac{1}{3})^{2t}\sigma^2,
\end{align*}
where the last inequality chooses $\eta$ as $2/3.$ 

Overall, we know with probability at least $1-m\expn-\exp(-\Omega(\epsilon^4 md/n)),$
$$\hftr(2/3)\leq 20(1-\frac{1}{3})^{2t}\sigma^2+\frac{n-d}{2n}\sigma^2+\frac{5\epsilon^2 d\sigma^2}{2n}.$$
We finish the proof by changing $\frac{5\epsilon^2}{2}$ by $(\epsilon')^2/20.$
\end{proofof}

\subsection{Lower bounding $\hftr$ for $\eta\in(\heta,\infty)$}\label{sec:high_loss_large_sample}
In this section, we show the empirical meta objective is large when the step size exceeds certain threshold. Recall Lemma~\ref{lem:diverge_large_sample} as follows.

\DivergeLargeSample*

Roughly speaking, we define $\heta$ such that for any step size larger than $\heta$ the GD sequence has a reasonable probability being truncated. The definition is very similar as $\teta$ in Definition~\ref{def:teta}. 

\begin{definition}\label{def:heta}
Given a training task $P,$ let $\calE_1$ be the event that $\sqrt{n}/2 \leq \sigma_i(\xtr)\leq 3\sqrt{n}/2$ and $1/2 \leq \lambda_i(\htr) \leq 3/2$ for all $i\in[d]$ and $\sqrt{n}\sigma/2 \leq \n{\xitr}\leq 2\sqrt{n}\sigma.$ Let $\bar{\calE}_2(\eta)$ be the event that the GD sequence is truncated with step size $\eta.$ Given $1>\epsilon>0$, define $\heta$ as follows,
$$\heta=\inf\left\{\eta\geq 0 \middle| \Estr \frac{1}{2}\ns{\wt-\wtr}_{\htr}\indic{\calE_1\cap \bar{\calE}_2(\eta)}\geq \frac{\epsilon^2 d\sigma^2}{n}\right\}.$$
\end{definition}

Similar as in Lemma~\ref{lem:monotone}, we show $\indic{\calE_1\cap \bar{\calE}_2(\eta')}\geq \indic{\calE_1\cap \bar{\calE}_2(\eta)}$ for any $\eta'\geq \eta.$ This means conditioning on $\calE_1,$ if a GD sequence gets truncated with step size $\eta,$ it has to be truncated with any step size $\eta'\geq \eta.$ The proof is deferred into Section~\ref{sec:technical_train_large}. 

\begin{lemma}\label{lem:monotone_large_sample}
Fixing a training set $\str,$ let $\calE_1$ and $\bar{\calE}_2(\eta)$ be as defined in Definition~\ref{def:heta}. We have 
$$\indic{\calE_1\cap \bar{\calE}_2(\eta')}\geq \indic{\calE_1\cap \bar{\calE}_2(\eta)},$$
for any $\eta'\geq \eta.$ 
\end{lemma}

Next, we show $\heta$ does exist and is a constant. Similar as in Lemma~\ref{lem:range_teta}, we show that the GD sequence almost never diverges when $\eta$ is small and diverges with high probability when $\eta$ is large. The proof is left in Section~\ref{sec:technical_train_large}.

\begin{lemma}\label{lem:range_heta}
Let $\heta$ be as defined in Definition~\ref{def:heta}. Suppose $\sigma$ is a constant. Assume $n\geq cd,t\geq c_2,d\geq c_4$ for some constants $c,c_2,c_4.$ 
We have 
$$\frac{4}{3}<\teta<6.$$
\end{lemma}

Next, we show the empirical loss is large for any $\eta$ larger than $\teta.$ The proof is very similar as the proof of Lemma~\ref{lem:diverge_train_GD}.

\begin{proofof}{Lemma~\ref{lem:diverge_large_sample}}
By Lemma~\ref{lem:range_heta}, we know $\heta$ is a constant as long as $n\geq cd,t\geq c_2,d\geq c_4$ for some constants $c,c_2,c_4.$ 
Let $\calE_1$ and $\bar{\calE}_2(\eta)$ be as defined in Definition~\ref{def:heta}.
For the simplicity of the proof, we assume $\Estr \frac{1}{2}\ns{w_{t,\heta}-\wtr}_{\htr}\indic{\calE_1\cap \bar{\calE}_2(\heta)}\geq \frac{\epsilon^2 d\sigma^2}{n}.$ The other case can be resolved using same techniques in Lemma~\ref{lem:diverge_train_GD}

Conditioning on $\calE_1,$ we know $\frac{1}{2}\ns{w_{t,\heta}-\wtr}_{\htr}\leq \frac{3}{4}45^2\sigma^2.$ Therefore, we know $\Pr[\calE_1\cap \bar{\calE}_2(\heta)]\geq \frac{4\epsilon^2 d}{3\times 45^2 n}.$ 
For each task $k$, define $\calE_1^{(k)}$ and $\bar{\calE}_2^{(k)}(\eta)$ as the corresponding events on training set $\strk.$ By Hoeffding's inequality, we know with probability at least $1-\exp(-\Omega(\epsilon^4 md^2/n^2)),$ 
$$\frac{1}{m}\sum_{k=1}^m \indic{\calE_1^{(k)}\cap \bar{\calE}_2^{(k)}(\heta)}\geq \frac{\epsilon^2 d}{45^2 n}.$$
By Lemma~\ref{lem:monotone_large_sample}, we know $\indic{\calE_1^{(k)}\cap \bar{\calE}_2^{(k)}(\eta)}\geq\indic{\calE_1^{(k)}\cap \bar{\calE}_2^{(k)}(\heta)}$ for any $\eta\geq \heta.$ 

Recall that 
\begin{align*}
\hftr(\eta)=\frac{1}{m}\sum_{k=1}^m \frac{1}{2}\ns{\wtk - \wtrk}_{\htrk}+\frac{1}{m}\sum_{k=1}^m\frac{1}{2n}\ns{(I_n-\proj_{\xtrk})\xitrk}.
\end{align*}
We can lower bound the first term for any $\eta>\heta$ as follows,
\begin{align*}
\hftr(\eta)=\frac{1}{m}\sum_{k=1}^m \frac{1}{2}\ns{\wtk-\wtrk}_{\htrk}
\geq& \frac{1}{m}\sum_{k=1}^m \frac{1}{2}\ns{\wtk-\wtrk}_{\htrk}\indic{\calE_1^{(k)}\cap \bar{\calE}_2^{(k)}(\eta)}\\
\geq& \frac{35^2\sigma^2}{4}\frac{1}{m}\sum_{k=1}^m\indic{\calE_1^{(k)}\cap \bar{\calE}_2^{(k)}(\eta)}\\
\geq& \frac{35^2\sigma^2}{4}\frac{1}{m}\sum_{k=1}^m\indic{\calE_1^{(k)}\cap \bar{\calE}_2^{(k)}(\heta)}\geq \frac{\epsilon^2 d\sigma^2}{8n},
\end{align*}
where the second inequality lower bounds the loss for one task by $35^2\sigma^2$ when the sequence gets truncated. 

For the second term, according to the analysis in Lemma~\ref{lem:good_point_large_sample}, with probability at least $1-\exp(-\Omega(\epsilon^4 md/n)),$ 
$$\frac{1}{m}\sum_{k=1}^m\frac{1}{2n}\ns{(I_n-\proj_{\xtrk})\xitrk} \geq \frac{n-d}{2n}\sigma^2-\frac{\epsilon^2 d\sigma^2}{20n}.$$

Overall, with probability at least $1-\exp(-\Omega(\epsilon^4 md^2/n^2)),$
$$\hftr(\eta)\geq \frac{\epsilon^2 d\sigma^2}{8n}+\frac{n-d}{2n}\sigma^2-\frac{\epsilon^2 d\sigma^2}{20n},$$
for all $\eta>\heta.$
\end{proofof}

\subsection{Generalization for $\eta\in[0,\heta]$}\label{sec:genera_large_sample}
Combing Lemma~\ref{lem:good_point_large_sample} and Lemma~\ref{lem:diverge_large_sample}, it's not hard to see that the optimal step size $\etatr$ lies in $[0,\heta].$ In this section, we show a generalization result for step sizes in $[0,\heta].$ The proof of Lemma~\ref{lem:genera_large_sample} is given at the end of this section.

\generaLargeSample*

In Lemma~\ref{lem:genera_fixed_large_sample}, we show $\hftr$ concentrates on $\bftr$ at any fixed step size. The proof is almost the same as Lemma~\ref{lem:genera_fixed_eta_train_GD}. We omit its proof.
\begin{lemma}\label{lem:genera_fixed_large_sample}
Suppose $\sigma$ is a constant. For any fixed $\eta$ and any $1>\epsilon>0,$ with probability at least $1-\expepsm,$
$$\absr{\hftr(\eta)-\bftr(\eta) }\leq \epsilon.$$
\end{lemma}

Next, we construct an $\epsilon$-net for $\bftr$ in $[0,\heta].$ The proof is very similar as in Lemma~\ref{lem:eps_net_train_expect}. We defer its proof into Section~\ref{sec:technical_train_large}. 

\begin{lemma}\label{lem:eps_net_expect_large_sample}
Let $\heta$ be as defined in Definition~\ref{def:heta} with $1>\epsilon>0$.
Assume the conditions in Lemma~\ref{lem:range_heta} hold.
Assume $n\geq c\log(\frac{n}{\epsilon d})d$ for some constant $c.$ There exists an $\frac{8\epsilon^2 d\sigma^2}{n}$-net $N\subset [0,\heta]$ for $\bftr$ with $|N|=O(\frac{tn}{\epsilon^2 d}).$ That means, for any $\eta\in[0,\heta],$
$$|\bftr(\eta)-\bftr(\eta')|\leq \frac{8\epsilon^2 d\sigma^2}{n},$$
for $\eta'=\arg\min_{\eta''\in N,\eta''\leq \eta}(\eta-\eta'').$
\end{lemma}

We also construct an $\epsilon$-net for the empirical meta objective. The proof is very similar as in Lemma~\ref{lem:eps_net_train_empirical}. We leave its proof into Section~\ref{sec:technical_train_large}.

\begin{lemma}\label{lem:eps_net_empirical_large_sample}
Let $\heta$ be as defined in Definition~\ref{def:heta} with $1>\epsilon>0$.
Assume the conditions in Lemma~\ref{lem:range_heta} hold.
Assume $n\geq 40d.$ With probability at least $1-m\expn,$ there exists an $\frac{\epsilon^2 d\sigma^2}{n}$-net $N'\subset [0,\heta]$ for $\hftr$ with $|N'|=O(\frac{tn}{\epsilon^2 d}+m).$ That means, for any $\eta\in[0,\heta],$
$$|\hftr(\eta)-\hftr(\eta')|\leq \frac{\epsilon^2 d\sigma^2}{n},$$
for $\eta'=\arg\min_{\eta''\in N',\eta''\leq \eta}(\eta-\eta'').$
\end{lemma}

Combing the above three lemmas, we give the proof of Lemma~\ref{lem:genera_large_sample}.

\begin{proofof}{Lemma~\ref{lem:genera_large_sample}}
We assume $\sigma$ as a constant in this proof.
By Lemma~\ref{lem:genera_fixed_large_sample}, we know with probability at least $1-\exp(-\Omega(m\epsilon^4 d^2/n^2)),$
$\absr{\hftr(\eta)-\bftr(\eta) }\leq \frac{\epsilon^2 d\sigma^2}{n}$ for any fixed $\eta.$
By Lemma~\ref{lem:eps_net_expect_large_sample}, we know as long as $n\geq c\log(\frac{n}{\epsilon d})d$ for some constant $c$, there exists an $\frac{8\epsilon^2 d\sigma^2}{n}$-net $N$ for $\bftr$ with size $O(\frac{tn}{\epsilon^2 d}).$ By Lemma~\ref{lem:eps_net_empirical_large_sample}, we know with probability at least $1-m\expn$, there exists an $\frac{\epsilon^2 d\sigma^2}{n}$-net $N'$ for $\hftr$ with size $O(\frac{tn}{\epsilon^2 d}+m).$   
It's not hard to verify that $N\cup N'$ is still an $\frac{8\epsilon^2 d\sigma^2}{n}$-net for $\hfva$ and $\bfva$. That means, for any $\eta\in[0,\heta],$ we have
$$|\bftr(\eta)-\bftr(\eta')|,|\hftr(\eta)-\hftr(\eta')|\leq \frac{8\epsilon^2 d\sigma^2}{n},$$ 
for $\eta'=\arg\min_{\eta''\in N\cup N',\eta''\leq \eta}(\eta-\eta'').$

Taking a union bound over $N\cup N',$ we have with probability at least $1-O(\frac{tn}{\epsilon^2 d}+m)\exp(-\Omega(m\epsilon^4 d^2/n^2)),$
$$\absr{\hftr(\eta)-\bftr(\eta) }\leq \frac{\epsilon^2 d\sigma^2}{n}$$ for all $\eta\in N\cup N'.$

Overall, we know with probability at least $1-m\expn-O(\frac{tn}{\epsilon^2 d}+m)\exp(-\Omega(m\epsilon^4 d^2/n^2)),$ for all $\eta\in[0,\heta],$
\begin{align*}
&|\bftr(\eta)-\hftr(\eta)| \\
\leq& |\bftr(\eta)-\bftr(\eta')|+|\hftr(\eta)-\hftr(\eta')| + |\hftr(\eta')-\bftr(\eta')|\\
\leq& \frac{17\epsilon^2 d\sigma^2}{n},
\end{align*}
where $\eta'=\arg\min_{\eta''\in N\cup N',\eta''\leq \eta}(\eta-\eta'').$
\end{proofof}

\subsection{Proofs of Technical Lemmas}\label{sec:technical_train_large}
\begin{proofof}{Lemma~\ref{lem:isotropic_eps}}
According to Lemma~\ref{lem:sig_matrix}, we know with probability at least $1-2\exp(-t^2/2),$
$$\sqrt{n}-\sqrt{d}-t \leq \sigma_i(\xtr)\leq \sqrt{n}+\sqrt{d}+t$$
for all $i\in[d].$ Since $d\leq \frac{\epsilon^2 n}{10},$ we have $\sqrt{n}-\frac{\epsilon\sqrt{n}}{\sqrt{10}}-t \leq \sigma_i(\xtr)\leq \sqrt{n}+\frac{\epsilon\sqrt{n}}{\sqrt{10}}+t$. Choosing $t=(\frac{1}{3}-\frac{1}{\sqrt{10}})\epsilon\sqrt{n},$ we have with probability at least $1-\expepsn,$
$$(1-\frac{\epsilon}{3})\sqrt{n}\leq \sigma_i(\xtr)\leq (1+\frac{\epsilon}{3})\sqrt{n}.$$
Since $\lambda_i(\htr)=1/n\sigma^2_i(\xtr),$ we have
$1-\epsilon\leq \lambda_i(\htr)\leq 1+\epsilon.$
\end{proofof}

\begin{proofof}{Lemma~\ref{lem:monotone_large_sample}}
The proof is almost the same as in Lemma~\ref{lem:monotone}. We omit the details here. Basically, in Lemma~\ref{lem:monotone}, the only property we rely on is that the norm threshold is larger than $2\n{\wtr}$ conditioning on $\calE_1.$ Conditioning on $\calE_1,$ we know $\n{\wtr}\leq 5\sigma.$ Recall that the norm threshold is still set as $40\sigma.$ So this property is preserved and the previous proof works.
\end{proofof}

\begin{proofof}{Lemma~\ref{lem:range_heta}}
The proof is very similar as in Lemma~\ref{lem:range_teta}. 
Conditioning on $\calE_1,$ we know $\n{\htr}\leq 3/2$ and $\n{\wtr}\leq 5\sigma.$ So the GD sequence never exceeds the norm threshold $40\sigma$ for any $\eta\leq 4/3.$ That means, 
$$\Estr \frac{1}{2}\ns{\wt-\wtr}_{\htr}\indic{\calE_1\cap \bar{\calE}_2(\eta)}=0$$
for all $\eta\leq 4/3.$

To lower bound the loss for large step size, we need to first lower bound $\n{\wtr}.$ Recall that $\wtr=w^*+(\xtr)^\dagger\xitr.$ Conditioning on $\calE_1,$ we know $\n{\xitr}\leq 2\sqrt{n}\sigma$ and $\sigma_d(\xtr)\geq \sqrt{n}/2,$ which implies $\n{(\xtr)^\dagger}\leq 2/\sqrt{n}.$ By Johnson-Lindenstrauss Lemma (Lemma~\ref{lem:JL_subspace}), we have $\n{\proj_{\xtr}\xitr}\leq \frac{3}{2}\sqrt{d/n}\n{\xitr}$ with probability at least $1-\expd.$ Call this event $\calE_3.$ Conditioning on $\calE_1\cap \calE_3,$ we have
$$\n{(\xtr)^\dagger\xitr}\leq 2\sqrt{n}\sigma\frac{2}{\sqrt{n}}\frac{3}{2}\sqrt{\frac{d}{n}}\leq 6\sqrt{\frac{d}{n}}\sigma,$$
which is smaller than $1/2$ as long as $n\geq 12^2 d\sigma^2.$ Note that we assume $\sigma$ is a constant. This then implies $\n{\wtr}\geq 1/2.$

Let $\{w_{\tau,\eta}'\}$ be the GD sequence without truncation. For any step size $\eta\in[6,\infty],$ conditioning on $\calE_1\cap \calE_3,$ we have
$$\n{\wt'}\geq \pr{(6\times \frac{1}{2}-1)^t -1}\n{\wtr}\geq \pr{2^t -1}\frac{1}{2}\geq 40\sigma,$$
where the last inequality holds as long as $t\geq c_2$ for some constant $c_2$. Therefore, we know when $\eta\in[6,\infty),$ $\indic{\calE_1\cap \bar{\calE}_2(\eta)}=\indic{\calE_1\cap\calE_3}$. Assuming $n\geq 40d,$ we know $\calE_1$ holds with probability at least $1-\expn.$ 
Then, we have for any $\eta\geq 6,$
\begin{align*}
\Estr \frac{1}{2}\ns{\wt-\wtr}_{\htr}\indic{\calE_1\cap \bar{\calE}_2(\eta)}
\geq& \frac{1}{4}\pr{40\sigma-5\sigma}^2\Pr[\calE_1\cap \calE_3]\geq \frac{\epsilon^2 d\sigma^2}{n},
\end{align*}
where the last inequality assumes $n\geq c,d\geq c_4$ for some constant $c,c_4.$

Overall, we know $\Estr \frac{1}{2}\ns{\wt-\wtr}_{\htr}\indic{\calE_1\cap \bar{\calE}_2(\eta)}$ equals zero for all $\eta\in[0,4/3]$ and is at least $\frac{\epsilon^2 d\sigma^2}{n}$ for all $\eta\in[6,\infty).$ By definition, we know $\heta\in(4/3,6).$ 
\end{proofof}

\begin{proofof}{Lemma~\ref{lem:eps_net_expect_large_sample}}
By Lemma~\ref{lem:range_heta}, we know $\heta$ is a constant.
The proof is very similar as in Lemma~\ref{lem:eps_net_train_expect}.
Let $\calE_1$ and $\bar{\calE}_2(\eta)$ be as defined in Definition~\ref{def:heta}.
For the simplicity of the proof, we assume $\Estr \frac{1}{2}\ns{w_{t,\heta}-\wtr}_{\htr}\indic{\calE_1\cap \bar{\calE}_2(\heta)}\leq \frac{\epsilon^2 d\sigma^2}{n}.$ The other case can be resolved using techniques in the proof of Lemma~\ref{lem:eps_net_train_expect}.

Recall the population meta objective 
$$\bftr(\eta)=\Estr \frac{1}{2}\ns{w_{t,\eta}-\wtr}_{\htr}+\frac{n-d}{2n}\sigma^2.$$
Therefore, we only need to construct an $\epsilon$-net for the first term.

We can divide $\Estr \frac{1}{2}\ns{w_{t,\eta}-\wtr}_{\htr}$ as follows,
\begin{align*}
&\Estr \frac{1}{2}\ns{w_{t,\eta}-\wtr}_{\htr}\\ 
=& \Estr \frac{1}{2}\ns{w_{t,\eta}-\wtr}_{\htr}\indic{\calE_1\cap \calE_2(\heta)}+\Estr \frac{1}{2}\ns{w_{t,\eta}-\wtr}_{\htr}\indic{\calE_1\cap \bar{\calE}_2(\heta)}\\
&+\Estr \frac{1}{2}\ns{w_{t,\eta}-\wtr}_{\htr}\indic{\bar{\calE}_1}.
\end{align*}

We will construct an $\epsilon$-net for the first term and show the other two terms are small. Let's first consider the third term. Assuming $n\geq 40d,$ we know $\Pr[\calE_1]\leq \expn.$ Since $\frac{1}{2}\ns{w_{t,\eta}-\wtr}_{\htr}$ is $O(1)$-subexponential, by Cauchy-Schwarz inequality, we have $\Estr \frac{1}{2}\ns{w_{t,\eta}-\wtr}_{\htr}\indic{\bar{\calE}_1}=O(1)\expn.$ Choosing $n\geq c\log(n/(\epsilon d))$ for some constant $c$, 
we know $\frac{1}{2}\ns{w_{t,\heta}-\wtr}_{\htr}\indic{\bar{\calE}_1}\leq \frac{\epsilon^2 d\sigma^2}{n}.$

Then we upper bound the second term. Since $\Estr \frac{1}{2}\ns{w_{t,\heta}-\wtr}_{\htr}\indic{\calE_1\cap \bar{\calE}_2(\heta)}\leq \frac{\epsilon^2 d\sigma^2}{n}$ and \\$\frac{1}{2}\ns{w_{t,\heta}-\wtr}_{\htr}\geq \frac{35^2\sigma^2}{4}$ when $w_{t,\heta}$ diverges, we know $\Pr[\calE_1\cap \bar{\calE}_2(\heta)]\leq \frac{4\epsilon^2 d}{35^2 n}.$ Then, we can upper bound the second term as follows,
\begin{align*}
\Estr \frac{1}{2}\ns{w_{t,\eta}-\wtr}_{\htr}\indic{\calE_1\cap \bar{\calE}_2(\heta)}\leq \frac{3\times 45^2\sigma^2}{4} \frac{4\epsilon^2 d}{35^2 n}\leq \frac{6\epsilon^2 d\sigma^2}{n}
\end{align*}

Next, similar as in Lemma~\ref{lem:eps_net_train_expect}, we can show the first term $\frac{1}{2}\ns{w_{t,\eta}-\wtr}_{\htr}\indic{\calE_1\cap \calE_2(\heta)}$ is $O(t)$-lipschitz. Therefore, there exists an $\frac{\epsilon^2 d\sigma^2}{n}$-net $N$ for $\Estr \frac{1}{2}\ns{w_{t,\eta}-\wtr}_{\htr}\indic{\calE_1\cap \calE_2(\heta)}$ with size $O(\frac{tn}{\epsilon^2 d})$. That means, for any $\eta\in[0,\heta],$
$$\absr{\Estr \frac{1}{2}\ns{w_{t,\eta}-\wtr}_{\htr}\indic{\calE_1\cap \calE_2(\heta)}-\Estr \frac{1}{2}\ns{w_{t,\eta'}-\wtr}_{\htr}\indic{\calE_1\cap \calE_2(\heta)}}\leq \frac{\epsilon^2 d\sigma^2}{n}$$ 
for $\eta'=\arg\min_{\eta''\in N,\eta''\leq \eta}(\eta-\eta'').$ 

Combing with the upper bounds on the second term and the third term, we have for any $\eta\in[0,\heta],$
$$\absr{\bftr(\eta)-\bftr(\eta')}\leq \frac{8\epsilon^2 d\sigma^2}{n}$$ 
for $\eta'=\arg\min_{\eta''\in N,\eta''\leq \eta}(\eta-\eta'').$
\end{proofof}

\begin{proofof}{Lemma~\ref{lem:eps_net_empirical_large_sample}}
By Lemma~\ref{lem:range_heta}, we know $\heta$ is a constant.
For each $k\in[m],$ let $\calE_{1,k}$ be the event that $\sqrt{n}/2 \leq \sigma_i(\xtrk)\leq 3\sqrt{n}/2$ and $1/2 \leq \lambda_i(\htrk) \leq 3/2$ for all $i\in[d]$ and $\sqrt{n}\sigma/2 \leq \n{\xitrk}\leq 2\sqrt{n}\sigma$. Assuming $n\geq 40d,$ by Lemma~\ref{lem:isotropic_eps}, we know with probability at least $1-m\expn,$ $\calE_{1,k}$'s hold for all $k\in[m].$ 

Then, similar as in Lemma~\ref{lem:eps_net_train_empirical}, there exists an $\frac{\epsilon^2 d\sigma^2}{n}$-net $N'$ with $|N'|=O(\frac{nt}{\epsilon^2 d}+m)$ for $\hftr$. That means, for any $\eta\in[0,\heta],$
$$\absr{\hftr(\eta)-\hftr(\eta')}\leq \frac{\epsilon^2 d\sigma^2}{n}$$ 
for $\eta'=\arg\min_{\eta''\in N',\eta''\leq \eta}(\eta-\eta'').$ 
\end{proofof}

\section{Proofs of train-by-train v.s. train-by-validation (SGD)}\label{sec:proofs_SGD}
Previously, we have shown that train-by-validation generalizes better than train-by-train when the tasks are trained by GD and when the number of samples is small. In this section, we show a similar phenomenon also appears in the SGD setting.

In the train-by-train setting, each task $P$ contains a training set $\str=\{(x_i,y_i)\}_{i=1}^n$. The inner objective is defined as $\hat{f}(w)=\frac{1}{2n}\sum_{(x,y)\in\str}\pr{\inner{w}{x}-y}^2.$ Let $\{w_{\tau,\eta}\}$ be the SGD sequence running on $\hat{f}(w)$ from initialization $0$ (without truncation). That means,
$w_{\tau,\eta} =w_{\tau-1,\eta}-\eta\hat{\nabla} \hat{f}(w_{\tau-1,\eta}),$
where 
$\hat{\nabla} \hat{f}(w_{\tau-1,\eta}) = \pr{\inner{w_{\tau-1,\eta}}{x_{i(\tau-1)}}-y_{i(\tau-1)}}x_{i(\tau-1)}.$
Here index $i(\tau-1)$ is independently and uniformly sampled from $[n].$ We denote the SGD noise as $n_{\tau-1,\eta}:=\hat{\nabla} \hat{f}(w_{\tau-1,\eta})-\nabla \hat{f}(w_{\tau-1,\eta}).$ The meta-loss on task $P$ is defined as follows, 
$$\Delta_{TbT(n)}(\eta,P)=\Esgd \hat{f}(\wt)=\Esgd \frac{1}{2n}\sum_{(x,y)\in\str}\pr{\inner{\wt}{x}-y}^2,$$
where the expectation is taken over the SGD noise. Note $\wt$ depends on the SGD noise along the trajectory. Then, the empirical meta objective $\hat{F}_{TbT(n)}(\eta)$ is the average of the meta-loss across $m$ different specific tasks
\begin{equation}
\hat{F}_{TbT(n)}(\eta)=\frac{1}{m}\sum_{k=1}^m \Delta_{TbT(n)}(\eta,P_k).\label{eqn:train_SGD}
\end{equation}

In order to control the SGD noise in expectation, we restrict the feasible set of step sizes into $O(1/d).$ We show within this range, the optimal step size under $\hat{F}_{TbT(n)}$ is $\Omega(1/d)$ and the learned weight is far from ground truth $w^*$ on new tasks. We prove Theorem~\ref{thm:train_SGD} in Section~\ref{sec:proofs_train_SGD}.

\begin{restatable}
{theorem}{thmTrainSGD}\label{thm:train_SGD}
Let the meta objective $\hat{F}_{TbT(n)}$ be as defined in Equation~\ref{eqn:train_SGD} with $n\in[d/4,3d/4].$
Suppose $\sigma$ is a constant. Assume unroll length $t\geq c_2d$ and dimension $d\geq c_4\log(m)$ for certain constants $c_2,c_4.$ Then, with probability at least $0.99$ in the sampling of training tasks $P_1,\cdots,P_m$ and test task $P$,
$$\etatr = \Omega(1/d) \mbox{ and } \Esgd \ns{w_{t,\etatr}-w^*}=\Omega(\sigma^2),$$
for all $\etatr \in \arg\min_{0\leq \eta\leq \frac{1}{2L^3 d}}\hat{F}_{TbT(n)}(\eta),$ where $L=100$ and $w_{t,\etatr}$ is trained by running SGD on test task $P.$ 
\end{restatable}

In the train-by-validation setting, each task $P$ contains a training set $\str$ with $n_1$ samples and a validation set with $n_2$ samples. The inner objective is defined as 
$\hat{f}(w)=\frac{1}{2n_1}\sum_{(x,y)\in\str}\pr{\inner{w}{x}-y}^2.$ Let $\{w_{\tau,\eta}\}$ be the SGD sequence running on $\hat{f}(w)$ from initialization $0$ (with the same truncation defined in Section~\ref{sec:least-squares}). For each task $P$, the meta-loss $\Delta_{TbV(n_1,n_2)}(\eta,P)$ is defined as 
$$\Delta_{TbV(n_1,n_2)}(\eta,P)=\Esgd \frac{1}{2n_2}\sum_{(x,y)\in\sva}\pr{\inner{\wt}{x}-y}^2.$$
The empirical meta objective $\hat{F}_{TbV(n_1,n_2)}(\eta)$ is the average of the meta-loss across $m$ different tasks $P_1,P_2,...,P_m$,
\begin{equation}
\hat{F}_{TbV(n_1,n_2)}(\eta)=\frac{1}{m}\sum_{k=1}^m \Delta_{TbV(n_1,n_2)}(\eta,P_k).\label{eqn:valid_SGD}
\end{equation}

In order to bound the SGD noise with high probability, we restrict the feasible set of the step sizes into $O(\frac{1}{d^2\log^2 d}).$ Within this range, we prove the optimal step size under $\hat{F}_{TbV(n_1,n_2)}$ is $\Theta(1/t)$ and the learned weight is better than initialization $0$ by a constant on new tasks. Theorem~\ref{thm:valid_SGD} is proved in Section~\ref{sec:proofs_valid_SGD}.  

\begin{restatable}
{theorem}{thmValidSGD}\label{thm:valid_SGD}
Let the meta objective $\hat{F}_{TbV(n_1,n_2)}$ be as defined in Equation~\ref{eqn:valid_SGD} with $n_1,n_2\in[d/4,3d/4]$. Assume noise level $\sigma$ is a large constant $c_1$. Assume unroll length $t\geq c_2d^2\log^2(d)$, number of training tasks $m\geq c_3$ and dimension $d\geq c_4$ for certain constants $c_2,c_3,c_4.$ There exists constant $c_5$ such that with probability at least $0.99$ in the sampling of training tasks, we have
$$\etava=\Theta(1/t)\mbox{ and }\E\ns{w_{t,\etava}-w^*}=\ns{w^*}-\Omega(1)$$
for all $\etava\in \arg\min_{0\leq \eta\leq \frac{1}{c_5d^2\log^2(d)}}\hat{F}_{TbV(n_1,n_2)}(\eta),$ where the expectation is taken over the new tasks and SGD noise.
\end{restatable}

\paragraph{Notations:}
In the following proofs, we use the same set of notations defined in Appendix~\ref{sec:proofs_GD}. We use $\Ep$ to denote the expectation over the sampling of tasks and use $\Esgd$ to denote the expectation over the SGD noise. We use $\E$ to denote $\Ep\Esgd.$ Same as in Appendix~\ref{sec:proofs_GD}, we use letter $L$ to denote constant $100$, which upper bounds $\n{\htr}$ with high probability.

\subsection{Train-by-train (SGD)}\label{sec:proofs_train_SGD}
Recall Theorem~\ref{thm:train_SGD} as follows.

\thmTrainSGD*

In order to prove Theorem~\ref{thm:train_SGD}, we first show that $\etatr$ is $\Omega(1/d)$ in Lemma~\ref{lem:large_step_SGD}. The proof is similar as in the GD setting. As long as $\eta=O(1/d),$ the SGD noise is dominated by the full gradient. Then, we can show that $\dtr$ is roughly $(1-\Theta(1)\eta)^t,$ which implies that $\etatr=\Omega(1/d).$ We leave the proof of Lemma~\ref{lem:large_step_SGD} into Section~\ref{sec:technical_train_SGD}.

\begin{lemma}\label{lem:large_step_SGD}
Assume $t\geq c_2 d$ with certain constant $c_2.$ With probability at least $1-m\expd$ in the sampling of $m$ training tasks,
$$\etatr \geq \frac{1}{6 L^5 d},$$
for all $\etatr\in \arg\min_{0\leq \eta\leq \frac{1}{2L^3 d}}\hftr(\eta).$
\end{lemma}

Let $P=(\calD(w^*),\str,\ell)$ be an independently sampled test task with $|\str|=n\in[d/4,3d/4].$ For any step size $\eta\in [\frac{1}{6L^5 d},\frac{1}{2L^3d}]$, let $\wt$ be the weight obtained by running SGD on $\hat{f}(w)$ for $t$ steps. Next, we show $\Esgd \ns{w_{t,\eta}-w^*}=\Omega(\sigma^2)$ with high probability in the sampling of $P.$

\begin{lemma}\label{lem:diverge_SGD}
Suppose $\sigma$ is a constant. Assume unroll length $t\geq c_2d$ for some constant $c_2$. With probability at least $1-\expd$ in the sampling of test task $P$,
$$\Esgd \ns{w_{t,\eta}-w^*}\geq \frac{\sigma^2}{128L},$$
for all $\eta\in[\frac{1}{6 L^5 d},\frac{1}{2 L^3 d}],$ where $\wt$ is obtained by running SGD on task $P$ for $t$ iterations.
\end{lemma}

With Lemma Lemma~\ref{lem:large_step_SGD} and Lemma~\ref{lem:diverge_SGD}, the proof of Theorem~\ref{thm:train_SGD} is straightforward.

\begin{proofof}{Theorem~\ref{thm:train_SGD}}
Combing Lemma~\ref{lem:large_step_SGD} and Lemma~\ref{lem:diverge_SGD}, we know as long as $\sigma$ is a constant, $t\geq c_2 d, d\geq c_4\log(m),$ with probability at least $0.99,$ $\etatr = \Omega(1/d) \mbox{ and } \Esgd \ns{w_{t,\etatr}-w^*}=\Omega(\sigma^2),$
for all $\etatr \in \arg\min_{0\leq \eta\leq \frac{1}{2L^3 d}}\hftr(\eta).$
\end{proofof}

\subsubsection{Detailed Proofs}\label{sec:technical_train_SGD}
\begin{proofof}{Lemma~\ref{lem:large_step_SGD}}
The proof is very similar to the proof of Lemma~\ref{lem:small_step_train} except that we need to bound the SGD noise term.
For each $k\in[m],$ let $\calE_k$ be the event that  $\sqrt{d}/\sqrt{L} \leq \sigma_i(\xtr)\leq  \sqrt{Ld}$ and $1/L \leq \lambda_i(\htr) \leq L$ for all $i\in[n]$ and $\sqrt{d}\sigma/4 \leq \n{\xitr}\leq \sqrt{d}\sigma$. According to Lemma~\ref{lem:isotropic} and Lemma~\ref{lem:norm_vector}, we know for each $k\in[m],$ $\calE_k$ happens with probability at least $1-\expd.$ Taking a union bound over all $k\in[m],$ we know $\cap_{k\in[m]}\calE_k$ holds with probability at least $1-m\expd.$ From now on, we assume $\cap_{k\in[m]}\calE_k$ holds.

For each $k\in[m],$ we have 
$$\dtrk:=\frac{1}{2}\Esgd \ns{\wtk -\wtrk}_{\htrk}.$$
Since $1/L \leq \lambda_i(\htrk)\leq L$ and $(\wtk-\wtrk)$ is in the span of $\htrk$, we have 
$$\frac{1}{2L}\Esgd \ns{\wtk -\wtrk}\leq \dtrk \leq \frac{L}{2}\Esgd \ns{\wtk -\wtrk}.$$

Recall the updates of stochastic gradient descent,
$$\wtk-\wtrk = (I-\eta\htrk)(w_{t-1,\eta}^{(k)}-\wtrk)-\eta n_{t-1,\eta}^{(k)}.$$
Therefore, 
$$\Esgd \br{ \ns{\wtk-\wtrk}| w_{t-1,\eta}^{(k)}} = \ns{(I-\eta\htrk)(w_{t-1,\eta}^{(k)}-\wtrk)}+\eta^2 \Esgd \br{\ns{n_{t-1,\eta}^{(k)}}|w_{t-1,\eta}^{(k)}}.$$
We know for any $\eta\leq 1/L,$
$$(1-2\eta L)\ns{w_{t-1,\eta}^{(k)}-\wtrk} \leq \ns{(I-\eta\htrk)(w_{t-1,\eta}^{(k)}-\wtrk)}\leq (1-\frac{\eta}{L})\ns{w_{t-1,\eta}^{(k)}-\wtrk}.$$
The noise can be bounded as follows,
\begin{align*}
&\eta^2 \Esgd \br{\ns{n_{t-1,\eta}^{(k)}}|w_{t-1,\eta}^{(k)}}\\
=& \eta^2\Esgd \br{\ns{x_{i(t-1)}x_{i(t-1)}^\top(w_{t-1,\eta}^{(k)}-\wtrk)-\htrk(w_{t-1,\eta}^{(k)}-\wtrk)}|w_{t-1,\eta}^{(k)}}\\
\leq& \eta^2 \Esgd \br{\ns{x_{i(t-1)}x_{i(t-1)}^\top(w_{t-1,\eta}^{(k)}-\wtrk)}|w_{t-1,\eta}^{(k)}}\\
\leq& \eta^2 \max_{i(t-1)}\ns{x_{i(t-1)}}\ns{w_{t-1,\eta}^{(k)}-\wtrk}_{\htrk}.
\end{align*}
Since $\n{\xtr}\leq \sqrt{L}\sqrt{d},$ we immediately know $\max_{i(t-1)}\n{x_{i(t-1)}}\leq \sqrt{L}\sqrt{d}.$ Therefore, we can bound the noise as follows,
\begin{align*}
\eta^2 \Esgd \br{\ns{n_{t-1,\eta}^{(k)}}|w_{t-1,\eta}^{(k)}}
\leq& \eta^2 \max_{i(t-1)}\ns{x_{i(t-1)}}\ns{w_{t-1,\eta}^{(k)}-\wtrk}_{\htrk}\\
\leq& L^2\eta^2 d\ns{w_{t-1,\eta}^{(k)}-\wtrk}.
\end{align*}
As long as $\eta\leq \frac{1}{2L^3d},$ we have 
$$(1-\eta L)\ns{w_{t-1,\eta}^{(k)}-\wtrk} \leq \Esgd \br{ \ns{\wtk-\wtrk}| w_{t-1,\eta}^{(k)}}\leq (1-\frac{\eta}{2L})\ns{w_{t-1,\eta}^{(k)}-\wtrk}.$$
This further implies
$$(1-\eta L)^t \ns{\wtr} \leq \Esgd \ns{\wtk-\wtrk}\leq (1-\frac{\eta}{2L})^t \ns{\wtr}.$$
Let $\eta_2 := \frac{1}{2L^3 d},$ we have 
$$\dtrk \leq \frac{L}{2}(1-\frac{1}{4L^4 d})^t \ns{\wtr}$$
Let $\eta_1 :=  \frac{1}{6 L^5 d},$ for all $\eta\in[0,\eta_1]$ we have 
$$\dtrk \geq \frac{1}{2L}(1-\frac{1}{6L^4 d})^t \ns{\wtr}.$$
As long as $t\geq c_2d$ for certain constant $c_2,$ we know 
$$\frac{1}{2L}(1-\frac{1}{6L^4 d})^t \ns{\wtr} > \frac{L}{2}(1-\frac{1}{4L^4 d})^t \ns{\wtr}.$$
As this holds for all $k\in[m]$ and $\hftr = 1/m\sum_{i=1}^m \dtrk,$
we know the optimal step size $\etatr$ is within $[\frac{1}{6 L^5 d}, \frac{1}{2L^3 d}].$
\end{proofof}

We rely the following technical lemma to prove Lemma~\ref{lem:diverge_SGD}. 

\begin{lemma}\label{lem:crossing_train_union}
Suppose $\sigma$ is a constant.
Given any $\epsilon>0$, with probability at least $1-O(1/\epsilon)\expepsd$,
$$\left|\inner{\bt\wtr^* - w^*}{\bt(\xtr)^\dagger\xitr}\right|\leq \epsilon,$$
for all $\eta\in[0,\frac{1}{2L^3 d}].$
\end{lemma}

\begin{proofof}{Lemma~\ref{lem:crossing_train_union}}
By Lemma~\ref{lem:isotropic}, with probability at least $1-\expd,$ $\sqrt{d}/\sqrt{L} \leq \sigma_i(\xtr)\leq  \sqrt{Ld}$ and $1/L \leq \lambda_i(\htr) \leq L$ for all $i\in[n]$. Therefore $\n{[(\xtr)^\dagger]^\top \bt(\bt\wtr^* - w^*)}\leq 2\sqrt{L}/\sqrt{d}.$ Notice that $\xitr$ is independent with $[(\xtr)^\dagger]^\top \bt(\bt\wtr^* - w^*).$ By Hoeffding's inequality, with probability at least $1-\expepsd,$ 
$$\absr{\inner{[(\xtr)^\dagger]^\top \bt(\bt\wtr^* - w^*)}{\xitr}}\leq \epsilon.$$

Next, we construct an $\epsilon$-net for $\eta$ and show the crossing term is small for all $\eta\in[0,\frac{1}{2L^3 d}].$ For simplicity, denote $g(\eta):=\inner{\bt\wtr^* -w^*}{\bt(\xtr)^\dagger\xitr}.$
Taking the derivative of $g(\eta),$ we have
\begin{align*}
g'(\eta)=&t\inner{\htr(I-\eta\htr)^{t-1}\wtr^*}{\bt(\xtr)^\dagger\xitr}\\
&+t\inner{\bt\wtr^* -w^*}{\htr(I-\eta \htr)^{t-1}(\xtr)^\dagger\xitr}
\end{align*}

According to Lemma~\ref{lem:norm_vector}, we know with probability at least $1-\exp(-\Omega(d)),$ $\n{\xitr}\leq \sqrt{d}\sigma.$ Therefore, the derivative $g'(\eta)$ can be bounded as follows,
\begin{align*}
|g'(\eta)|= O(1)t(1-\frac{\eta}{L})^{t-1}
\end{align*}

Similar as in Lemma~\ref{lem:crossing_term2}, there exists an $\epsilon$-net $N_\epsilon$ with size $O(1/\epsilon)$
such that for any $\eta\in[0,\frac{1}{3L^3 d}],$ there exists $\eta'\in N_\epsilon$ with $|g(\eta)-g(\eta')|\leq \epsilon.$
Taking a union bound over $N_\epsilon,$ we have with probability at least $1-O(1/\epsilon)\expepsd$, for every $\eta\in N_\epsilon,$
$$\left|\inner{\bt\wtr^* - w^*}{\bt(\xtr)^\dagger\xitr}\right|\leq \epsilon.$$
which implies for every $\eta\in[0,\frac{1}{3L^3 d}].$
$$\left|\inner{\bt\wtr^* - w^*}{\bt(\xtr)^\dagger\xitr}\right|\leq 2\epsilon.$$
Changing $\epsilon$ to $\epsilon'/2$ finishes the proof.
\end{proofof}

\begin{proofof}{Lemma~\ref{lem:diverge_SGD}}
According to Lemma~\ref{lem:isotropic} and Lemma~\ref{lem:norm_vector}, we know with probability at least $1-\expd,$ $\sqrt{d}/\sqrt{L} \leq \sigma_i(\xtr)\leq  \sqrt{Ld}$ and $1/L \leq \lambda_i(\htr) \leq L$ for all $i\in[n]$ and $\sqrt{d}\sigma/4 \leq \n{\xitr}\leq \sqrt{d}\sigma$. We assume these properties hold in the proof and take a union bound at the end.

Recall that $\Esgd \ns{\wt - w^*}$ can be lower bounded as follows,
\begin{align*}
\Esgd \ns{\wt - w^*} 
=& \Esgd \ns{\bt(\wtr^*+(\xtr)^\dagger\xitr)-\eta\sum_{\tau=0}^{t-1}(I-\eta\htr)^{t-1-\tau}n_{\tau,\eta}-w^*}\\
\geq& \ns{\bt(\wtr^*+(\xtr)^\dagger\xitr)-w^*}\\
\geq& \ns{B_{t,\eta}(\xtr)^\dagger\xitr}+2\inner{B_{t,\eta}\wtr^* - w^*}{B_{t,\eta}(\xtr)^\dagger\xitr}
\end{align*}

For any $\eta \in[\frac{1}{6 L^5 d}, \frac{1}{2 L^3d}]$, we can lower bound the first term as follows,
\begin{align*}
\ns{B_{t,\eta}(\xtr)^\dagger\xitr}
\geq& \pr{1-\exp\pr{-\frac{\eta t}{L}}}^2\frac{\sigma^2}{16L}\\
\geq& \pr{1-\exp\pr{-\frac{t}{6L^6 d}}}^2\frac{\sigma^2}{16L}\\
\geq& \frac{\sigma^2}{64L},
\end{align*}
where the last inequality holds as long as $t\geq c_2d$ for certain constant $c_2.$

Choosing $\epsilon=\frac{\sigma^2}{256L}$ in Lemma~\ref{lem:crossing_train_union}, we know with probability at least $1-\expd,$
$$\absr{\inner{\bt\wtr^* -w^*}{\bt(\xtr)^\dagger\xitr}}\leq \frac{\sigma^2}{256L},$$
for all $\eta\in[0,\frac{1}{2L^3d}].$

Overall, we have $\Esgd \ns{\wt - w^*} \geq \frac{\sigma^2}{128 L}.$ Taking a union bound over all the bad events, we know this happens with probability at least $1-\expd.$
\end{proofof}
\subsection{Train-by-validation (SGD)}\label{sec:proofs_valid_SGD}
Recall Theorem~\ref{thm:valid_SGD} as follows. 

\thmValidSGD*

To prove Theorem~\ref{thm:valid_SGD}, we first study the behavior of the population meta objective $\bfva$. That is,
\begin{align*}
\bfva(\eta) := \Ep \dva=&\Ep \Esgd \frac{1}{2}\ns{\wt - w^* -(\xva)^\dagger \xiva}_{\hva}\\
=&\Ep \Esgd \frac{1}{2}\ns{\wt - w^*}+\frac{\sigma^2}{2}.
\end{align*}
We show that the optimal step size for the population meta objective $\bfva$ is $\Theta(1/t)$ and $\Ep \Esgd \ns{\wt - w^*}=\ns{w^*}-\Omega(1)$ under the optimal step size. 
\begin{restatable}
{lemma}{expectationSGD}\label{lem:expectation_SGD}
Suppose $\sigma$ is a large constant $c_1$. Assume $t\geq c_2 d^2\log^2(d), d\geq c_4$ for some constants $c_2,c_4.$ There exist $\eta_1,\eta_2,\eta_3=\Theta(1/t)$ with $\eta_1<\eta_2<\eta_3$ and constant $c_5$ such that 
\begin{align*}
\bfva(\eta_2)&\leq \frac{1}{2}\ns{w^*}-\frac{9}{10}C+\frac{\sigma^2}{2}\\
\bfva(\eta)&\geq \frac{1}{2}\ns{w^*}-\frac{6}{10}C+\frac{\sigma^2}{2},\forall \eta\in[0,\eta_1]\cup [\eta_3,\frac{1}{c_5d^2\log^2(d)}] 
\end{align*}
where $C$ is a positive constant.
\end{restatable}

In order to relate the behavior of $\bfva$ to $\hfva,$ we show a generalization result from $\hfva$ to $\bfva$ for $\eta\in[0,\frac{1}{c_5 d^2\log^2(d/\epsilon)}].$

\begin{restatable}
{lemma}{generalizationSGD}\label{lem:generalization_SGD}
For any $1>\epsilon>0,$ assume $\sigma$ is a constant and $d\geq c_4\log(1/\epsilon)$ for some constant $c_4.$ There exists constant $c_5$ such that with probability at least $1-O(1/\epsilon)\expepsm$, 
$$|\hfva(\eta)-\bfva(\eta)|\leq \epsilon,$$
for all $\eta\in[0,\frac{1}{c_5 d^2\log^2(d/\epsilon)}].$
\end{restatable}

Combining Lemma~\ref{lem:expectation_SGD} and Lemma~\ref{lem:generalization_SGD}, we give the proof of Theorem~\ref{thm:valid_SGD}.

\begin{proofof}{Theorem~\ref{thm:valid_SGD}}
The proof is almost the same as in the GD setting (Theorem~\ref{thm:valid_GD}). We omit the details here.
\end{proofof}

\subsubsection{Behavior of $\bfva$ for $\eta\in[0,\frac{1}{c_5d^2\log^2 d}]$}
In this section, we give the proof of Lemma~\ref{lem:expectation_SGD}. Recall the lemma as follows,
\expectationSGD*

Recall that $\bfva(\eta)=\Ep\Esgd 1/2\ns{\wt-w^*}+\sigma^2/2.$ Denote $Q(\eta):=\Esgd 1/2\ns{\wt-w^*}$. Recall that we truncate the SGD sequence once the weight norm exceeds $4\sqrt{L}\sigma.$ Due to the truncation, the expectation of $1/2\ns{\wt-w^*}$ over SGD noise is very tricky to analyze. 

Instead, we define an auxiliary sequence $\{w_{\tau,\eta}'\}$ that is obtained by running SGD on task $P$ without truncation and we first study $Q'(\eta):=1/2\Esgd \ns{\wt'-w^*}.$ In Lemma~\ref{lem:whp_valid_SGD}, we show that with high probability in the sampling of task $P$, the minimizer of $Q'(\eta)$ is $\Theta(1/t).$ The proof is very similar as the proof of Lemma~\ref{lem:whp_valid} except that we need to bound the SGD noise at step size $\eta_2$. We defer the proof into Section~\ref{sec:technical_valid_SGD}.

\begin{lemma}\label{lem:whp_valid_SGD}
Given a task $P$, let $\{w_{\tau,\eta}'\}$ be the weight obtained by running SGD on task $P$ without truncation. Choose $\sigma$ as a large constant $c_1$. Assume unroll length $t\geq c_2d$ for some constant $c_2$. With probability at least $1-\expd$ over the sampling of task $P,$ $\sqrt{d}/\sqrt{L} \leq \sigma_i(\xtr)\leq  \sqrt{Ld}$ and $1/L \leq \lambda_i(\htr) \leq L$ for all $i\in[n]$ and $\sqrt{d}\sigma/4 \leq \n{\xitr}\leq \sqrt{d}\sigma$ and there exists $\eta_1,\eta_2,\eta_3=\Theta(1/t)$ with $\eta_1<\eta_2<\eta_3$ such that 
\begin{align*}
Q'(\eta_2):=1/2\Esgd \ns{w'_{t,\eta_2} - w^*}&\leq \frac{1}{2}\ns{w^*}-C\\
Q'(\eta):=1/2\Esgd \ns{\wt' - w^*}&\geq \frac{1}{2}\ns{w^*}-\frac{C}{2},\forall \eta\in[0,\eta_1]\cup [\eta_3,1/L] 
\end{align*}
where $C$ is a positive constant.
\end{lemma}

To relate the behavior of $Q'(\eta)$ defined on $\{w_{\tau,\eta}'\}$ to the behavior of $Q(\eta)$ defined on $\{w_{\tau,\eta}\}$. We show when the step size is small enough, the SGD sequence gets truncated with very small probability so that sequence $\{w_{\tau,\eta}\}$ almost always coincides with sequence $\{w_{\tau,\eta}'\}$. The proof of Lemma~\ref{lem:q_n_qprime} is deferred into Section~\ref{sec:technical_valid_SGD}.

\begin{lemma}\label{lem:q_n_qprime}
Given a task $P$, assume $\sqrt{d}/\sqrt{L} \leq \sigma_i(\xtr)\leq  \sqrt{Ld}$ and $1/L \leq \lambda_i(\htr) \leq L$ for all $i\in[n]$ and $\sqrt{d}\sigma/4 \leq \n{\xitr}\leq \sqrt{d}\sigma$. Given any $\epsilon>0,$ suppose $\eta\leq \frac{1}{c_5 d^2\log^2(d/\epsilon)}$ for some constant $c_5$, we have 
$$\absr{Q(\eta)-Q'(\eta)}\leq \epsilon.$$
\end{lemma}

Combining Lemma~\ref{lem:whp_valid_SGD} and Lemma~\ref{lem:q_n_qprime}, we give the proof of lemma~\ref{lem:expectation_SGD}.

\begin{proofof}{Lemma~\ref{lem:expectation_SGD}}
Recall that we define $Q(\eta):=1/2\Esgd \ns{\wt - w^*}$ and $Q'(\eta)=1/2\Esgd \ns{\wt' - w^*}.$ Here, $\{w_{\tau,\eta}'\}$ is a SGD sequence running on task $P$ without truncation.

According to Lemma~\ref{lem:whp_valid_SGD}, with probability at least $1-\expd$ over the sampling of task $P,$ $\sqrt{d}/\sqrt{L} \leq \sigma_i(\xtr)\leq  \sqrt{Ld}$ and $1/L \leq \lambda_i(\htr) \leq L$ for all $i\in[n]$ and $\sqrt{d}\sigma/4 \leq \n{\xitr}\leq \sqrt{d}\sigma$ and there exists $\eta_1,\eta_2,\eta_3=\Theta(1/t)$ with $\eta_1<\eta_2<\eta_3$ such that
\begin{align*}
Q'(\eta_2)&\leq \frac{1}{2}\ns{w^*}-C\\
Q'(\eta)&\geq \frac{1}{2}\ns{w^*}-\frac{C}{2},\forall \eta\in[0,\eta_1]\cup [\eta_3,1/L] 
\end{align*}
where $C$ is a positive constant. Call this event $\calE.$ Suppose the probability that $\calE$ happens is $1-\delta$. We can write $\Ep Q(\eta)$ as follows,
\begin{align*}
\Ep Q(\eta) = \Ep[Q(\eta)|\calE]\Pr[\calE]+\Ep[Q(\eta)|\bar{\calE}]\Pr[\bar{\calE}].
\end{align*}
According to the algorithm, we know $\n{\wt}$ is always bounded by $4\sqrt{L}\sigma.$ Therefore, $Q(\eta):=1/2\ns{\wt-w^*}\leq 13L\sigma^2.$ By Lemma~\ref{lem:q_n_qprime}, we know conditioning on $\calE,$ $|Q(\eta)-Q'(\eta)|\leq \epsilon$ for any $\eta\leq \frac{1}{c_5 d^2\log^2(d/\epsilon)}.$ As long as $t\geq c_2 d^2\log^2(d/\epsilon)$ for certain constant $c_2,$ we know $\eta_3\leq \frac{1}{c_5 d^2\log^2(d/\epsilon)}.$

When $\eta=\eta_2,$ we have
\begin{align*}
\Ep Q(\eta_2) \leq& \pr{Q'(\eta_2)+\epsilon}(1-\delta)+13L\sigma^2\delta\\
\leq& \pr{\frac{1}{2}\ns{w^*}-C+\epsilon}(1-\delta)+13L\sigma^2\delta\\
\leq& \frac{1}{2}\ns{w^*}-C+13L\sigma^2\delta+\epsilon
\leq \frac{1}{2}\ns{w^*}-\frac{9C}{10},
\end{align*}
where the last inequality assumes $\delta\leq \frac{C}{260L\sigma^2}$ and $\epsilon\leq \frac{C}{20}.$

When $\eta\in[0,\eta_1]\cup [\eta_3,\frac{1}{c_5 d^2\log^2(d/\epsilon)}],$ we have
\begin{align*}
\Ep Q(\eta_2)\geq& \pr{Q'(\eta)-\epsilon}(1-\delta)-13L\sigma^2\delta\\
\geq& \pr{\frac{1}{2}\ns{w^*}-\frac{C}{2}-\epsilon}(1-\delta)-13L\sigma^2\delta\\
\geq& \frac{1}{2}\ns{w^*}-\frac{C}{2}-\frac{\delta}{2}-13L\sigma^2\delta-\epsilon
\geq \frac{1}{2}\ns{w^*}-\frac{6C}{10},
\end{align*}
where the last inequality holds as long as $\delta\leq \frac{C}{280L\sigma^2}$ and $\epsilon\leq \frac{C}{20}.$

According to Lemma~\ref{lem:whp_valid_SGD}, we know $\delta\leq \expd.$ Therefore, the conditions for $\delta$ can be satisfied as long as $d$ is larger than certain constant. The condition on $\epsilon$ can be satisfied as long as $\eta\leq \frac{1}{c_5 d^2\log^2(d)}$ for some constant $c_5 $.
\end{proofof}

\subsubsection{Generalization for $\eta\in[0,\frac{1}{c_5d^2\log^2 d}]$}
In this section, we prove Lemma~\ref{lem:generalization_SGD} by showing that $\hfva(\eta)$ is point-wise close to $\bfva(\eta)$ for all $\eta\in[0,\frac{1}{c_5 d^2\log^2(d/\epsilon)}].$ Recall Lemma~\ref{lem:generalization_SGD} as follows.
\generalizationSGD*

In order to prove Lemma~\ref{lem:generalization_SGD}, we first show that for a fixed $\eta$ with high probability $\hfva(\eta)$ is close to $\bfva(\eta)$. Similar as in Lemma~\ref{lem:genera_fixed_eta_valid_GD}, we can still show that each $\dva$ is $O(1)$-subexponential. The proof is deferred into Section~\ref{sec:technical_valid_SGD}.

\begin{lemma}\label{lem:genera_fixed_eta_SGD}
Suppose $\sigma$ is a constant. Given any $1>\epsilon>0,$ for any fixed $\eta$ with probability at least $1-\expepsm,$
$$\absr{\hfva(\eta)-\bfva(\eta) }\leq \epsilon.$$
\end{lemma}

Next, we show that there exists an $\epsilon$-net for $\bfva$ with size $O(1/\epsilon).$ By $\epsilon$-net, we mean there exists a finite set $N_\epsilon$ of step sizes such that $|\bfva(\eta)-\bfva(\eta')|\leq \epsilon$ for any $\eta$ and $\eta'\in\arg\min_{\eta\in N_\epsilon}|\eta-\eta'|.$ The proof is very similar as in Lemma~\ref{lem:eps_net_expect_valid_GD}. We defer the proof of Lemma~\ref{lem:eps_net_expect_SGD} into Section~\ref{sec:technical_valid_SGD}.

\begin{lemma}\label{lem:eps_net_expect_SGD}
Suppose $\sigma$ is a constant. For any $1>\epsilon>0,$ assume $d\geq c_4\log(1/\epsilon)$ for some $c_4.$ 
There exists constant $c_5$ and an $\epsilon$-net $N_\epsilon\subset [0,\frac{1}{c_5 d^2\log^2(d/\epsilon)}]$ for $\bfva$ with $|N_\epsilon|=O(1/\epsilon).$ That means, for any $\eta\in[0,\frac{1}{c_5 d^2\log^2(d/\epsilon)}],$
$$|\bfva(\eta)-\bfva(\eta')|\leq \epsilon,$$
for $\eta'\in\arg\min_{\eta\in N_\epsilon}|\eta-\eta'|.$
\end{lemma}

Next, we show that with high probability, there also exists an $\epsilon$-net for $\hfva$ with size $O(1/\epsilon).$ The proof is very similar as the proof of Lemma~\ref{lem:eps_net_empirical_valid_GD}. We defer the proof into Section~\ref{sec:technical_valid_SGD}.

\begin{lemma}\label{lem:eps_net_empirical_SGD}
Suppose $\sigma$ is a constant. For any $1>\epsilon>0,$ assume $d\geq c_4 \log(1/\epsilon)$ for some $c_4.$ With probability at least $1-\expepsm$, there exists constant $c_5$ and an $\epsilon$-net $N_\epsilon'\subset [0,\frac{1}{c_5 d^2\log^2(d/\epsilon)}]$ for $\hfva$ with $|N_\epsilon|=O(1/\epsilon).$ That means, for any $\eta\in[0,\frac{1}{c_5 d^2\log^2(d/\epsilon)}],$
$$|\hfva(\eta)-\hfva(\eta')|\leq \epsilon,$$
for $\eta'\in\arg\min_{\eta\in N_\epsilon}|\eta-\eta'|.$
\end{lemma}

Combing Lemma~\ref{lem:genera_fixed_eta_SGD}, Lemma~\ref{lem:eps_net_expect_SGD} and Lemma~\ref{lem:eps_net_empirical_SGD}, now we give the proof of Lemma~\ref{lem:generalization_SGD}.

\begin{proofof}{Lemma~\ref{lem:generalization_SGD}}
The proof is almost the same as the proof of Lemma~\ref{lem:genera_valid_GD}. We omit the details here.
\end{proofof}

\subsubsection{Proofs of Technical Lemmas}\label{sec:technical_valid_SGD}
In Lemma~\ref{lem:SGD_noise}, we show when the step size is small, the expected SGD noise square is well bounded. The proof follows from the analysis in Lemma~\ref{lem:large_step_SGD}.

\begin{lemma}\label{lem:SGD_noise}
Let $\{w_{\tau,\eta}'\}$ be an SGD sequence running on task $P$ without truncation. Let $n'_{\tau,\eta}$ be the SGD noise at $w_{\tau,\eta}'$. Assume $\sqrt{d}/\sqrt{L} \leq \sigma_i(\xtr)\leq \sqrt{L}\sqrt{\sigma}$ for all $i\in[n]$ and $\n{\xitr}\leq \sqrt{d}\sigma$. Suppose $\eta\in[0,\frac{1}{2L^3 d}],$ we have 
$$\Esgd \ns{n'_{\tau,\eta}}\leq 4L^3 \sigma^2 d$$
for all $\tau\leq t.$
\end{lemma}

\begin{proofof}{Lemma~\ref{lem:SGD_noise}}
Similar as the analysis in Lemma~\ref{lem:large_step_SGD}, for $\eta\leq \frac{1}{2L^3 d},$
we have 
$$\Esgd \br{\ns{n'_{\tau,\eta}}|w'_{\tau-1,\eta}}\leq L^2d \ns{w'_{\tau-1,\eta}-\wtr}.$$
and
$$\Esgd \ns{w'_{\tau-1,\eta}-\wtr}\leq (1-\frac{\eta}{2L})^{\tau-1} \ns{\wtr}\leq \ns{\wtr^*+(\xtr)^\dagger\xitr}\leq 4L\sigma^2.$$

Therefore, we have
$$\Esgd \ns{n'_{\tau,\eta}}\leq L^2 d\Esgd \ns{w_{\tau,\eta}'-\wtr}\leq 4L^3\sigma^2 d.$$
\end{proofof}

\begin{proofof}{Lemma~\ref{lem:whp_valid_SGD}}
We can expand $Q'(\eta)$ as follows,
\begin{align*}
Q'(\eta):=& \frac{1}{2}\Esgd \ns{\wt'-w^*}\\
=& \frac{1}{2}\Esgd \ns{\bt\wtr^*+\bt(\xtr)^\dagger\xitr-\eta\sum_{\tau=0}^{t-1}(I-\eta\htr)^{t-1-\tau}n'_{\tau,\eta}-w^*}\\
=& \frac{1}{2}\ns{\bt\wtr^* -w^*}+\frac{1}{2}\ns{\bt(\xtr)^\dagger\xitr}+\frac{\eta^2}{2}\Esgd \ns{\sum_{\tau=0}^{t-1}(I-\eta\htr)^{t-1-\tau}n'_{\tau,\eta}}\\
&+\inner{\bt\wtr^* -w^*}{ \bt(\xtr)^\dagger\xitr}
\end{align*}
Denote
 $$G(\eta):=\frac{1}{2}\ns{\bt\wtr^* -w^*}+\frac{1}{2}\ns{\bt(\xtr)^\dagger\xitr}+\frac{\eta^2}{2}\Esgd \ns{\sum_{\tau=0}^{t-1}(I-\eta\htr)^{t-1-\tau}n'_{\tau,\eta}}.$$
We first show that with probability at least $1-\expd,$ there exist $\eta_1,\eta_2,\eta_3=\Theta(1/t)$ with $\eta_1<\eta_2<\eta_3$ such that $G(\eta_2)\leq 1/2\ns{w^*}-5C/4$ and $G(\eta)\geq 1/2\ns{w^*}-C/4$ for all $\eta\in[0,\eta_1]\cup [\eta_3,1/L]$.

According to Lemma~\ref{lem:isotropic}, we know with probability at least $1-\expd,$ $\sqrt{d}/\sqrt{L} \leq \sigma_i(\xtr)\leq  \sqrt{L}\sqrt{d}$ and $1/L \leq \lambda_i(\htr) \leq L$ for all $i\in[n].$ According to Lemma~\ref{lem:norm_vector}, we know with probability at least $1-\expd,$ $\sqrt{d}\sigma/4 \leq \n{\xitr}\leq \sqrt{d}\sigma$.

\paragraph{Upper bounding $G(\eta_2)$:} We can expand $G(\eta)$ as follows:
\begin{align*}
G(\eta):=& \frac{1}{2}\ns{B_{t,\eta}\wtr^*-w^*}+\frac{1}{2}\ns{B_{t,\eta}(\xtr)^\dagger\xitr}+\frac{\eta^2}{2}\Esgd \ns{\sum_{\tau=0}^{t-1}(I-\eta\htr)^{t-1-\tau}n'_{\tau,\eta}}\\
=& \frac{1}{2}\ns{w^*}+ \frac{1}{2}\ns{B_{t,\eta}\wtr^*}+\frac{1}{2}\ns{B_{t,\eta}(\xtr)^\dagger\xitr}+\frac{\eta^2}{2}\Esgd \ns{\sum_{\tau=0}^{t-1}(I-\eta\htr)^{t-1-\tau}n'_{\tau,\eta}}\\
&- \inner{B_{t,\eta}\wtr^*}{w^*}.
\end{align*}
Same as in Lemma~\ref{lem:whp_valid}, we know 
$
\frac{1}{2}\ns{B_{t,\eta}\wtr^*}+\frac{1}{2}\ns{B_{t,\eta}(\xtr)^\dagger\xitr}
\leq L^3\eta^2 t^2\sigma^2.
$
For the SGD noise, by Lemma~\ref{lem:SGD_noise} we know $\Esgd \ns{n'_{\tau,\eta}}\leq 4L^3\sigma^2 d$ for all $\tau\leq t$ as long as $\eta\leq \frac{1}{2L^3d}.$ 
Therefore,
\begin{align*}
\frac{\eta^2}{2}\Esgd \ns{\sum_{\tau=0}^{t-1}(I-\eta\htr)^{t-1-\tau}n'_{\tau,\eta}}
\leq \frac{\eta^2}{2} \sum_{\tau=0}^{t-1}\Esgd \ns{n'_{\tau,\eta}}
\leq 2L^3\eta^2\sigma^2 dt\leq 2L^3\eta^2\sigma^2 t^2,
\end{align*}
where the last inequality assumes $t\geq d.$
According to Lemma~\ref{lem:improvement}, for any fixed $\eta\in [0,L/t]$, with probability at least $1-\exp(-\Omega(d))$ over $\xtr$,
$$\inner{B_{t,\eta}\wtr^*}{ w^*} \geq \frac{\eta t}{16L}.$$

Therefore, for any step size $\eta\leq \frac{1}{2L^3 d},$
\begin{align*}
G(\eta)\leq \frac{1}{2}\ns{w^*} + 3L^3\eta^2\sigma^2 t^2 - \frac{\eta t}{16L}\leq \frac{1}{2}\ns{w^*}- \frac{\eta t}{32L},
\end{align*}
where the second inequality holds as long as $\eta\leq \frac{1}{96 L^4 \sigma^2 t}.$ Choosing $\eta_2:= \frac{1}{96 L^4 \sigma^2 t}$ that is smaller than $\frac{1}{2L^3 d}$ assuming $t\geq d.$ Then, we have 
$$G(\eta_2)\leq \frac{1}{2}\ns{w^*}- \frac{5C}{4},$$
where constant $C=\frac{1}{3072 L^5\sigma^2}.$

\paragraph{Lower bounding $G(\eta)$ for $\eta\in[0,\eta_1]:$} Now, we prove that there exists $\eta_1=\Theta(1/t)$ with $\eta_1 < \eta_2$ such that for any $\eta\in[0,\eta_1], G(\eta)\geq \frac{1}{2}\ns{w^*}-\frac{C}{4}.$
Recall that 
\begin{align*}
G(\eta)
=& \frac{1}{2}\ns{w^*}+ \frac{1}{2}\ns{B_{t,\eta}\wtr^*}+\frac{1}{2}\ns{B_{t,\eta}(\xtr)^\dagger\xitr}+\frac{\eta^2}{2}\Esgd \ns{\sum_{\tau=0}^{t-1}(I-\eta\htr)^{t-1-\tau}n'_{\tau,\eta}}\\
&- \inner{B_{t,\eta}\wtr^*}{w^*}.\\
\geq& \frac{1}{2}\ns{w^*}- \inner{B_{t,\eta}\wtr^*}{w^*}.
\end{align*}
Same as in Lemma~\ref{lem:whp_valid}, by choosing $\eta_1= \frac{C}{4L t},$ we have for any $\eta\in[0,\eta_1],$
$$G(\eta)\geq \frac{1}{2}\ns{w^*}- \frac{C}{4}.$$

\paragraph{Lower bounding $G(\eta)$ for $\eta\in [\eta_3,1/L]$:} Now, we prove that there exists $\eta_3 = \Theta(1/t)$ with $\eta_3 > \eta_2$ such that for all $\eta\in[\eta_3,1/L]$,
$$G(\eta)\geq \frac{1}{2}\ns{w^*}- \frac{C}{4}.$$
Recall that 
\begin{align*}
G(\eta)
=& \frac{1}{2}\ns{B_{t,\eta}\wtr^*-w^*}+\frac{1}{2}\ns{B_{t,\eta}(\xtr)^\dagger\xitr}+\frac{\eta^2}{2}\Esgd \ns{\sum_{\tau=0}^{t-1}(I-\eta\htr)^{t-1-\tau}n'_{\tau,\eta}}\\
\geq& \frac{1}{2}\ns{B_{t,\eta}(\xtr)^\dagger\xitr}.
\end{align*}
Same as in Lemma~\ref{lem:whp_valid}, by choosing $\eta_3=\log(2)L/t,$ as long as $\sigma\geq 8\sqrt{L},$ we have
$$G(\eta)\geq \frac{1}{2}\ns{w^*}$$
for all $\eta\in[\eta_3,1/L].$ Note $\eta_3\leq 1/L$ as long as $t\geq \log(2)L^2.$

Overall, we have shown that there exist $\eta_1,\eta_2,\eta_3=\Theta(1/t)$ with $\eta_1<\eta_2<\eta_3$ such that $G(\eta_2)\leq 1/2\ns{w^*}-5C/4$ and $G(\eta)\geq 1/2\ns{w^*}-C/4$ for all $\eta\in[0,\eta_1]\cup [\eta_3,1/L]$. Recall that $Q'(\eta) = G(\eta)+\inner{B_{t,\eta}\wtr^*-w^*}{B_{t,\eta}(\xtr)^\dagger\xitr}.$ Choosing $\epsilon=C/4$ in Lemma~\ref{lem:crossing_term2}, we know with probability at least $1-\expd,$ $\absr{\inner{B_{t,\eta}\wtr^*-w^*}{B_{t,\eta}(\xtr)^\dagger\xitr}}\leq C/4$ for all $\eta\in[0,1/L].$ Therefore, we know $Q'(\eta_2)\leq 1/2\ns{w^*}-C$ and $Q'(\eta)\geq 1/2\ns{w^*}-C/2$ for all $\eta\in[0,\eta_1]\cup [\eta_3,1/L]$.
\end{proofof}

In order to prove Lemma~\ref{lem:q_n_qprime}, we first construct a super-martingale to show that as long as task $P$ is well behaved, with high probability in SGD noise, the weight norm along the trajectory never exceeds $4\sqrt{L}\sigma.$

\begin{lemma}\label{lem:martingale}
Assume $\sqrt{d}/\sqrt{L} \leq \sigma_i(\xtr)\leq  \sqrt{Ld}$ and $1/L \leq \lambda_i(\htr) \leq L$ for all $i\in[n]$ and $\sqrt{d}\sigma/4 \leq \n{\xitr}\leq \sqrt{d}\sigma$. Given any $1>\delta>0,$ suppose $\eta\leq \frac{1}{c_5 d^2\log^2(d/\delta)}$ for some constant $c_5$, with probability at least $1-\delta$ in the SGD noise,
$$\n{w_{\tau,\eta}'}< 4\sqrt{L}\sigma$$
for all $\tau \leq t.$
\end{lemma}

\begin{proofof}{Lemma~\ref{lem:martingale}}
According to the proofs of Lemma~\ref{lem:SGD_noise}, as long as $\eta\leq \frac{1}{2L^3 d},$ we have 
$$\Esgd \br{ \ns{\wt'-\wtr}| w_{t-1,\eta}'}\leq (1-\frac{\eta}{2L})\ns{w_{t-1,\eta}'-\wtr}.$$

Since $\log$ is a concave function, by Jenson's inequality, we know 
\begin{align*}
&\Esgd \br{\log \ns{\wt'-\wtr}| w_{t-1,\eta}'} \\
\leq& \log \Esgd \br{ \ns{\wt'-\wtr}| w_{t-1,\eta}'}\leq \log \ns{w_{t-1,\eta}'-\wtr} + \log(1-\frac{\eta}{2L}).
\end{align*}
Defining $G_t = \log \ns{\wt'-\wtr} -t\log(1-\frac{\eta}{2L}),$ we know $G_t$ is a super-martingale. Next, we bound the martingale differences.

We can bound $|G_t-\Esgd [G_t|w_{t-1,\eta}']|$ as follows,
\begin{align*}
|G_t-\Esgd [G_t|w_{t-1,\eta}']|\leq \max_{n'_{t-1,\eta},n''_{t-1,\eta}}\log\pr{\frac{\ns{(I-\eta\htr)(w_{t-1,\eta}'-\wtr)-\eta n_{t-1,\eta}'}}{\ns{(I-\eta\htr)(w_{t-1,\eta}'-\wtr)-\eta n_{t-1,\eta}''}}}
\end{align*}
We can expand $\ns{(I-\eta\htr)(w_{t-1,\eta}'-\wtr)-\eta n_{t-1,\eta}'}$ as follows,
\begin{align*}
&\ns{(I-\eta\htr)(w_{t-1,\eta}'-\wtr)-\eta n_{t-1,\eta}'}\\
=& \ns{(I-\eta\htr)(w'_{t-1,\eta}-\wtr)} -2\eta\inner{ n'_{t-1,\eta}}{(I-\eta\htr)(w'_{t-1,\eta}-\wtr)}+\eta^2\ns{n'_{t-1,\eta}}
\end{align*}
We can bound the norm of the noise as follows,
\begin{align*}
\n{n'_{t-1,\eta}}=& \n{x_{i(t-1)}x_{i(t-1)}^\top(w'_{t-1,\eta}-\wtr)-\htr(w'_{t-1,\eta}-\wtr)}\\
\leq& \n{x_{i(t-1)}x_{i(t-1)}^\top(w'_{t-1,\eta}-\wtr)}+\n{\htr(w'_{t-1,\eta}-\wtr)}\\
\leq& \pr{Ld+L}\n{w'_{t-1,\eta}-\wtr}\leq 2Ld\n{w'_{t-1,\eta}-\wtr},
\end{align*}
where the second inequality uses $\n{x_{i(t-1)}}\leq \sqrt{Ld}$. Therefore, we have 
\begin{align*}
&\absr{2\eta\inner{ n'_{t-1,\eta}}{(I-\eta\htr)(w'_{t-1,\eta}-\wtr)}}\leq 4L\eta d \ns{w'_{t-1,\eta}-\wtr},\\
&\eta^2\ns{n'_{t-1,\eta}}\leq 4L^2\eta^2 d^2 \ns{w'_{t-1,\eta}-\wtr}.
\end{align*}
This further implies,
\begin{align*}
&|G_t-\Esgd [G_t|w'_{t-1,\eta}]|\\
\leq& \log\pr{\frac{\ns{(I-\eta\htr)(w'_{t-1,\eta}-\wtr)}+\pr{4L\eta d+4L^2\eta^2 d^2 }\ns{w'_{t-1,\eta}-\wtr}}{\ns{(I-\eta\htr)(w'_{t-1,\eta}-\wtr)}-4L \eta d\ns{w'_{t-1,\eta}-\wtr}}}\\
\leq& \log\pr{1+\frac{8L\eta d+4L^2\eta^2 d^2 }{(1-2L\eta -4L\eta d)}}\leq 16L\eta d+8L^2\eta^2 d^2,
\end{align*}
where the second inequality uses $\ns{(I-\eta\htr)(w_{t-1,\eta}'-\wtr)}\geq (1-2L\eta)\ns{w_{t-1,\eta}'-\wtr}.$ The last inequality assumes $\eta \leq \frac{1}{12L d}$ and uses numerical inequality $\log(1+x)\leq x.$ Assuming $\eta\leq 1/(L d),$ we further have $|G_t-\Esgd [G_t|w'_{t-1,\eta}]|\leq L^2 \eta d.$

By Azuma's inequality, we know with probability at least $1-\delta/t,$
$$G_t\leq G_0 + L^2\sqrt{2t}\eta d\log(t/\delta).$$
Plugging in $G_t = \log \ns{\wt'-\wtr} -t\log(1-\frac{\eta}{2L})$ and $G_0 = \log \ns{w_0-\wtr}=\log \ns{\wtr},$ we have 
\begin{align*}
\log \ns{\wt'-\wtr} 
\leq& \log \ns{\wtr}+t\log(1-\frac{\eta}{2L})+L^2\sqrt{2t}\eta d\log(t/\delta)\\
\leq& \log \ns{\wtr}-\frac{\eta}{2L}t+L^2\sqrt{2t}\eta d\log(t/\delta).
\end{align*} 
This implies,
\begin{align*}
\ns{\wt'-\wtr} \leq& \ns{\wtr}\exp\pr{\eta\pr{-\frac{1}{2L}t+L^2\sqrt{2}\log(t/\delta) d\sqrt{t} }}\\
=& \ns{\wtr}\exp\pr{O(d^2\log^2(d/\delta))\eta}\\
\leq& \ns{\wtr}\exp\pr{2/3},
\end{align*}
where the second inequality assumes $\eta\leq \frac{1}{c_5 d^2log^2(d/\delta)}$ for some constant $c_5 .$ 
Furthermore, since $\n{\wtr}\leq (1+\sqrt{L})\sigma$, we have $\n{\wt'}\leq (1+e^{1/3})\n{\wtr}< 4\sqrt{L}\sigma.$

Overall, we know as long as $\eta\leq \frac{1}{c_5 d^2log^2(d/\delta)}$, with probability at least $1-\delta/t,$ $\n{\wt'}\leq 4\sqrt{L}\sigma.$ Since this analysis also applies to any $\tau\leq t,$ we know for any $\tau,$ with probability at least $1-\delta/t,$ $\n{w_{\tau,\eta}'}< 4\sqrt{L}\sigma.$ Taking a union bound over $\tau\leq t,$ we have with probability at least $1-\delta,$ $\n{w_{\tau,\eta}'} < 4\sqrt{L}\sigma$ for all $\tau\leq t.$
\end{proofof}

\begin{proofof}{Lemma~\ref{lem:q_n_qprime}}
Let $\calE$ be the event that $\n{w_{\tau,\eta}'}< 4\sqrt{L}\sigma$ for all $\tau\leq t.$ We first show that $\Esgd \ns{\wt-w^*}$ is close to $\Esgd  \ns{\wt'-w^*}\indic{\calE}$. It's not hard to verify that 
$$\Esgd \ns{\wt-w^*}=\Esgd  \ns{\wt'-w^*}\indic{\calE}+\ns{u-w^*}\Pr[\bar{\calE}],$$
where $u$ is a fixed vector with norm $4\sqrt{L}\sigma.$
By Lemma~\ref{lem:martingale}, we know $\Pr[\bar{\calE}]\leq \epsilon/(25L\sigma^2)$ as long as $\eta\leq \frac{1}{c_5 d^2\log^2(d/\epsilon)}$ for some constant $c_5 $. Therefore, we have 
\begin{align*}
\absr{\Esgd \ns{\wt-w^*}-\Esgd  \ns{\wt'-w^*}\indic{\calE}}\leq \epsilon.
\end{align*}

Next, we show that $\Esgd  \ns{\wt'-w^*}\indic{\calE}$ is close to $\Esgd  \ns{\wt'-w^*}$. For any $1\leq \tau\leq t,$ let $\calE_\tau$ be the event that $\n{w_{\tau,\eta}'}\geq 4\sqrt{L}\sigma$ and $\n{w_{\tau',\eta}'}< 4\sqrt{L}\sigma$ for all $\tau'<\tau.$ Basically $\calE_\tau$ means the weight norm exceeds the threshold at step $\tau$ for the first time. It's easy to see that $\cup_{\tau=1}^t \calE_\tau = \bar{\calE}.$ Therefore, we have
\begin{align*}
\Esgd  \ns{\wt'-w^*} = \Esgd  \ns{\wt'-w^*}\indic{\calE}+\sum_{\tau=1}^t \Esgd  \ns{\wt'-w^*}\indic{\calE_\tau}.
\end{align*}

Conditioning on $\calE_\tau,$ we know $\n{w_{\tau-1,\eta}'}< 4\sqrt{L}\sigma.$ Since we assume $\frac{\sqrt{d}}{\sqrt{L}}\leq \sigma_i(\xtr)\leq \sqrt{L}\sqrt{d}$ for all $i\in[n]$ and $\xitr\leq \sqrt{d}\sigma,$ we know $\n{\wtr}\leq 2\sqrt{L}\sigma.$ Therefore, we have $\n{w_{\tau-1,\eta}'-\wtr}\leq 6\sqrt{L}\sigma$. Recall the SGD updates,
$$w_{\tau,\eta}'-\wtr = (I-\eta\htr)(w_{\tau-1,\eta}'-\wtr)-\eta n'_{\tau-1,\eta}.$$
For the noise term, we have $\eta\n{n'_{\tau-1,\eta}}\leq 2\eta Ld \n{w_{\tau-1,\eta}'-\wtr}$ that is at most $\n{w_{\tau-1,\eta}'-\wtr}$ assuming $\eta\leq \frac{1}{2Ld}.$
Therefore, we have $\n{w_{\tau,\eta}'-\wtr}\leq 2\n{w_{\tau-1,\eta}'-\wtr}\leq 12\sqrt{L}\sigma.$ Note that event $\calE_{\tau}$ is independent with the SGD noises after step $\tau$. Therefore, according to the previous analysis, we know as long as $\eta\leq \frac{1}{2L^3 d},$
$$\Esgd \br{\ns{\wt'-\wtr}|\calE_\tau}\leq \ns{w_{\tau,\eta}'-\wtr}\leq 2L^2\sigma^2. $$
Then, we can bound $\Esgd \br{\ns{\wt'-w^*}|\calE_\tau}$ as follows,
\begin{align*}
&\Esgd \br{\ns{\wt'-w^*}|\calE_\tau} \\
=& \Esgd \br{\ns{\wt'-\wtr+\wtr-w^*}|\calE_\tau}\\
\leq& \Esgd \br{\ns{\wt'-\wtr}|\calE_\tau}+2\Esgd \br{\n{\wt'-\wtr}|\calE_\tau}\n{\wtr-w^*}+\ns{\wtr-w^*}\\
\leq& 2L^2 \sigma^2 + 2\cdot 2 L\sigma\cdot 3\sqrt{L}\sigma + 9L\sigma^2\leq 3L^2\sigma^2.
\end{align*}
Therefore, we have 
\begin{align*}
\sum_{\tau=1}^t \Esgd  \ns{\wt'-w^*}\indic{\calE_\tau} =& \sum_{\tau=1}^t \Esgd \br{\ns{\wt'-w^*}|\calE_\tau}\Pr[\calE_\tau]\\
\leq& 3L^2\sigma^2\sum_{\tau=1}^t \Pr[\calE_\tau]=3L^2\sigma^2\Pr[\bar{\calE}]\leq 3L^2\sigma^2\epsilon.
\end{align*}
This then implies that $\absr{\Esgd  \ns{\wt'-w^*} - \Esgd  \ns{\wt'-w^*}\indic{\calE}}\leq 3L^2 \sigma^2\epsilon.$

Finally, we have
\begin{align*}
&\absr{\Esgd \ns{\wt-w^*}-\Esgd \ns{\wt'-w^*} }\\
\leq& \absr{\Esgd \ns{\wt-w^*}-\Esgd  \ns{\wt'-w^*}\indic{\calE}}+\absr{\Esgd  \ns{\wt'-w^*} - \Esgd  \ns{\wt'-w^*}\indic{\calE}}\\
\leq& \pr{3L^2\sigma^2+1}\epsilon
\end{align*}
as long as $\eta\leq \frac{1}{c_5 d^2\log^2(d/\epsilon)}$. Therefore, $\absr{Q(\eta)-Q'(\eta)}\leq \pr{3L^2\sigma^2+1}\epsilon/2$.
Choosing $\epsilon'=\frac{2\epsilon}{\pr{3L^2\sigma^2+1}}$ finishes the proof.
\end{proofof}

\begin{proofof}{Lemma~\ref{lem:genera_fixed_eta_SGD}}
Recall that 
$$\hfva(\eta):=\frac{1}{m}\sum_{k=1}^m \dva = \frac{1}{m}\sum_{k=1}^m \Esgd \frac{1}{2}\ns{\wtk-\wvak}_{\hvak}.$$  
Similar as in Lemma~\ref{lem:genera_valid_GD}, we can show $\frac{1}{2}\ns{\wtk-\wvak}_{\hvak}$ is $O(1)$-subexponential, which implies \\$\Esgd \frac{1}{2}\ns{\wtk-\wvak}_{\hvak}$ is $O(1)$-subexponential. Therefore, $\hfva(\eta)$ is the average of $m$ i.i.d. $O(1)$-subexponential random variables. By standard concentration inequality, we know for any $1>\epsilon>0,$ with probability at least $1-\expepsm,$
$$\absr{\hfva(\eta)-\bfva(\eta) }\leq \epsilon.$$
\end{proofof}

\begin{proofof}{Lemma~\ref{lem:eps_net_expect_SGD}}
Recall that 
\begin{align*}
\bfva(\eta)=&\Ep \Esgd \frac{1}{2}\ns{\wt-w^*}+\sigma^2/2
\end{align*}
We only need to construct an $\epsilon$-net for $\Ep \Esgd \frac{1}{2}\ns{\wt-w^*}$.
Let $\calE$ be the event that  $\sqrt{d}/\sqrt{L} \leq \sigma_i(\xtr)\leq  \sqrt{Ld}$ and $1/L \leq \lambda_i(\htr) \leq L$ for all $i\in[n]$ and $\sqrt{d}\sigma/4 \leq \n{\xitr}\leq \sqrt{d}\sigma$ We have
\begin{align*}
&\Ep \Esgd \frac{1}{2}\ns{\wt-w^*} \\
=& \Ep \br{\frac{1}{2}\Esgd\ns{\wt-w^*}| \calE}\Pr[\calE]+ \Ep \br{\frac{1}{2}\Esgd\ns{\wt-w^*}| \bar{\calE} }\Pr[\bar{\calE}]
\end{align*}

According to Lemma~\ref{lem:q_n_qprime}, we know conditioning on $\calE,$ 
$$\absr{\frac{1}{2}\Esgd\ns{\wt-w^*}-\frac{1}{2}\Esgd\ns{\wt'-w^*}}\leq \epsilon,$$
as long as $\eta\leq \frac{1}{c_5 d^2\log^2(d/\epsilon)}.$ Note $\{w_{\tau,\eta}'\}$ is the SGD sequence without truncation.

For the second term,
we have 
\begin{align*}
\Ep \br{\frac{1}{2}\Esgd \ns{\wt-w^*}| \bar{\calE} }\Pr[\bar{\calE}]
\leq 13L\sigma^2\Pr[\bar{\calE}]
\leq \epsilon,
\end{align*}
where the last inequality assumes $\Pr[\bar{\calE}]\leq \frac{\epsilon}{13L\sigma^2}.$ According to Lemma~\ref{lem:isotropic} and Lemma~\ref{lem:norm_vector}, we know $\Pr[\bar{\calE}]\leq \expd.$ Therefore, given any $\epsilon>0,$ we have $\Pr[\bar{\calE}]\leq \frac{\epsilon}{13L\sigma^2}$ as long as $d\geq c_4\log(1/\epsilon)$ for some constant $c_4$.

Then, we only need to construct an $\epsilon$-net for $\Ep \br{\frac{1}{2}\Esgd\ns{\wt'-w^*}| \calE}\Pr[\calE].$
By the analysis in Lemma~\ref{lem:large_step_SGD}, it's not hard to prove
$$\absr{\frac{\partial}{\partial \eta} \Ep \br{\frac{1}{2}\Esgd\ns{\wt'-w^*}| \calE}\Pr[\calE]}=O(1)t(1-\frac{\eta}{2L})^{t-1},$$
for all $\eta\in[0,\frac{1}{c_5 d^2\log^2(d/\epsilon)}].$ Similar as in Lemma~\ref{lem:crossing_term2}, for any $\epsilon>0,$ we know there exists an $\epsilon$-net $N_\epsilon$ with size $O(1/\epsilon)$ such that for any $\eta\in[0,\frac{1}{c_5 d^2\log^2(d/\epsilon)}],$ 
$$\absr{\Ep \br{\frac{1}{2}\Esgd\ns{\wt'-w^*}| \calE}\Pr[\calE]-\Ep \br{\frac{1}{2}\Esgd\ns{w_{t,\eta'}'-w^*}| \calE}\Pr[\calE]}\leq \epsilon$$ 
for $\eta'\in\arg\min_{\eta\in N_\epsilon}|\eta-\eta'|.$ 

Combing with the bounds on $\absr{\frac{1}{2}\Esgd\ns{\wt-w^*}\indic{\calE}-\frac{1}{2}\Esgd\ns{\wt'-w^*}\indic{\calE}}$ and \\$\Ep \br{\frac{1}{2}\Esgd\ns{\wt-w^*}| \bar{\calE} }\Pr[\bar{\calE}]$, we have for any $\eta\in[0,\frac{1}{c_5 d^2\log^2(d/\epsilon)}],$ 
$$\bfva(\eta)-\bfva(\eta')\leq 4\epsilon$$ 
for $\eta'\in\arg\min_{\eta\in N_\epsilon}|\eta-\eta'|.$ 
We finish the proof by replacing $4\epsilon$ by $\epsilon'.$
\end{proofof}

\begin{proofof}{Lemma~\ref{lem:eps_net_empirical_SGD}}
The proof is very similar as the proof of Lemma~\ref{lem:eps_net_empirical_valid_GD}. The only difference is that we need to first relate the SGD sequence with truncation to the SGD sequence without truncation and then bound the Lipschitzness on the SGD sequence without truncation (as we did in Lemma~\ref{lem:eps_net_expect_SGD}). We omit the details here.
\end{proofof}

\section{Tools}
\subsection{Norm of random vectors}
We use the following lemma to bound the noise in least squares model.

\begin{lemma}[Theorem 3.1.1 in~\cite{vershynin2018high}]\label{lem:norm_vector}
Let $X=(X_1, X_2, \cdots, X_n)\in \R^n$ be a random vector with each entry independently sampled from $\mathcal{N}(0,1).$
Then
$$\Pr[\absr{\n{x}-\sqrt{n}}\geq t]\leq 2\exp(-t^2/C^2),$$
where $C$ is an absolute constant.
\end{lemma}

\subsection{Singular values of Gaussian matrices}

Given a random Gaussian matrix, in expectation its smallest and largest singular value can be bounded as follows.
\begin{lemma}[Theorem 5.32 in~\cite{vershynin2010introduction}]\label{lem:sig_matrix_expect}
Let $A$ be an $N\times n$ matrix whose entries are independent standard normal random variables. Then
$$\sqrt{N}-\sqrt{n} \leq \E s_{\min}(A)\leq \E s_{\max}(A)\leq \sqrt{N}+\sqrt{n}$$
\end{lemma}

Lemma~\ref{lem:concen_gauss} shows a lipchitz function over i.i.d. Gaussian variables concentrate well on its mean. We use this lemma to argue for any fixed step size, the empirical meta objective concentrates on the population meta objective.

\begin{lemma}[Proposition 5.34 in~\cite{vershynin2010introduction}]\label{lem:concen_gauss}
Let $f$ be a real valued Lipschitz function on $\R^n$ with Lipschitz constant $K$. Let $X$ be the standard normal random vector in $\R^n.$ Then for every $t\geq 0$ one has 
$$\Pr[f(X)-\E f(X)\geq t]\leq \exp(-\frac{t^2}{2K^2}).$$
\end{lemma}

The following lemma shows a tall random Gaussian matrix is well-conditioned with high probability. The proof follows from Lemma~\ref{lem:sig_matrix_expect} and Lemma~\ref{lem:concen_gauss}. We use Lemma~\ref{lem:sig_matrix} to show the covariance matrix is well conditioned in the least squares model.

\begin{lemma}[Corollary 5.35 in~\cite{vershynin2010introduction}]\label{lem:sig_matrix}
Let $A$ be an $N\times n$ matrix whose entries are independent standard normal random variables. Then for every $t\geq 0$ with probability at least $1-2\exp(-t^2/2)$ one has 
$$\sqrt{N}-\sqrt{n}-t \leq s_{\min}(A)\leq s_{\max}(A)\leq \sqrt{N}+\sqrt{n}+t$$
\end{lemma}

\subsection{Johnson-Lindenstrauss lemma}
We also use Johnson-Lindenstrauss Lemma in some of the lemmas. Johnson-Lindenstrauss Lemma tells us the projection of a fixed vector on a random subspace concentrates well as long as the subspace is reasonably large.

\begin{lemma}[\cite{johnson1984extensions}]\label{lem:JL_subspace}
Let $P$ be a projection in $\R^d$ onto a random $n$-dimensional subspace uniformly distributed in $G_{d,n}.$ Let $z\in\R^d$ be a fixed point and $\epsilon>0$, then with probability at least $1-2\exp(-c\epsilon^2 n),$
$$(1-\epsilon)\sqrt{\frac{n}{d}}\n{z}\leq \n{Pz}\leq (1+\epsilon)\sqrt{\frac{n}{d}}\n{z}.$$ 
\end{lemma}
\section{Experiment details}\label{sec:experiment_details}
We describe the detailed settings of our experiments in Section~\ref{sec:exp_setting} and give more experimental results in Section~\ref{sec:exp_more}.

\subsection{Experiment settings}\label{sec:exp_setting}

\paragraph{Optimizing step size for quadratic objective}

In this experiment, we meta-train a learning rate for gradient descent on a fixed quadratic objective. Our goal is to show that the autograd module in popular deep learning softwares, such as Tensorflow, can have numerical issues when using the log-transformed meta objective. Therefore, we first implement the meta-training process with Tensorflow to see the results. We then re-implement the meta-training using the hand-derived meta-gradient (see Eqn~\ref{eqn:metagrad_hand}) to compare the result.

A general setting for both implementations is as follows. The inner problem is fixed as a 20-dimensional quadratic objective as described in Section \ref{sec:quadratic}, and we use the log-transformed meta objective for training. The positive semi-definite matrix $H$ is generated by first sampling a $20\times20$ matrix $X$ with all entries drawn from the standard normal distribution and then setting $H=X^TX$. The initial point $w_0$ is drawn from standard normal as well. Note that we use the same quadratic problem (i.e., the same $H$ and $w_0$) throughout the meta-training. We do 1000 meta-training iterations, and collect results for different settings of the initial learning rate $\eta_0$ and the unroll length $t$.

We first implement the meta-training code with Tensorflow. Our code is adapted from \cite{wichrowska2017learned} \footnote{Their open source code is available at \url{https://github.com/tensorflow/models/tree/master/research/learned_optimizer}}. We use their global learning rate optimizer and specify the problem set to have only one quadratic objective instance. We implemented the quadratic objective class ourselves (the "MyQuadratic" class). We also turned off multiple advanced features in the original code, such as attention and second derivatives, by assigning their flags as false. This ensures that the experiments have exactly the same settings as we described. The meta-training learning rate is set to be 0.001, which is of similar scale as our next experiment.
We also try RMSProp as the meta optimizer, which alleviates some of the numerical issues as it renormalizes the gradient, but our experiments show that even RMSProp is still much worse than our implementation. 

We then implement the meta-training by hand to show the accurate training results that avoid numerical issues. Specifically, we compute the meta-gradient using Eq (\ref{eqn:metagrad_hand}), where we also scaled the numerator and denominator as described in Claim \ref{clm:max} to avoid numerical issues. We use the algorithm suggested in Theorem~\ref{thm:quadratic_converge}, except we choose the meta-step size to be $1/(100\sqrt{k})$ as the constants in Theorem~\ref{thm:quadratic_converge} were not optimized.


\paragraph{Train-by-train vs. train-by-validation, synthetic data}

In this experiment, we find the optimal learning rate $\eta^*$ for least-squares problems trained in train-by-train and train-by-validation settings and then see how the learning rate works on new tasks.

Specifically, we generate 300 different 1000-dimensional least-squares tasks with noise as defined in Section \ref{sec:least-squares}  for inner-training and then use the meta-objectives defined in Eq (\ref{eqn:train_GD}) and (\ref{eqn:valid_GD}) to find the optimal learning rate. The inner-training number of steps $t$ is set as 40. We try different sample sizes and different noise levels for comparison. Subsequently, in order to test how the two $\eta^*$ (for train-by-train and train-by-validation respectively) work, we use them on 10 test tasks (the same setting as the inner-training problem) and compute training and testing root mean squared error (RMSE).

Note that since we only need the final optimal $\eta^*$ found under the two meta-objective settings (regardless of how we find it), we do not need to actually do the meta-training. Instead, we do a grid search on the interval $[10^{-6}, 1]$, which is divided log-linearly to 25 candidate points. For both the train-by-train and train-by-validation settings, we average the meta-objectives over the 300 inner problems and see which $\eta$ minimizes this averaged meta-objective. The results are shown in Appendix \ref{sec:exp_more}.

\paragraph{Train-by-train vs. train-by-validation, MLP optimizer on MNIST}

To observe the trade-off between train-by-train and train-by-validation in a broader and more realistic case, we also do experiments to meta-train an MLP optimizer as in ~\cite{metz2019understanding} to solve the MNIST classification problem. We use part of their code \footnote{Their code is available at \url{https://github.com/google-research/google-research/tree/master/task_specific_learned_opt}} to integrate with our code in the first experiment, and we use exactly the same default setting as theirs, which is summarized below.

The MLP optimizer is a trainable optimizer that works on each parameter separately. When doing inner-training, for each parameter, we first compute some statistics of that parameter (explained below), which are combined into a feature vector, and then feed that feature vector to a Muti-Layer Perceptron (MLP) with ReLU activations, which outputs two scalars, the update direction and magnitude. The update is computed as the direction times the exponential of the magnitude. The feature vector is 31-dimensional, which includes gradient, parameter value, first-order moving averages (5-dim), second-order moving averages (5-dim), normalized gradient (5-dim), reciprocal of square root second-order moving averages (5-dim) and a step embedding (9-dim). All moving averages are computed using 5 different decay rates (0.5, 0.9, 0.99, 0.999, 0.9999), and the step embedding is $\tanh$ distortion of the current number of steps divided by 9 different scales (3, 10, 30, 100, 300, 1000, 3000, 10000, 300000). After expanding the 31-dimensional feature vector for each parameter, we also normalize the set of vectors dimension-wise across all the parameters to have mean 0 and standard deviation 1 (except for the step embedding part). More details can be found in their original paper and original implementation.

The inner-training problem is defined as using a two-layer fully connected network (i.e., another ``MLP'') with ReLU activations to solve the classic MNIST 10-class classification problem. We use a very small network for computational efficiency, and the two layers have 100 and 20 neurons. We fix the cross-entropy loss as the inner-objective and use mini-batches of 32 samples when inner-training.

When we meta-train the MLP optimizer, we use exactly the same process as fixed in experiments by \cite{wichrowska2017learned}. We use 100 different inner problems by shuffling the 10 classes and also sampling a new subset of data if we do not use the complete MNIST data set. We run each of the problems with three inner-training trajectories starting with different initialization. Each inner-training trajectory is divided into a certain number of unrolled segments, where we compute the meta-objective and update the meta-optimizer after each segment. The number of unrolled segments in each trajectory is sampled from $10+\text{Exp}(30)$, and the length of each segment is sampled from $50+\text{Exp}(100)$, where $\text{Exp}(\cdot)$ denotes the exponential distribution. Note that the meta-objective computed after each segment is defined as the average of all the inner-objectives (evaluated on the train/validation set for train-by-train/train-by-val) within that segment for a better convergence. 
We also do not need to log-transform the inner-objective this time because the cross entropy loss has a log operator itself. The meta-training, i.e. training the parameters of the MLP in the MLP optimzier, is completed using a classic RMSProp optimizer with meta learning rate 0.01.

For each settings of sample sizes and noise levels, we train two MLP optimizer: one for train-by-train, and one for train-by-validation. When we test the learned MLP optimizer, we use similar settings as the inner-training problem, and we run the trajectories longer for full convergence (4000 steps for small data sets; 40000 steps for the complete data set). We run 5 independent tests and collect training accuracy and test accuracy for evaluation. The plots show the mean of the 5 tests. We have also tuned a SGD optimizer (with the same mini-batch size) by doing a grid-search of the learning rate as baseline.

\subsection{Additional results}\label{sec:exp_more}

\paragraph{Optimizing step size for quadratic objective}

We try experiments for the same settings of the initial $\eta_0$ and inner training length $t$ for all of three implementations (our hand-derived GD version, Tensorflow GD version and the Tensorflow RMSProp version). We do 1000 meta-training steps for all the experiments.

For both Tensorflow versions, we always see infinite meta-objectives if $\eta_0$ is large or $t$ is large, whose meta-gradient is usually treated as zero, so the training get stuck and never converge. Even for the case that both $\eta_0$ and $t$ is small, it still has very large meta-objectives (the scale of a few hundreds), and that is why we also try RMSProp, which should be more robust against the gradient scales. Our hand-derived version, however, does not have the numerical issues and can always converge to the optimal $\eta^*$. The detailed convergence is summarized in Tab \ref{tab:qua1} and Tab \ref{tab:qua2}. Note that the optimal $\eta^*$ is usually around 0.03 under our settings.

\begin{table}[]
\centering
\caption{Whether the implementation converges for different $t$ (fixed $\eta_0 = 0.1$)}
\label{tab:qua1}
\begin{tabular}{c|cccc}
\hline
$t$                & 10          & 20          & 40 & 80 \\ \hline
Ours               & $\checkmark$ & $\checkmark$ &  $\checkmark$  & $\checkmark$   \\
Tensorflow GD      & $\times$   &  $\times$           & $\times$   &  $\times$  \\
Tensorflow RMSProp & $\checkmark$   &   $\checkmark$          & $\times$   &  $\times$  \\ \hline
\end{tabular}
\end{table}

\begin{table}[]
\centering
\caption{Whether the implementation converges for different $\eta_0$ (fixed $t = 40$)}
\label{tab:qua2}
\begin{tabular}{c|cccc}
\hline
$\eta_0$                & 0.001          & 0.01          & 0.1 & 1 \\ \hline
Ours               & $\checkmark$ & $\checkmark$ &  $\checkmark$  & $\checkmark$   \\
Tensorflow GD      & $\times$   &  $\times$           & $\times$   &  $\times$  \\
Tensorflow RMSProp & $\checkmark$   &   $\checkmark$          & $\times$   &  $\times$  \\ \hline
\end{tabular}
\end{table}

\paragraph{Train-by-train vs. train-by-validation, MLP optimizer on MNIST}

\begin{figure}[h]
\begin{minipage}[t]{\linewidth}
     \centering
     \includegraphics[width=3in]{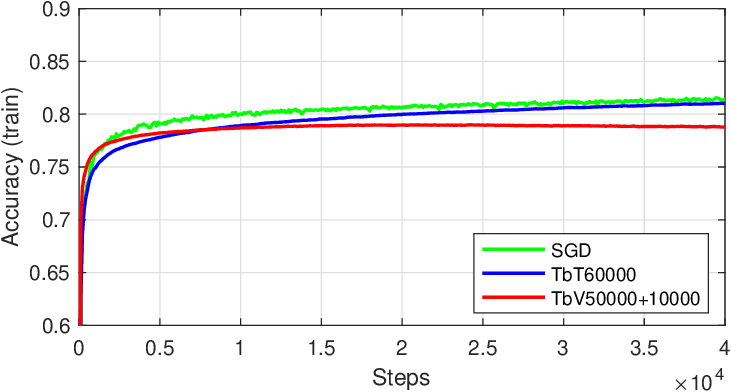}
\end{minipage}
\hfill
\begin{minipage}[t]{\linewidth}
     \centering
     \includegraphics[width=3in]{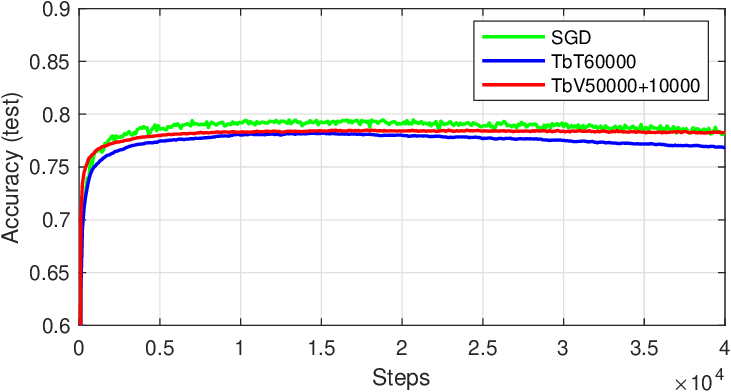}
\end{minipage}
\caption{Training and testing accuracy for different models (all samples, 20\% noise)}
\label{fig:mnist_acc_noise60000}
\end{figure}

We also do additional experiments on training an MLP optimizer on the MNIST classification problem. We first try using all samples under the 20\% noised setting. The results are shown in Fig \ref{fig:mnist_acc_noise60000}. The train-by-train setting can perform well if we have a large data set, but since there is also noise in the data, the train-by-train model still overfits and is slightly worse than the train-by-validation model.

\begin{figure}[h]
\begin{minipage}[t]{\linewidth}
     \centering
     \includegraphics[width=3in]{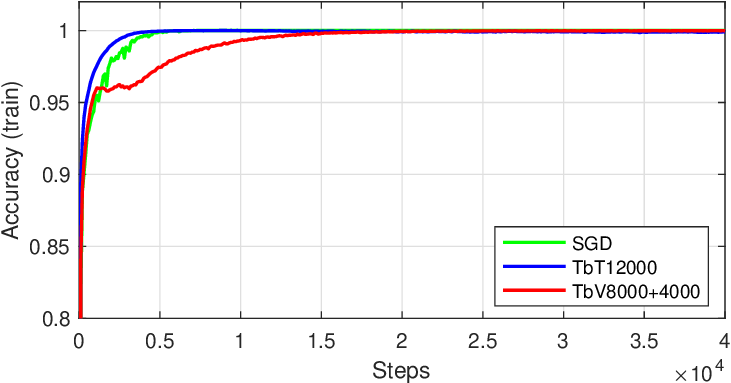}
\end{minipage}
\hfill
\begin{minipage}[t]{\linewidth}
     \centering
     \includegraphics[width=3in]{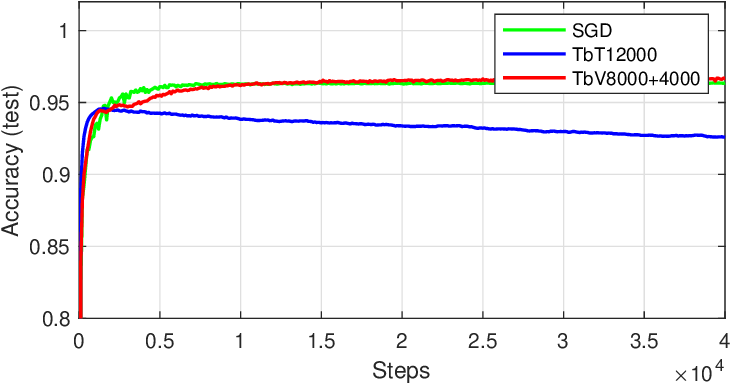}
\end{minipage}
\caption{Training and testing accuracy for different models (12000 samples, no noise)}
\label{fig:mnist_acc_12000}
\end{figure}

\begin{figure}[h]
\begin{minipage}[t]{\linewidth}
     \centering
     \includegraphics[width=3in]{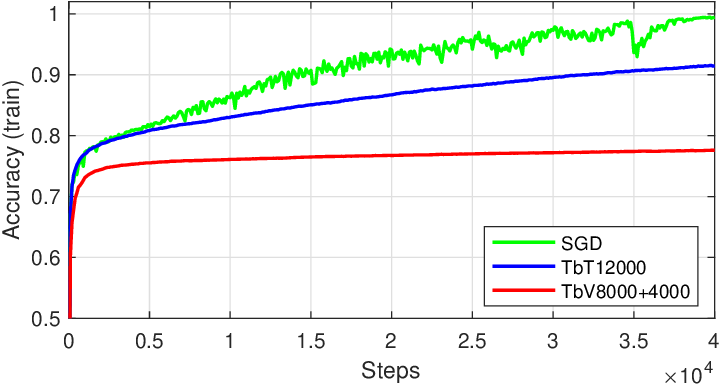}
\end{minipage}
\hfill
\begin{minipage}[t]{\linewidth}
     \centering
     \includegraphics[width=3in]{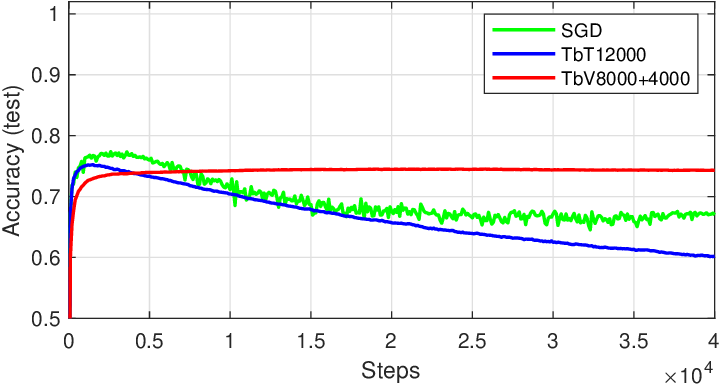}
\end{minipage}
\caption{Training and testing accuracy for different models (12000 samples, 20\% noise)}
\label{fig:mnist_acc_noise12000}
\end{figure}

We then try an intermediate sample size 12000. The results are shown in Fig \ref{fig:mnist_acc_12000} (no noise) and Fig \ref{fig:mnist_acc_noise12000} (20\% noise). We can see that as the theory predicts, as the amount of data increases (from 1000 samples to 12000 samples and then to 60000 samples) the gap between train-by-train and train-by-validation decreases. Also, when we condition on the same number of samples, having additional label noise always makes train-by-train model much worse compared to train-by-validation. 

\end{document}